\documentclass[9pt,compsoc]{IEEEtran}

\usepackage{hyperref}
\usepackage{url}
\usepackage{amsmath}
\usepackage{amssymb}
\usepackage{amsthm}
\usepackage{graphicx}
\usepackage{epstopdf}
\usepackage{float}
\usepackage[ruled,vlined]{algorithm2e}

\DeclareMathOperator{\Tr}{tr}
\DeclareMathOperator{\diag}{diag}
\DeclareMathOperator{\rank}{rank}

\begin{document}

\def\httilde{\mbox{\tt\raisebox{-.5ex}{\symbol{126}}}}
\def\balpha{\mbox{\boldmath $\alpha$}}
\def\bDelta{{\bf \Delta}}
\def\bLambda{{\bf \Lambda}}
\def\bvarphi{{\bf \varphi}}
\def\bTheta{{\bf \Theta}}
\def\btheta{\mbox{\boldmath $\theta$}}

\def\bPhi{\mbox{\boldmath{$\Phi$}}}
\def\vbphi{\vec{\mbox{\boldmath $\phi$}}}
\def\bb{{\bf b}}
\def\h{{\bf h}}
\def\bd{{\bf d}}
\def\ba{{\bf a}}
\def\bc{{\bf c}}
\def\p{{\bf p}}
\def\e{{\bf e}}
\def\s{{\bf s}}
\def\X{{\bf X}}
\def\x{{\bf x}}
\def\Y{{\bf Y}}
\def\y{{\bf y}}
\def\K{{\bf K}}
\def\k{{\bf k}}
\def\p{{\bf p}}
\def\bc{{\bf c}}
\def\A{{\bf A}}
\def\B{{\bf B}}
\def\C{{\bf C}}
\def\V{{\bf V}}
\def\S{{\bf S}}
\def\T{{\bf T}}
\def\W{{\bf W}}
\def\I{{\bf I}}
\def\U{{\bf U}}
\def\g{{\bf g}}
\def\G{{\bf G}}
\def\Q{{\bf Q}}
\def\d{{\bf d}}
\def\eg{{\it e.g.}}
\def\etal{{\it et. al}}
\def\H{{\bf H}}
\def\cR{{\bf R}}
\def\J{{\bf J}}
\def\bt{{\bf t}}
\def\bv{{\bf v}}
\def\R{{\bf R}}

\def\balpha{\mbox{\boldmath $\alpha$}}
\def\bdelta{\mbox{\boldmath $\delta$}}
\def\bzeta{\mbox{\boldmath $\zeta$}}
\def\bphi{\mbox{\boldmath $\phi$}}
\def\btau{\mbox{\boldmath $\tau$}}
\def\bmu{\mbox{\boldmath $\mu$}}
\def\bsigma{\mbox{\boldmath $\sigma$}}
\def\bSigma{{\bm \Sigma} }
\def\btheta{\mbox{\boldmath $\theta$}}
\def\dbphi{\dot{\mbox{\boldmath $\phi$}}}
\def\dbtau{\dot{\mbox{\boldmath $\tau$}}}
\def\dbtheta{\dot{\mbox{\boldmath $\theta$}}}
\def\bGamma{\mbox{\boldmath $\Gamma$}}
\def\bDelta{\mbox{\boldmath $\Delta$}}
\def\blambda{\mbox{\boldmath $\lambda $}}
\def\bOmega{\mbox{\boldmath $\Omega $}}
\def\bbeta{\mbox{\boldmath $\beta $}}
\def\bupsilon{\mbox{\boldmath $\Upsilon$}}
\def\myphi{\phi}
\def\bPhi{\mbox{\boldmath{$\Phi$}}}
\def\bLambda{\mbox{\boldmath{$\Lambda$}}}
\def\bSigma{\mbox{\boldmath{$\Sigma$}}}

\def\balpha{\mbox{\boldmath{$\alpha$}}}
\def\bbeta{\mbox{\boldmath{$\beta$}}}
\def\bdelta{\mbox{\boldmath{$\delta$}}}
\def\bgamma{\mbox{\boldmath{$\gamma$}}}
\def\blambda{\mbox{\boldmath{$\lambda$}}}
\def\bsigma{\mbox{\boldmath{$\sigma$}}}
\def\btheta{\mbox{\boldmath{$\theta$}}}
\def\bomega{\mbox{\boldmath{$\omega$}}}
\def\bxi{\mbox{\boldmath{$\xi$}}}

\def\bigO2{\mbox{${\cal O}$}}
\def\bigO{O}

\newcommand{\bH}{\mathbf{H}}
\def\mA{\mathcal{A}}
\def\mB{\mathcal{B}}
\def\mC{\mathcal{C}}
\def\mD{\mathcal{D}}
\def\mG{\mathcal{G}}
\def\mV{\mathcal{V}}
\def\mE{\mathcal{E}}
\def\mF{\mathcal{F}}
\def\mH{\mathcal{H}}
\def\mL{\mathcal{L}}
\def\mM{\mathcal{M}}
\def\mN{\mathcal{N}}
\def\mK{\mathcal{K}}
\def\mR{\mathcal{R}}
\def\mS{\mathcal{S}}
\def\mT{\mathcal{T}}
\def\mU{\mathcal{U}}
\def\mW{\mathcal{W}}
\def\mX{\mathcal{X}}
\def\mY{\mathcal{Y}}
\def\1n{\mathbf{1}_n}
\def\0{\mathbf{0}}
\def\1{\mathbf{1}}
\def\etal{{\em et al.}}

\def\balpha{\mbox{\boldmath $\alpha$}}
\def\bdelta{\mbox{\boldmath $\delta$}}
\def\bzeta{\mbox{\boldmath $\zeta$}}
\def\bphi{\mbox{\boldmath $\phi$}}
\def\btau{\mbox{\boldmath $\tau$}}
\def\bmu{\mbox{\boldmath $\mu$}}
\def\bsigma{\mbox{\boldmath $\sigma$}}
\def\bSigma{{\bm \Sigma} }
\def\btheta{\mbox{\boldmath $\theta$}}
\def\dbphi{\dot{\mbox{\boldmath $\phi$}}}
\def\dbtau{\dot{\mbox{\boldmath $\tau$}}}
\def\dbtheta{\dot{\mbox{\boldmath $\theta$}}}
\def\bGamma{\mbox{\boldmath $\Gamma$}}
\def\bDelta{\mbox{\boldmath $\Delta$}}
\def\blambda{\mbox{\boldmath $\lambda $}}
\def\bOmega{\mbox{\boldmath $\Omega $}}
\def\bbeta{\mbox{\boldmath $\beta $}}
\def\bupsilon{\mbox{\boldmath $\Upsilon$}}
\def\myphi{\phi}
\def\bPhi{\mbox{\boldmath{$\Phi$}}}
\def\bLambda{\mbox{\boldmath{$\Lambda$}}}
\def\bSigma{\mbox{\boldmath{$\Sigma$}}}

\def\balpha{\mbox{\boldmath{$\alpha$}}}
\def\bbeta{\mbox{\boldmath{$\beta$}}}
\def\bdelta{\mbox{\boldmath{$\delta$}}}
\def\bgamma{\mbox{\boldmath{$\gamma$}}}
\def\blambda{\mbox{\boldmath{$\lambda$}}}
\def\bsigma{\mbox{\boldmath{$\sigma$}}}
\def\btheta{\mbox{\boldmath{$\theta$}}}
\def\bomega{\mbox{\boldmath{$\omega$}}}
\def\bxi{\mbox{\boldmath{$\xi$}}}

\def\bPsi{\mbox{\boldmath $\Psi $}}
\def\bone{\mbox{\bf 1}}
\def\bzero{\mbox{\bf 0}}

\def\WB{{\bf WB}}

\def\A{{\bf A}}
\def\B{{\bf B}}
\def\C{{\bf C}}
\def\D{{\bf D}}
\def\E{{\bf E}}
\def\F{{\bf F}}
\def\G{{\bf G}}
\def\H{{\bf H}}
\def\I{{\bf I}}
\def\J{{\bf J}}
\def\K{{\bf K}}
\def\L{{\bf L}}
\def\M{{\bf M}}
\def\N{{\bf N}}
\def\O{{\bf O}}
\def\P{{\bf P}}
\def\Q{{\bf Q}}
\def\R{{\bf R}}
\def\S{{\bf S}}
\def\T{{\bf T}}
\def\U{{\bf U}}
\def\V{{\bf V}}
\def\W{{\bf W}}
\def\X{{\bf X}}
\def\Y{{\bf Y}}
\def\Z{{\bf Z}}

\def\b{{\bf b}}
\def\bc{{\bf c}}
\def\bd{{\bf d}}
\def\e{{\bf e}}
\def\f{{\bf f}}
\def\g{{\bf g}}
\def\h{{\bf h}}
\def\i{{\bf i}}
\def\j{{\bf j}}
\def\k{{\bf k}}
\def\l{{\bf l}}
\def\m{{\bf m}}
\def\n{{\bf n}}
\def\o{{\bf o}}
\def\p{{\bf p}}
\def\q{{\bf q}}
\def\br{{\bf r}}
\def\s{{\bf s}}
\def\t{{\bf t}}
\def\u{{\bf u}}
\def\v{{\bf v}}
\def\w{{\bf w}}
\def\bx{{\bf x}}
\def\y{{\bf y}}
\def\z{{\bf z}}

\def\vbphi{\vec{\mbox{\boldmath $\phi$}}}
\def\vbtau{\vec{\mbox{\boldmath $\tau$}}}
\def\vbtheta{\vec{\mbox{\boldmath $\theta$}}}
\def\vI{\vec{\bf I}}
\def\vR{\vec{\bf R}}
\def\vV{\vec{\bf V}}

\def\mvec{\vec{m}}
\def\fvec{\vec{f}}
\def\appfvec{\vec{f}_k}
\def\avec{\vec{a}}
\def\bvec{\vec{b}}
\def\evec{\vec{e}}
\def\uvec{\vec{u}}
\def\xvec{\vec{x}}
\def\wvec{\vec{w}}
\def\gradvec{\vec{\nabla}}

\def\aM{\mbox{\bf a}_M}
\def\aS{\mbox{\bf a}_S}
\def\aO{\mbox{\bf a}_O}
\def\aL{\mbox{\bf a}_L}
\def\aP{\mbox{\bf a}_P}
\def\ai{\mbox{\bf a}_i}
\def\aj{\mbox{\bf a}_j}
\def\an{\mbox{\bf a}_n}
\def\a1{\mbox{\bf a}_1}
\def\a2{\mbox{\bf a}_2}
\def\a3{\mbox{\bf a}_3}
\def\a4{\mbox{\bf a}_4}

\def\sx{\mbox{\scriptsize\bf x}}
\def\st{\mbox{\scriptsize\bf t}}
\def\ss{\mbox{\scriptsize\bf s}}
\def\cR{{\cal R}}
\def\calD{{\cal D}}
\def\calS{{\cal S}}

\def\sigmae{\sigma}
\def\sigmam{\sigma}

\def\balpha{\mbox{\boldmath{$\alpha$}}}
\def\bbeta{\mbox{\boldmath{$\beta$}}}
\def\bdelta{\mbox{\boldmath{$\delta$}}}
\def\bgamma{\mbox{\boldmath{$\gamma$}}}
\def\blambda{\mbox{\boldmath{$\lambda$}}}
\def\bsigma{\mbox{\boldmath{$\sigma$}}}
\def\btheta{\mbox{\boldmath{$\theta$}}}
\def\bomega{\mbox{\boldmath{$\omega$}}}
\def\bxi{\mbox{\boldmath{$\xi$}}}

\def\dx{{\delta \x}}
\def\dref{{\d_{ref}}}
\def\px{{\partial \x}}
\def\fxp{\f(\x, \p)}

\def\dfp{\mathbf{d}(\mathbf{f}(\x,\mathbf{p}))}
\def\dfpk{\mathbf{d}(\mathbf{f}(\x,\mathbf{p}^k))}
\def\Ep{E(\mathbf{\d, \p})}
\newcommand{\deltap}[1]{\Delta^{#1}}
\newcommand{\Jp}[1]{\J^{#1}}
\newcommand{\Hp}[1]{\H^{#1}}
\newcommand{\Hpnewton}[1]{\H^{#1}_{nt}}
\newcommand{\Psip}[1]{\Psi^{#1}}
\newcommand{\Phip}[1]{\Phi^{#1}}
\newcommand{\dPsip}[1]{\Psi_{#1}}
\newcommand{\dPhip}[1]{\Phi_{#1}}

\newcommand{\dn}{\d_{s}}
\newcommand{\dc}{\d}
\newcommand{\dnxt}[1]{\dn(\f(\x, #1))}
\newcommand{\dcur}[1]{\dc(\f(\x, #1))}

\newcommand{\one}{\mathbf{1}}
\newcommand{\zero}{\mathbf{0}}
\newcommand{\real}{\mathbb{R}}

\newcommand{\denselist}{\itemsep -1pt}
\newcommand{\sparselist}{\itemsep 1pt}

\title{Error-Correcting Factorization}

\author{Miguel Angel Bautista, Oriol Pujol, Fernando De la Torre and Sergio Escalera 
\IEEEcompsocitemizethanks{\IEEEcompsocthanksitem Miguel Angel Bautista, Oriol Pujol and Sergio Escalera are with the Department of Applied Mathematics and Analisis, University of Barcelona, Barcelona, 08007, Spain and the Computer Vision Center, Autonomous University of Barcelona, 08193, Bellaterra (Cerdanyola), Barcelona, Spain  \protect\\
E-mail: $\{$mbautista,sescalera,oriol$\_$pujol$\}$@ub.edu
\IEEEcompsocthanksitem Fernando De la Torre is in the Robotics Institute at Carnegie Mellon University, Pittsburgh, 15216, PA.\protect\\
E-mail: ftorre@cs.cmu.edu}
\thanks{}}
\maketitle
\IEEEpeerreviewmaketitle
\markboth{MANUSCRIPT SUBMITTED TO IEEE TRANSACTIONS ON PATTERN ANALYSIS AND MACHINE INTELLIGENCE, 2015}{}

\begin{abstract}
Error Correcting Output Codes (ECOC) is a successful technique in multi-class classification, which is a core problem in Pattern Recognition and Machine Learning. A major advantage of ECOC over other methods is that the multi-class problem is decoupled into a set of binary problems that are solved independently.  However, literature defines a general error-correcting capability for ECOCs without analyzing how it distributes among classes, hindering a deeper analysis of pair-wise error-correction. To address these limitations this paper proposes an Error-Correcting Factorization (ECF) method, our contribution is three fold: (I) We propose a novel representation of the error-correction capability, called the design matrix, that enables us to build an ECOC on the basis of allocating correction to pairs of classes. (II) We derive the optimal code length of an ECOC using rank properties of the design matrix. (III) ECF is formulated as a discrete optimization problem, and a relaxed solution is found using an efficient constrained block coordinate descent approach. (IV) Enabled by the flexibility introduced with the design matrix we propose to allocate the error-correction on classes that are prone to confusion. Experimental results in several databases show that when allocating the error-correction to confusable classes ECF outperforms state-of-the-art approaches.
\end{abstract}

\section{Introduction}

In the last decade datasets have experimented an exponential growth rate, generating vast collections of data that need to automatically be analyzed. In particular, multimedia datasets have experienced an explosion on data availability, thanks to the almost negligible cost of gathering multi-media data from Internet. Therefore, there is a pushing need for efficient algorithms that are able to automatize knowledge extraction processes on those datasets. One of the classic problems in Pattern Recognition and Machine Intelligence is to perform automatic classification, i.e., automatically attributing a label to each sample of the dataset. In this sense, the classification process is often considered as first step for higher order representations or knowledge extractions. In multi-class classification problems the goal is to find a function $f: \mathbb{R}^n \rightarrow \mathbb{K}$, that maps samples to a finite discrete set $\mathbb{K}$ of labels with $|\mathbb{K}|>2$. While there exists a large set of approaches to estimate $f$ all of them can be grouped in two different categories: \textit{Single-Machine/Single-Loss} approaches and \textit{Divide and Conquer} approaches. The formers attempt to approximate a single $f$ for the complete multi-class problem, while the latter decouple $f$ into a set of binary sub-functions (binary classifiers) that are potentially easier to estimate and aggregate the results. 

\begin{figure}[!t]
\centering
\includegraphics[width=0.50\textwidth]{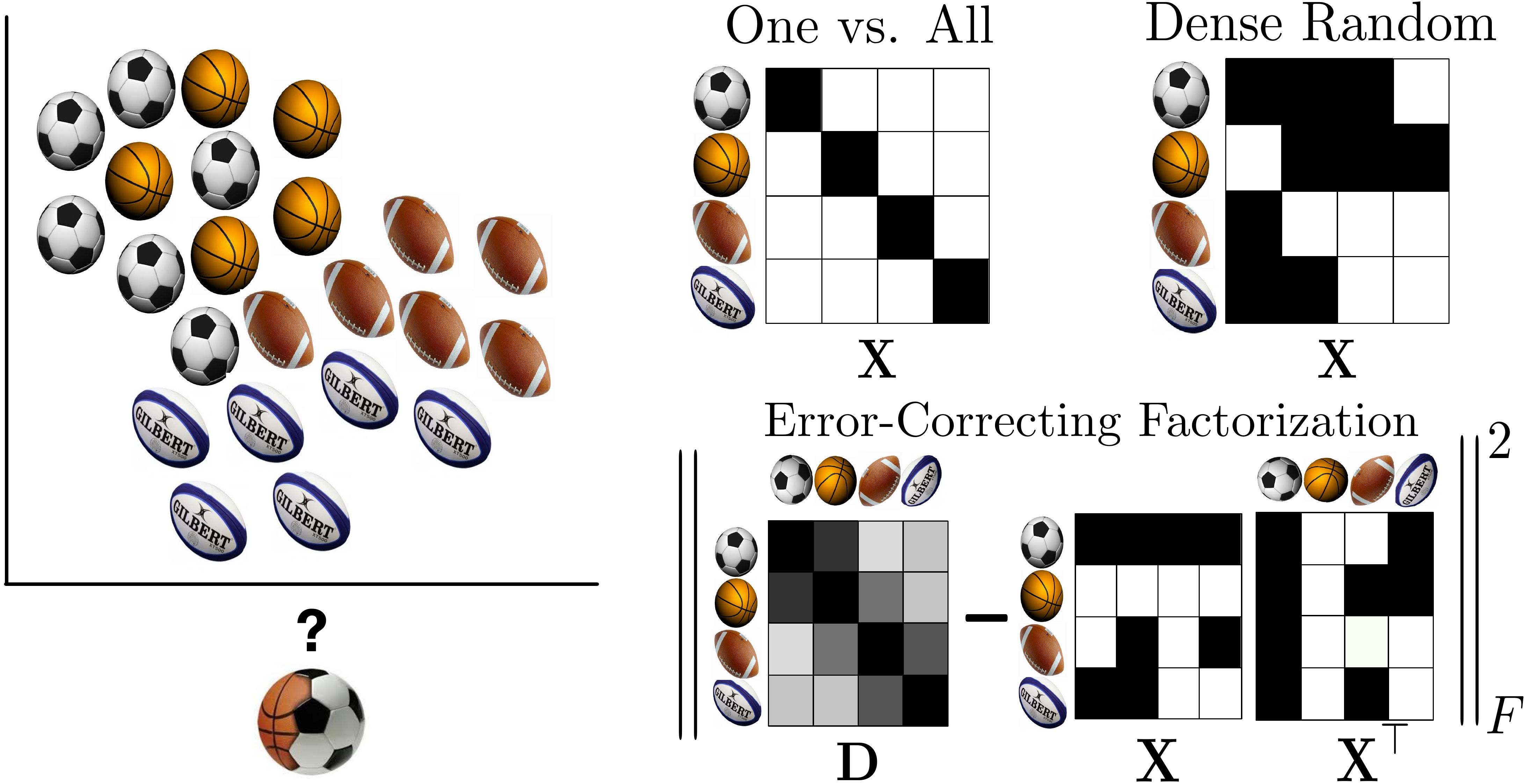}
\vspace{-0.7 cm}
\caption{Example of a classification problem of 4 different sports balls. Note how \textit{One vs. All} or \textit{Dense Random}
	ECOC designs do not take into account the data distribution while the proposed Error-Correcting Factorization method finds an ECOC matrix $\X$ by factorizing a design matrix $\D$. In addition, the codes (rows of $\X$) ECF assigns to similar categories are very dissimilar in order to benefit from Error-Correcting principles.}
\label{fig:introa}
\end{figure}

In this sense, Error-Correcting Output Codes (ECOC) is a divide and conquer approach that has proven to be very effective in many different multi-class contexts. The core property within an ECOC is its capability to correct errors in binary classifiers by using redundancy. However, existing literature represents the error-correcting capability of an ECOC as an scalar, hindering a deeper the analysis of error-correction and redundancy on class pairs. Furthermore, classical divide and conquer approaches that have been included in the ECOC framework like One vs. All \cite{rifkin} or Random \cite{random} approaches ignore the data distribution, thus not taking profit of allocating the error-correcting capabilities of ECOCs in a problem-dependent fashion. In addition, recent problem-dependent ECOC designs have focused on designing the binary sub-functions rather than analyzing the core error-correcting property. In order to overcome this limitations, our proposal builds an ECOC matrix by factorizing a design matrix $\D$ that encodes the desired 'correction properties' between classes (i.e a design matrix which can be obtained directly from data or be set by experts on the problem domain). The proposed method finds the ECOC coding that yields the closest configuration to the design matrix. We cast the task of designing an ECOC as a matrix factorization problem with binary constraints. A visual example is shown in Figure \ref{fig:introa}. 

\section{Related Work}
\subsection{Single-machine/Single-loss Approaches}

The multi-class problem can be directly treated by some methods that exhibit a multi-class behaviour off the shelf (i.e Nearest Neighbours \cite{nearestneighbour}, Decision Trees \cite{decisiontree}, Random Forests \cite{randomforest}). However, some of the most powerful methods for binary classification like Support Vector Machines (SVM) or Adaptive Boosting (AdaBoost) can not be directly extended to the multi-class case and further development is required. In this sense, literature is prolific on single-loss strategies to estimate $f$. One of the most well know approaches are the extensions of SVMs \cite{svmvapnik} to the multi-class case. For instance, the work of Weston and Watkins \cite{weston} presents a single-machine extension of the SVM method to cope with the multi-class case, in which $k$ predictor functions are trained, constrained with $k-1$ slack variables per sample. However, a more recent adaptation of \cite{crammerandsinger} reduces the number of constraints per samples to one, paying only for the second largest classification score among the $k$ predictors. To solve the optimization problem a dual decomposition algorithm is derived, which iteratively solves the quadratic programming problem associated with each training sample. Despite these efforts, single-machine approaches to estimate $f$ scale poorly with the number of classes and are often outperformed by simple decompositions \cite{rifkin,hastie}. In recent years various works that extended the classical Adaptive Boosting method \cite{adaboost} to the multi-class setting have been presented \cite{schapire,theory}. In \cite{multiclassAdaboost} the authors directly extend the AdaBoost algorithm to the multi-class case without reducing it to multiple binary problems, that is estimating a single $f$ for the whole multi-class problem. This algorithm is based on an exponential loss function for multi-class classification which is optimized on a forward stage-wise additive model. Furthermore, the work of Saberian and Vasconcenlos \cite{multiclassboosting} presents a derivation of a new margin loss function for multi-class classification altogether with the set of real class codewords that maximize the presented multi-class margin, yielding boundaries with max margin. However, though these methods are  consistently derived and supported with strong theoretical results, methodologies that jointly optimize a multi-class loss function present some limitations:

\begin{itemize}
\item They scale linearly with $k$, rendering them unsuitable for problems with a large $k$.

\item  Due to their single-loss architecture the exploitation of parallelization on modern multi-core processors is difficult. 

\item They can not recover from classification errors on the class predictors.
 \end{itemize}

\subsection{Divide and Conquer Approaches}

\begin{figure*}
	\centering
	\setlength{\tabcolsep}{1 pt}
	\begin{tabular}{c c c}
		\centering
		\includegraphics[width=0.33\textwidth]{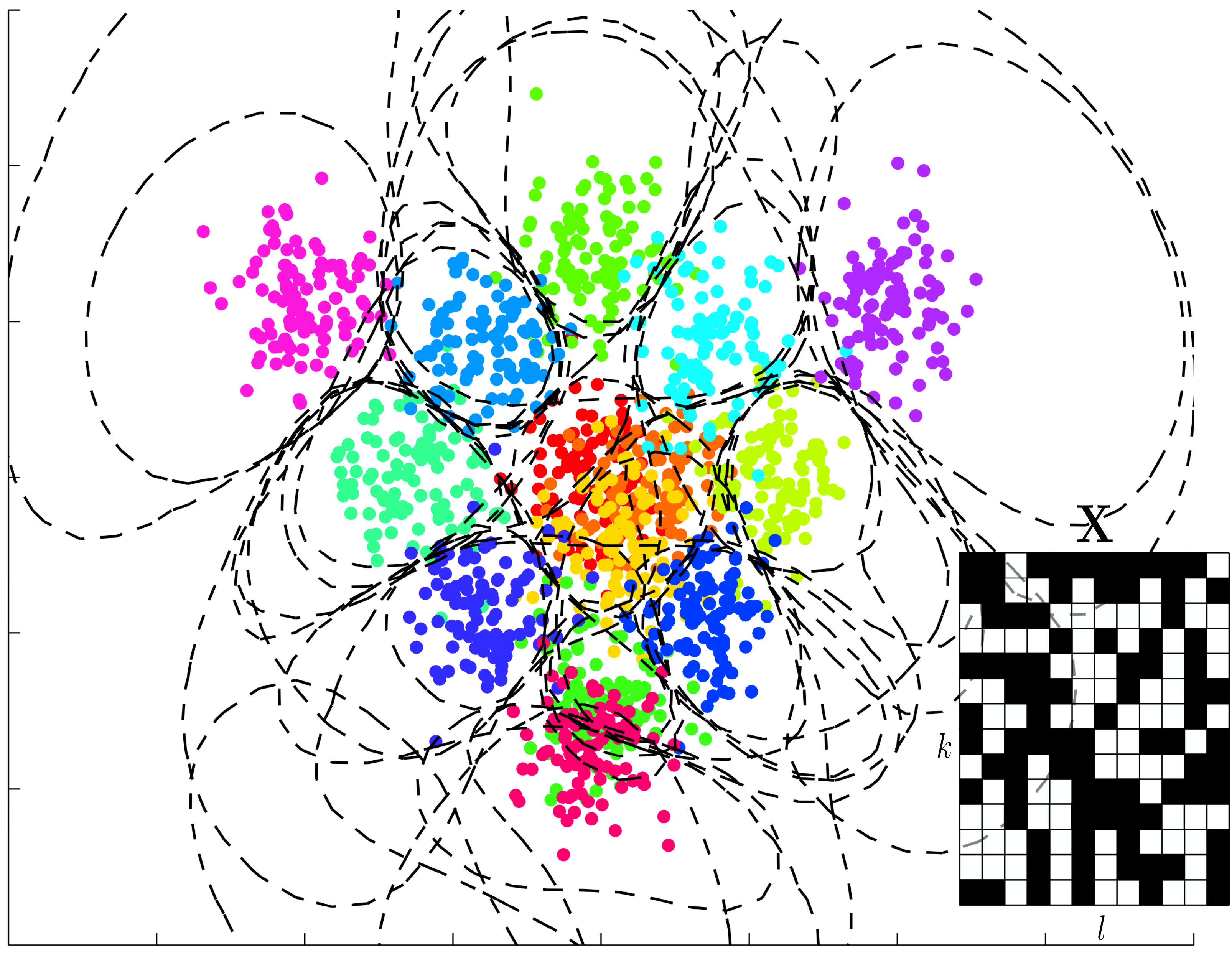} &
		\includegraphics[width=0.33\textwidth]{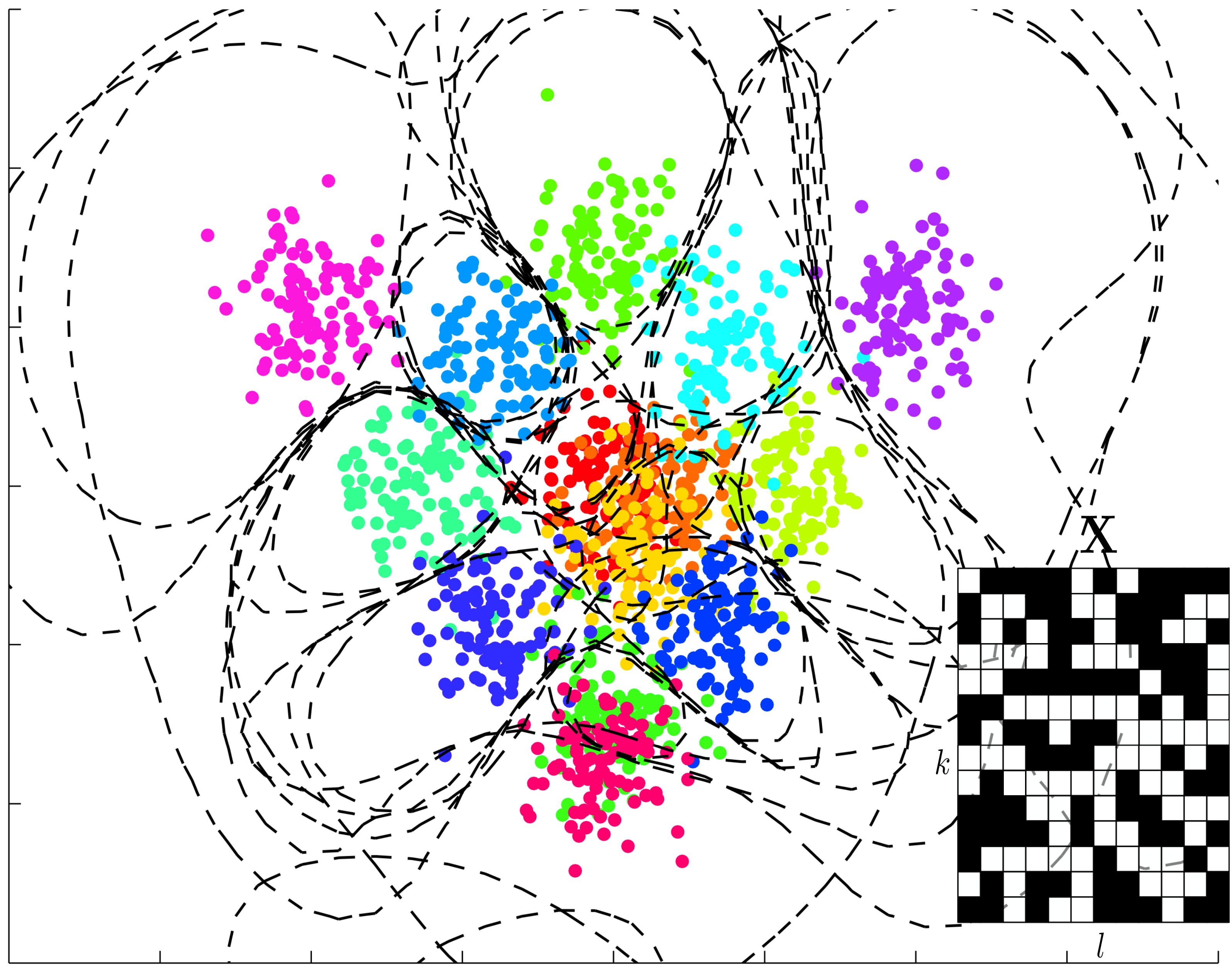} &
		\includegraphics[width=0.33\textwidth]{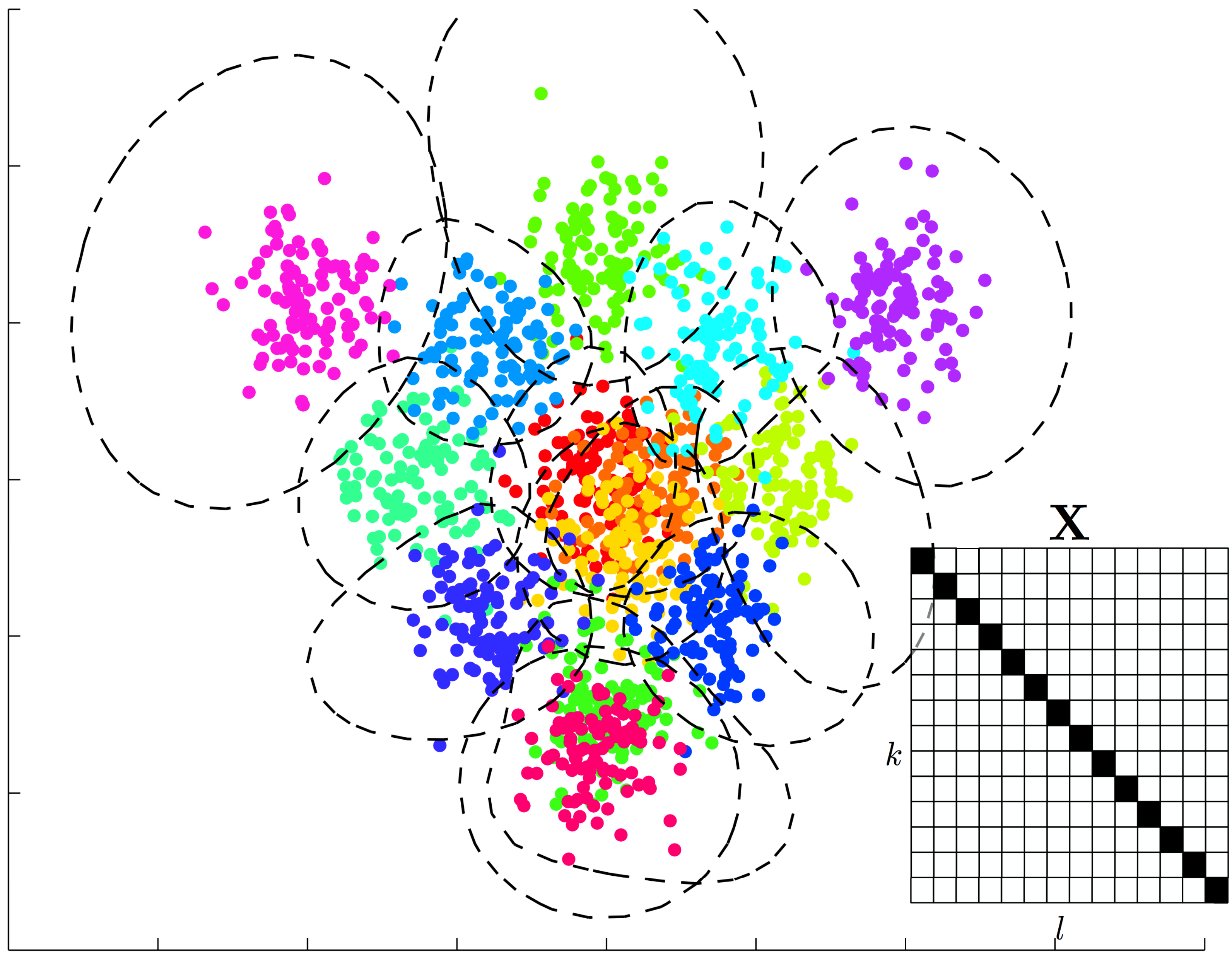} \\
		(a)&(b)&(c)\\
	\end{tabular}
	\vspace{-0.4 cm}
	\caption{(a) SVM RBF boundaries learned from Error-Correcting Factorization along with the ECOC coding matrix $\X$ in a Toy problem, $77.12\%$ classification accuracy (12 classifiers are trained). (b) Boundaries learned by the Dense Random ECOC coding design, $66.45\%$ classification accuracy (12 classifiers are trained). (c) SVM boundaries induced by the One vs. All approach, $49.53\%$ classification accuracy (14 classifiers are trained).}
	\label{fig:intro}
\end{figure*}

On the other hand, the divide and conquer approach has drawn a lot of attention due to its excellent results and easily parallelizable architecture \cite{rifkin,hastie,random,subclass,decoc,Bautista2013PR,marsalek,tree}. In this sense, instead of developing a method to cope with the multi-class case, divide and conquer approaches decouple $f$ into a set of $l$ binary problems which are treated separately. Once the responses of binary classifiers are obtained a committee strategy is used to find the final output. In this trend one can find three main lines of research: flat strategies, hierarchical classification, and ECOC. Flat strategies like One vs. One \cite{hastie} and One vs. All \cite{rifkin} are those that use a predefined problem partition scheme followed by a committee strategy to aggregate the binary classifier outputs. On the other hand, hierarchical classification relies on a similarity metric distance among classes to build a binary tree in which nodes correspond to different problem partitions \cite{iccv2011gao,marsalek,tree}. Finally, the ECOC framework consists of two steps: In the \textit{coding} step, a set of binary partitions of the original problem are encoded in a matrix of discrete codewords \cite{C6} (univocally defined, one code per class) (see Figure \ref{fig:intro}). At the \textit{decoding} step a final decision is obtained by comparing the test codeword resulting of the union of the binary classifier responses with every class codeword and choosing the class codeword at minimum distance \cite{decoding,Zhou2012}. The coding step has been widely studied in literature, yielding three different types of codings: predefined codings \cite{rifkin,hastie}, random codings \cite{random} and problem-dependent codings for ECOC \cite{subclass,decoc,Bautista2013PR,spectralECOC,icmlgao,zhao2013sparse}. Predefined codings like One vs. All or One vs. One are directly embeddable in the ECOC framework. In \cite{random}, the authors propose the Dense and Sparse Random coding designs with a fixed code length of $\{10,15\}\log_2(K)$, respectively. In \cite{random} the authors encourage to generate a set of $10^4$ random matrices and select the one that maximizes the minimum distance between rows, thus showing the highest correction capability. However, the selection of a suitable code length $l$ still remains an open problem.

\subsection{Problem-dependent Strategies}

Alternatively, problem-dependent strategies for ECOC have proven to be successful in multi-class classification tasks \cite{spectralECOC,iccv2011gao,icmlgao,zhao2013sparse,subclass,jointECOC,adaptiveECOC,decoc}. A common trend of these works is to exploit information of the multi-class data distribution obtained a priori in order to design a decomposition into binary problems that are easily separable. In that sense, \cite{spectralECOC} computes a spectral decomposition of the graph laplacian associated to the multi-class problem. The expected most separable partitions  correspond to the thresholded eigenvectors of the laplacian. However, this approach does not provide any warranties on defining unequivocal codewords (which is a core property of the ECOC coding framework) or obtaining a suitable code length $l$. In \cite{icmlgao}, Gao and Koller propose a method which adaptively learns an ECOC coding by optimizing a novel multi-class hinge loss function sequentially. On an update of their earlier work, Gao and Koller propose in \cite{iccv2011gao} a joint optimization process to learn a hierarchy of classifiers in which each node corresponds to a binary sub-problem that is optimized to find easily separable subproblems. Nonetheless, although the hierarchical configuration speeds up the testing step, it is highly prone to error propagation since node mis-classifications can not be recovered. Finally, the work of Zhao et. al \cite{zhao2013sparse} proposes a dual projected gradient method embedded on a constrained concave-convex procedure to optimize an objective composed of a measure of expected problem separability, codeword correlation and regularization terms. In the light of these results, a general trend of recent works is to optimize a measure of binary problem separability in order to induce easily separable sub-problems. This assumption leads to ECOC coding matrices that boost the boundaries of easily separable classes while modeling with low redundancy the ones with most confusion.

\subsection{Our approach}

In this paper we present the Error-Correcting Factorization (ECF) method for factorizing a design matrix of desired 'error-correcting properties' between classes into a discrete ECOC matrix. The proposed ECF method is a general framework for the ECOC coding step since the design matrix is a flexible tool for error-correction analysis. In this sense, the problem of designing the ECOC matrix is reduced to defining the design matrix, where higher level reasoning may be used. For example, following recent state-of-the-art works one could build a design matrix following a \textit{"hard classes are left behind"} spirit, boosting the boundaries of easily separable classes and disregarding the classes that are not easily separable. An alternative for building the design matrix is the \textit{"no class is left behind"} criteria, where we may boost those classes that are prone to be confused in the hope of recovering more errors. Note that the design matrix could also directly encode knowledge of domain experts on the problem, providing a great flexibility on the design of the ECOC coding matrix. Figure \ref{fig:intro} shows different coding schemes and the real boundaries learned by binary classifiers (SVM with RBF kernel) for a Toy problem of $14$ classes (see section \ref{sec:experiments} for further details on the dataset). We can see how the binary problems induced by ECF in Fig. \ref{fig:intro}(a) boost the boundaries of classes that are prone to be confused, while other approaches that use equal or higher number of classifiers like Dense Random \cite{random} in Fig. \ref{fig:intro}(b), or classic One vs. All designs in Fig. \ref{fig:intro}(c) fail in this task. 
The paper is organized as follows: Section \ref{sec:method} introduces the ECOC properties and derives ECF, where we cast the problem of finding an ECOC matrix that follows a certain distribution of correction as a discrete optimization problem. Section \ref{sec:discussion} presents a discussion of the method addressing important issues from the point of view of the ECOC framework. Concretely, we derive the optimal problem-dependent code length for ECOCs obtained by means of ECF, which to the best of our knowledge is the first time this question is tackled in the extended ECOC literature. In addition, we show how ECF converges to a solution with negligible objective value when the design matrix follows certain constraints. Section \ref{sec:experiments} shows how ECF yields ECOC coding matrices that obtain higher classification performances than state-of-the-art methods with comparable or lower computational complexity. Finally, Section \ref{sec:conclusions} concludes the paper.

\section{Methodology} \label{sec:method}

In this section, we review existing properties of the ECOC framework and propose to cast the ECOC coding matrix optimization as a Matrix Factorization problem that can be solved efficiently using a constrained coordinate descent approach.

\subsection{Error-Correcting Output Codes}

ECOC is a multi-class framework inspired on the basis of error-correcting principles of communication theory~\cite{C6}, which is composed of two different steps: \textit{coding} \cite{C6,random} and \textit{decoding} \cite{decoding,Zhou2012}. At the coding step an ECOC coding matrix $\X \in \{-1,+1\}^{{k \times l}}$ (see notation\footnote{Bold capital letters denote matrices (e.g. $\X$), bold lower-case letters represent vectors (e.g., $\x$). All non-bold letters denote scalar variables. $\x^i$ is the $i-$th row of the matrix $\textbf{X}$. $\x_j$ is the $j-$th column of the matrix $\X$. $\1$ is a matrix or vector of all ones of the appropriate size. $x_{ij}$ denotes the scalar in the $i-$th row and $j-$th column of $\X$. $\|\X\|_F=\Tr(\X^\top\X)$ denotes the Frobenius norm. $\|\cdot\|_p$ is used to denote the Lp-norm. $\x \oplus y$ is an operator which concatenates vectors $\x$ and $\textbf{y}$ . $\rank(\X)$ denotes the rank of $\X$. $\X \leq 0$ denotes the point-wise inequality}) is constructed, where $k$ denotes the number of classes in the problem and $l$ the number of bi-partitions (also known as dichotomies) to be learnt. In the coding matrix, the rows ($\x^i$'s, also known as \textit{codewords}) are unequivocally defined, since these are the identifiers of each category in the multi-class problem. On the other hand, the columns of $\X$ ($\x_j$'s) denote the bi-partitions to be learnt by base classifiers (also known as dichotomizer).  Therefore, for a certain column a dichotomizer learns the boundary between classes valued $+1$ and classes valued $-1$. However, \cite{random} introduced a third value, defining ternary valued coding matrices. $\X \in \{-1,+1,0\}^{{k \times l}}$. In this case, for any given dichotomy categories can be valued as $+1$ or $-1$ depending on the meta-class they belong to, or $0$ if they are ignored by the dichotomizer. This new value allows the inclusion of well-known decomposition techniques into the ECOC framework, such as One vs. One \cite{hastie}.

At the decoding step a data sample $\s$ is classified among the $\{c_1, \dots, c_k\}$ possible categories. In order to perform the classification task, each dichotomizer predicts a binary value for $\s$ whether it belongs to one of the bi-partitions defined by the correspondent dichotomy. Once the set of predictions $\y \in \{-1,+1\}^l$ is obtained, it is compared to the rows of $\X$ using a distance function $\delta$, known as the \textit{decoding function}. Usual decoding techniques are based on well-known distance measures such as the $l_1$ or Euclidean distance. These measures are proved to be effective for $\X \in \{+1,-1\}^{k \times l}$. Nevertheless, it is not until the work of \cite{decoding} that decoding functions took into account the meaning of the $0$ value at the decoding step. Generally, the final prediction for $\s$ is given by the class $c_i$, where $\underset{i}\arg\min \ \delta(\x^i,\y)$, $i \in \{1, \dots, k\}$.

\subsection{Good practices in ECOC}

Several works have studied the characteristics of a good ECOC coding matrix \cite{C6,kuncheva2003measures,Allwein00reducingmulticlass,spectralECOC,Bautista2013PR}, which are summed up in the following three properties:

\begin{enumerate}

\item \textbf{Correction capability:} let $\H \in \mathbb{R}^{k \times k}$ denote a symmetric matrix of hamming distances among all pairs of rows in $\X$, the correction capability is expressed as $ \lfloor \frac{\min(\H)-1}{2} \rfloor$ \footnote{In the case of ternary codes this correction capability can be easily adapted.}, considering only off-diagonal values of $\H$. In this sense, if $\min(\H)=3$, ECOC will be able to recover the correct multi-class prediction even if $\lfloor \frac{3-1}{2} \rfloor = 1$ binary classifier misses its prediction.\footnote{Note that for $\X$ to be valid all off-diagonal elements of $\H$ should be greater or equal than one.} 

\item \textbf{Uncorrelated binary sub-problems}: the induced binary problems should be as uncorrelated as possible for $\X$ to recover binary classifier errors.

\item \textbf{Use of powerful binary classifiers:} since the final class prediction consists of the aggregation of bit predictors, accurate binary classifiers are also required to obtain accurate multi-class predictions. 
\end{enumerate}

\subsection{From global to pair-wise correction capability}

\begin{figure}[h]
\includegraphics[width=0.48\textwidth]{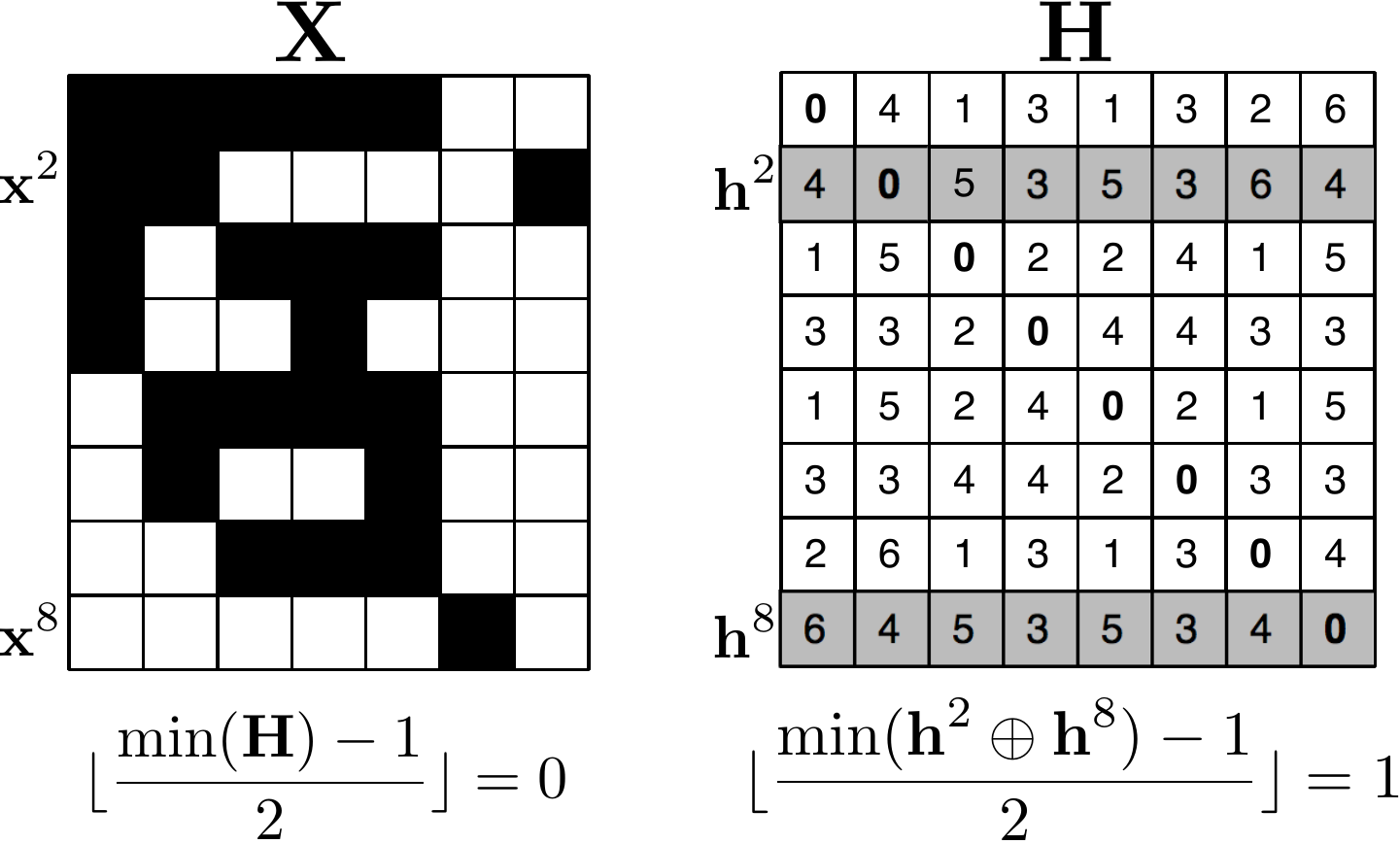}
\vspace{-0.4 cm}
\caption{Example of global versus pair-wise correction capability. On the left side of the figure the calculation of the global correction capability is shown. The right side of the image shows a sample of pair-wise correction calculation for codewords $\x^2$ and $\x^8$.}
\label{fig:correction}
\end{figure}

In literature, correction capability has been a core objective of problem-dependent designs of $\X$. In this sense, different authors have always agreed on defining correction capability for an ECOC coding matrix as a global value  \cite{C6,random,kuncheva2003measures,spectralECOC,iccv2011gao,pedrajas}. Hence, $\min(\H)$ is expected to be large in order for $\X$ to recover from as many binary classifier errors as possible. However, since $\H$ expresses the hamming distance between rows of $\X$, one can alternatively express the correction capability in a pair-wise fashion \cite{separability}, allowing for a deeper understanding of how correction is distributed among codewords. Figure \ref{fig:correction} shows an example of global and pair-wise correction capabilities calculation. Recall that the $\oplus$ operator between two vectors denotes its concatenation. Thus, the pair-wise correction capability is defined as follows:

\begin{enumerate}
\setcounter{enumi}{3}
\item The \textbf{pair-wise correction capability} of codewords $\x^i$ and $\x^j$ is expressed as: $\lfloor \frac{\min(\h^i \oplus \h^j)-1}{2} \rfloor$, where we only consider off-diagonal values of $\H$. This means that a sample of class $c_i$ is correctly discriminated from class $c_j$ even if $\lfloor \frac{\min(\h^i \oplus \h^j))-1}{2} \rfloor$ binary classifiers miss their predictions.
\end{enumerate}

Note that though in Figure \ref{fig:correction} the global correction capability of $\X$ is $0$, there are pairs of codewords with a higher correction, e.g. $\x^2$ and $\x^8$. In this case the global correction capability as defined in literature is overlooking ECOC coding characteristics that can potentially be exploited. This novel way of expressing the correction capability of an ECOC matrix enables a better understanding of how ECOC coding matrices distribute their correction capability, and gives an insight on how to design coding matrices. In this sense, it is straightforward to demand the correction capabilities of the ECOC matrix to be allocated according to those classes that are more prone to error, in order for them to have better recovery behavior (i.e. following a \textit{"no class is left behind"} criteria). However, recent works \cite{spectralECOC,iccv2011gao,zhao2013sparse} have focused on designing a matrix $\X$ where binary problems are easily separable. This assumption leads to a matrix $\X$ where classes that are not easily separable show a small hamming distance on their respective codewords (i.e. following a \textit{"hard classes are left behind"} scheme).

In addition to the proposal of a general method for ECOC coding by means of the definition of a design matrix, we explore the effect of focusing the learning effort of our method in those classes that have complex boundaries (i.e. those which show a small inter-class margin). It is important to take into account that though it is natural to estimate the design matrix from training data, it is not a limitation of ECF. In this sense, the design matrix can also code information of experts or any other distance measure directly set by the user. Formally, let $\X \in \{-1,+1\}^{k \times l}$ be a coding matrix, let $\H$ be a symmetric matrix of pair-wise $l_1$ distances between rows of $\X$ and let $\D \in \mathbb{R}^{k \times k}$ be a design matrix (e.g. pair-wise distance measure between class codewords). It is natural to see that the ordinal properties of the distance should hold in $\H$ and $\D$. Thus, if distance between codewords $\x^k$ and $\x^l$ ($d_{kl}$) is required to be larger than the distance between codewords $\x^i$ and $\x^j$ ($d_{ij}$), this order should be maintained in $\H$. Then we want to find a configuration of $\X$ such that $h_{ij}<h_{kl} \iff  d_{ij}<d_{kl} \forall_{i,j,k,l}$.

Note that the $l_1$ distances in $\H$ can be seen as a function of the dot product of the codewords $\|\x^i - \x^j \|_1 = \frac{ -({\x^i}{\x^j}^\top)+l}{2}$, where $ \x \in \{-1,+1\}$ . Therefore, instead of directly requiring $\H$ to match $\D$, we can equivalently require the product $\X\X^\top$ to match $\D$ \cite{weiss2012multidimensional}. This implies that we can cast the problem of finding $\X$ into a Matrix Factorization problem, where we find an $\X$ so that the matrix of inner products $\X \X^\top$ is closest to $\D$ under a given norm.

\subsection{Error-Correcting Factorization}

This section describes the objective function and the optimization strategy for the ECF algorithm.
\subsubsection{Objective}
Our goal is to find an ECOC coding matrix that encodes the properties denoted by the design matrix $\D$. In this sense, ECF seeks a factorization of the design matrix  $\D \in \mathbb{R}^{k \times k}$ into a discrete ECOC matrix $\X$. This factorization is formulated as the quadratic form $\X \X^\top$ that reconstructs $\D$ with minimal Frobenius distance under several constraints, as shown in Equation \eqref{eq:minproblem} \footnote{Recall that the $l_1$ distance is a function of the dot product $\|\x^i - \x^j \|_1 = \frac{-({\x^i}{\x^j}^\top)+l}{2}$.}.

\begin{eqnarray}
\underset{\X}{\text{minimize}} && \| \D - \X\X^\top  \|_{F}^{2}\label{eq:minproblem}  \\
\text{subject to} && \X \in \{-1,+1\}^{k \times l}  \label{constraint:disc}\\
 && \X\X^\top - \P \leq 0 \label{constraint:rows} \\
 && \X^\top\X - \1(l-1) \leq 0  \label{constraint:cols1} \\
 && -\X^\top\X - \1(l-1) \leq 0  \label{constraint:cols2} 
\end{eqnarray}

The component $\X^* \in \{-1,+1\}^{k \times l}$ that solves this optimization problem generates the inner product of discrete vectors that is closest to $\D$ under the Frobenius norm. In order for $\X$ to be a valid matrix under the ECOC framework we constraint $\X$ in Equations \eqref{constraint:disc}-\eqref{constraint:cols2}. Equation \eqref{constraint:disc} ensures that each binary problem classes will belong to one of the two possible meta-classes. In addition, to avoid the case of having two or more equivalent rows in $\X$, the constraints in \ref{constraint:rows} ensure that the correlation between rows of $\X$ less or equal than a certain user-defined matrix $-\1l \leq \P \leq \1l$ (recall that $\1$ denotes a matrix or vector of all 1s of the appropriate size when used), where $\P$ encodes the minimum distance between any pair of codewords. $\P$ is a symmetric matrix with $p_{ii}=l \ \forall i$. Thus, by setting the off diagonal values in $\P$ we can control the minimum inter-class correction capability. Hence, if we want the correction capability of rows $\x^i$ and $\x^j$ to be $\lfloor\frac{c-1}{2}\rfloor$, we set $\p^i=\p^j=\1 (l-c)$.

Finally, constraints in Equations \eqref{constraint:cols1} and \eqref{constraint:cols2} ensure the induced binary problems are not equivalent. Similar constraints have been studied thoroughly in literature \cite{C6,kuncheva2003measures,pedrajas} defining methods that rely on diversity measures for binary problems to obtain a coding matrix $\X$. Equations \eqref{constraint:cols1} and \eqref{constraint:cols2} can be considered as soft-constraints since its violation does not imply violating the ECOC properties in terms of row distance. This is easy to show since a coding matrix $\X \in \{-1,+1\}^{k \times l}$ that induces some equivalent binary problems but ensures that $\X \X^\top \leq \1(l-1),\ \forall i,j : i\neq j $ will define a matrix whose rows are unequivocally defined. In this sense, a coding matrix $\X$ can be easily projected on the set defined by constraints \eqref{constraint:cols1} and \eqref{constraint:cols2} by eliminating repeated columns, $\X = \x_j : \x_j \neq \x_i \forall j \neq i$. Thus, constraints in \ref{constraint:cols1} and \ref{constraint:cols2} ensure that uncorrelated binary sub-problems will be defined in our coding matrix $\X$. The discrete constraint in Equation \ref{constraint:disc} on the variable elevates the optimization problem to the NP-Hard class. To overcome this issue and following \cite{Crammer2,zhao2013sparse,boyd} we relax the discrete constraint in \ref{constraint:disc} an replace it by $\X \in [-1,+1]^{k \times l}$ in Equation \ref{constraint:relaxed}.

\subsubsection{Optimization}

In this section, we detail the process for optimizing $\X$. The minimization problem posed in Equation \eqref{eq:minproblem} with the relaxation of the boolean constraint in Equation \eqref{constraint:disc} is non-convex, thus, $\X^*$ is not guaranteed to be a global minimum. In this sense, although gradient descent techniques have been successfully applied in the literature to obtain local minimums \cite{bmds,mds1,agarwal2007generalized} these techniques do not enjoy the efficiency and scalability properties present in other optimization methods applied to Matrix Factorization problems, such as Coordinate Descent \cite{lin2007projected,De12}. Coordinate Descent techniques have been widely applied in Nonnegative Matrix Factorization obtaining satisfying results in terms of efficiency \cite{nnmf1,nnmf2}. In addition, it has been proved that if each of the coordinate sub-problems can be solved exactly, Coordinate Descent converges to a stationary point \cite{conv,tseng2001convergence}. Using this result, we decouple the problem in Equation \eqref{eq:minproblem} into a set of linear least-squares problems (one for each coordinate). Therefore, if the problem in Equation \eqref{eq:minproblem} is going to be minimized along the $i-$th coordinate of $\X$, we fix all rows of $\X$ except of $\x^i$ and we substitute $\X$ with $\begin{bmatrix} \x^i \\ \X^{\prime i} \end{bmatrix}$ in Equations \eqref{eq:minproblem} and \eqref{constraint:rows}, where $\X^{\prime i}$ denotes matrix $\X$ after removing the $i-$th row. In addition, we substitute ${\D}$ with $\begin{bmatrix} l & \d_i \\ {\d^{i}}^{T} & \D^{\prime i}_{\prime i} \end{bmatrix}$, where $\D^{\prime i}_{\prime i}$ denotes the matrix $\D$ after removing the $i-$th row and column. Equivalently, we substitute $\P = \begin{bmatrix} l & \p_i \\ {\p^{i}}^{T} & \P^{\prime i}_{\prime i} \end{bmatrix}$, obtaining the following block decomposition:

\vspace{-0.2 cm}
\begin{eqnarray}
\underset{\x^i}{\text{minimize}} && \left\|  \begin{bmatrix} l & \d_i \\ {\d^{i}}^{T} & \D^{\prime i}_{\prime i} \end{bmatrix}  -   \begin{bmatrix} \x^i {\x^{i}}^{T} & \X^{\prime i} \x_i \\ \X^{\prime i} {\x^{i}}^{T} & \X^{\prime i} {\X^{\prime i}}^\top \end{bmatrix}  \right\|_{F}^{2} \label{eq:minproblem2} \\
\text{subject to} && \x^i \in [-1,+1]^{l}  \label{constraint:relaxed}\\
 &&  \begin{bmatrix} \x^i {\x^{i}}^{T} & \X^{\prime i} \x_i \\ \X^{\prime i} {\x^{i}}^{T} & \X^{\prime i} {\X^{\prime i}}^\top \end{bmatrix} - \begin{bmatrix} l & \p_i \\ {\p^{i}}^{T} & \P^{\prime i}_{\prime i} \end{bmatrix} \leq 0. \label{constraint:rows2}
\end{eqnarray}

 Analyzing the block decomposition in Equation \eqref{eq:minproblem2} we can see that the only terms involving free variables are $\x^i{\x^i}^\top$, $\X^{\prime i} \x^i$ and ${\X^{\prime i} \x^i}^\top$. Thus, since $\D$ and $\X \X^\top$ are symmetric by definition, the minimizer ${\x^i}^*$ of Equation \eqref{eq:minproblem2} is the solution to the linear least-squares problem shown in Equation \eqref{eq:minproblem3}:

\vspace{-0.2 cm}
\begin{eqnarray}
\underset{\x^i}{\text{minimize}} && \left\|   \X^{\prime i} \x^i - \d^i  \right\|_{2}^{2} \label{eq:minproblem3} \\
\text{subject to} && -1 \leq \x^i \leq +1  \label{constraint:relaxed3}\\
 &&   \X^{\prime i} \x^i - \p^i \leq 0, \label{constraint:rows3}
\end{eqnarray}

where constraint \eqref{constraint:relaxed3} is the relaxation of the  discrete constraint \eqref{constraint:disc}. In addition, constraint \eqref{constraint:rows3} ensures the correlation of $\x^i$ with the rest of the rows of $\X$ is below a certain value $\p^i$. Algorithm \ref{alg:CD} shows the complete optimization process.

\begin{algorithm}
\KwData{$\tilde{\D} \in \mathbb{R}^{k \times k}, \P \in \mathbb{N}^{k \times k}, l$}
\KwResult{$\X \in \{-1,+1\}^{k \times l}$}
\Begin{
\Repeat{convergence}{
	\ForEach{$i \in \{1, 2, \dots, k\}$}{ 
	$\x^i  \leftarrow  \underset{\x^i \in \mathbb{R}^l}{\text{minimize}} \left\|  \X^{\prime i} \x^i - \d^i  \right\|_{2}^{2}, \ \text{subject to}:  -1 \leq \x^i \leq +1,  \  \X^{\prime i} \x^i - \p^i \leq 0$\;
	}
$\X \leftarrow \epsilon\text{-suboptimal}(\X)$\;
$\X = \{ \x_j : \x_j \neq \x_i \forall j \neq i\}$\tcp*[l]{Projection step to remove duplicate columns}
}}
\caption{Error-Correcting Factorization Algorithm. \label{alg:CD}}
\end{algorithm}

To solve the minimization problem in Algorithm \ref{alg:CD} we use the Active Set method described in \cite{gill2007numerical}, which finds an initial feasible solution by first solving a linear programming problem. Once ECF converges to a solution $\X^*$  with objective value $f_{obj}(\X^*)$ we obtain a discretized $\epsilon$-suboptimal solution $\X \in \{-1,+1\}$ with objective value $f_{obj}(\X)$ by sampling 1000 points that split the interval $[-1,+1]$ and choosing the point that minimizes $\|f_{obj}(\X^*)-f_{obj}(\X) \|_2$. Finally, we discard repeated columns if any appear \footnote{In all our runs of ECF this situation happened with a chance of less than $10^{-5}\%$.}.

\subsection{Connections to Singular Value Decomposition, Nearest Correlation Matrix and Discrete Basis problems}

Similar objective functions to the one defined in the ECF problem in Equation \eqref{eq:minproblem} are found in other contexts, for example, in the Singular Value Decomposition problem (SVD). The SVD uses the same objective function as ECF subjected to the constraint $\X \X^\top = \I$. However, the solution of SVD yields an orthogonal basis, disagreeing with the objective defined in Equation \eqref{eq:minproblem} which ensures different correlations between the $\x^i$'s. In addition, we can also find a common ground with the Nearest Correlation Matrix (NMC) Problem \cite{ncm,boyd2005least,malick2004dual}. However, the NMC solution does not yield a discrete factor $\X$, instead it seeks directly for the Gramian $\X\X^\top$ where $\X$ is not discrete, as in Equation \eqref{eq:nmc}. 

\begin{eqnarray}
\underset{\X}{\text{minimize}} && \| \X - \D \|_{F}^{2} \label{eq:nmc}  \\
\text{subject to} && \X \succeq 0 \\
&& \bc \X \bc^\top = \b  
\end{eqnarray}

In addition, the ECF has similarities with the Discrete Basis Problem (DBP) \cite{dbp}, since the factors are $\X$ discrete valued. Nevertheless, DBP factorizes $\D \in \{0,1\}^{k \times k}$ instead of $\D \in \mathbb{R}^{k \times k}$, as show in Equation \eqref{eq:dbp}.

\begin{eqnarray}
\underset{\X,\Y}{\text{minimize}} && \| \X\circ\Y - \D \|_{1} \label{eq:dbp}  \\
\text{subject to} && \X,\Y,\D  \in \{0,1\} 
\end{eqnarray}

\section{Discussion} \label{sec:discussion}
In this section we discuss how to ensure that the design matrix $\D$ is valid, as well as how to automatically estimate the code length for each problem given $\D$. Furthermore, we analyze the convergence of ECF in relation to the order of updating the coordinates. Finally we show that under certain conditions of $\D$ ECF converges to a solution with almost negligible objective value. 

\subsection{Ensuring a representable design matrix}\label{sec:delta}

An alternative interpretation for ECF is that it seeks for a discrete matrix $\X$ whose Gramian is closest to $\D$ under the Frobenius norm. However, since $\D$ can be directly set by the user we need to guarantee that $\D$ is a correlation matrix that is realizable in the $\mathbb{R}^{k \times k}$ space, that is, $\D$ has to be symmetric and positive semi-definite. In particular, we would like to find the correlation matrix $\tilde{\D} \in\mathbb{R}^{k \times k}$ that is closest to $\D$ under the Frobenius norm. This problem has been treated in several works \cite{ncm,boyd2005least,cayton2006robust,globerson2007euclidean}, resulting in various algorithms that often use an alternating projections approach. However, for this particular case in addition to be in the Positive Semidefinite (PSD) Cone and symmetric we also require $\D$ to be scaled in the $[-l,+l]$ range, with $\tilde{\delta}_{ii}=l \forall i$. In this sense, to find $\tilde{\D}$  we follow an alternating projections algorithm, similar as \cite{ncm}, which is shown in Algorithm \ref{alg:alternproj}. We first project $\D$ into the PSD cone by computing its eigenvectors and recovering $\D= \V\diag(\blambda_+)\V^\top$, where $\blambda_+$ are the non-negative eigenvalues of $\D$. Then, we scale $\D$ in the range $[-l,+l]$ and set $\delta_{ii}=l \forall i$.

\begin{algorithm}
	\KwData{$\D \in \mathbb{R}^{k \times k}$}
	\KwResult{$\tilde{\D} \in \mathbb{R}^{k \times k}$ }
	\Begin{
		\Repeat{convergence}{
			$\D \leftarrow\V \diag(\blambda_+)\V^\top$\;
			$\D \leftarrow \D \in [-l,+l]^{k \times k}$\;
			$\D \leftarrow d_{ii}=l \forall i$\;
		}
	}
	\caption{Projecting $\D$ into the PSD cone with additional constraints.\label{alg:alternproj}}
\end{algorithm}

\subsection{Defining a code length with representation guarantees}

The definition of a problem-dependent ECOC code length $l$, that is, choosing the number of binary partitions for a given multi-class task is a problem that has been overlooked in literature. For example, predefined coding designs like One vs. All or One vs. One have fixed code length. On the other hand, coding designs like Dense or Sparse Random codings (which are very often used in experimental comparisons \cite{spectralECOC,zhao2013sparse,Bautista2013PR,subclass}) are suggested \cite{random} to have a code length of $\lceil 10log_2(k)\rceil$ and $\lceil 15log_2(k)\rceil$ respectively. These values are arbitrary and unjustified. Additionally, to build a Dense or Sparse Random ECOC matrix one has to generate a set of 1000 matrices and chose the one that maximizes $\min(\H)$. Consider the Dense Random Coding design, of length $l=\lceil 10\log_2(k)\rceil$, the ECOC matrix will have in the best case a correction capability of $\lfloor \frac{10-1}{2}\rfloor = 4$, independently of the distribution of the multi-class data. In addition, the effect of maximizing $\min(\H)$ leads to an equi-distribution of the correction capability over the  classes. Other approaches, like Spectral ECOC \cite{spectralECOC} search for the code length by looking at the best performance on a validation set. Nevertheless, recent works have shown that the code length can be reduced to of $l=log_2(k)$ with very small loss in performance if the ECOC coding design is carefully chosen \cite{LorenaMulticlass} and classifiers are strong. In this paper, instead of fixing the code length or optimizing it on a validation subset, we derive the optimal length according to matrix rank properties. Consider the rank of a factorization of $\D$ into $\X\X^\top$, there are three different possibilities:

\begin{enumerate}
	
\item If $\rank(\X\X^\top)=\rank(\D)$, we obtain rank factorization algorithm that should be able to factorize $\D$ with minimal error.

\item In the case when $\rank(\X\X^\top)<\rank(\D)$ we obtain a low-rank factorization method that cannot guarantee to represent $\D$ with 0 error, but reconstructs the components of $\D$ with higher information.

\item If $\rank(\X\X^\top)>\rank(\D)$, the system is overdetermined and many possible solutions exist.
	
\end{enumerate}

\begin{figure}
	\setlength{\tabcolsep}{0pt}
	\centering
	\begin{tabular}{c c}
	
		\includegraphics[width=0.25\textwidth]{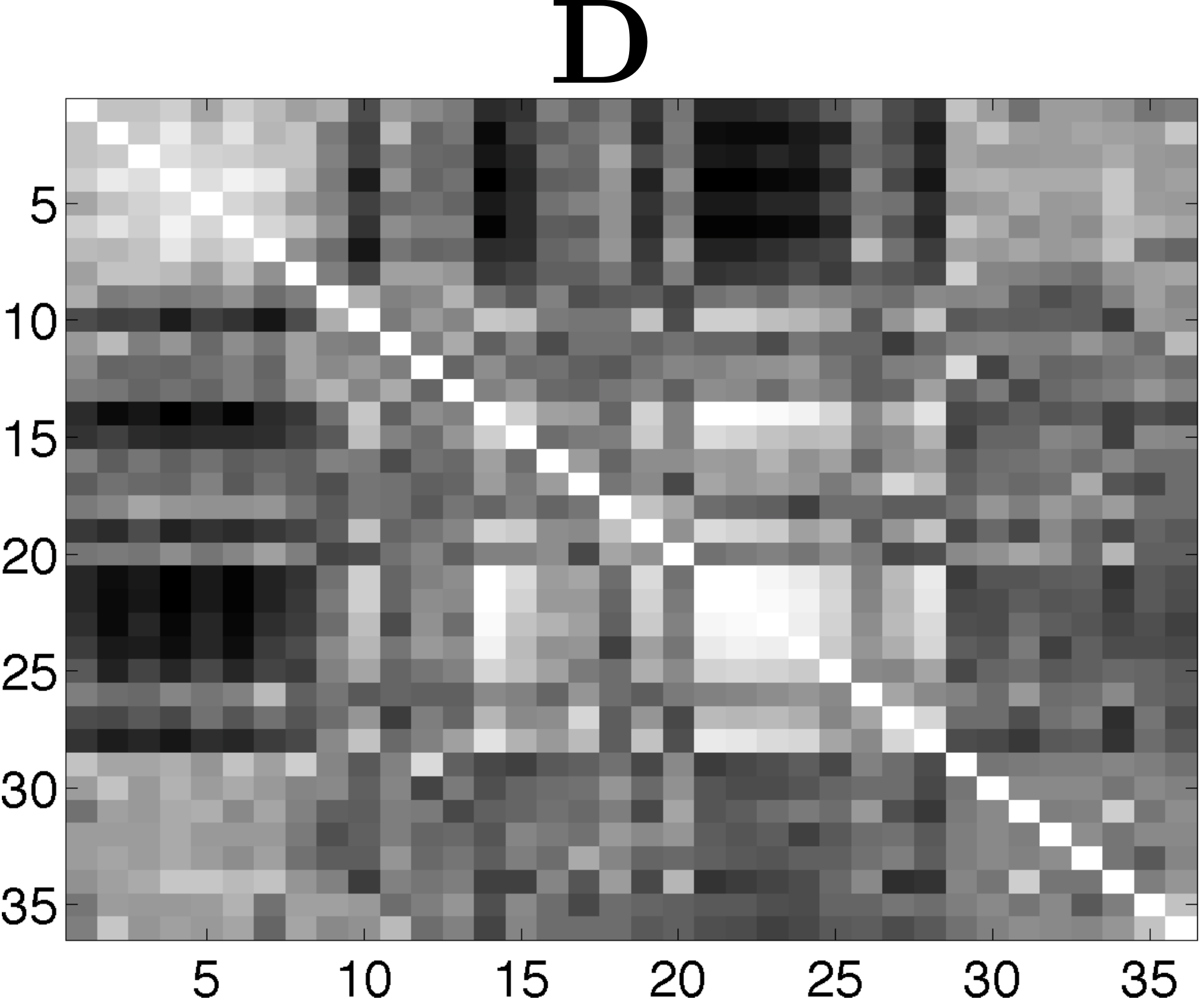} &\includegraphics[width=0.25\textwidth]{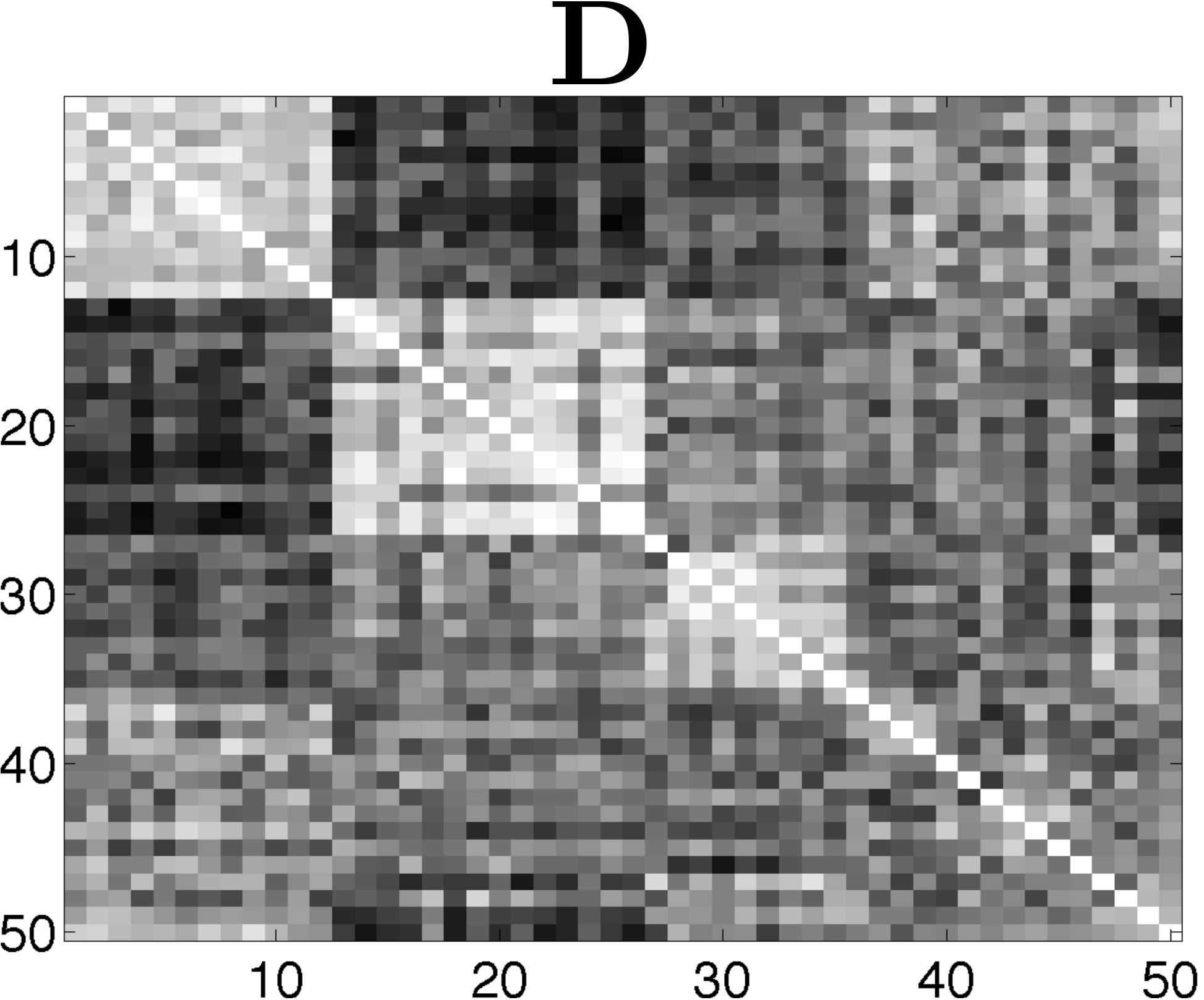}\\
		(a) &(b)\\
		\includegraphics[width=0.25\textwidth]{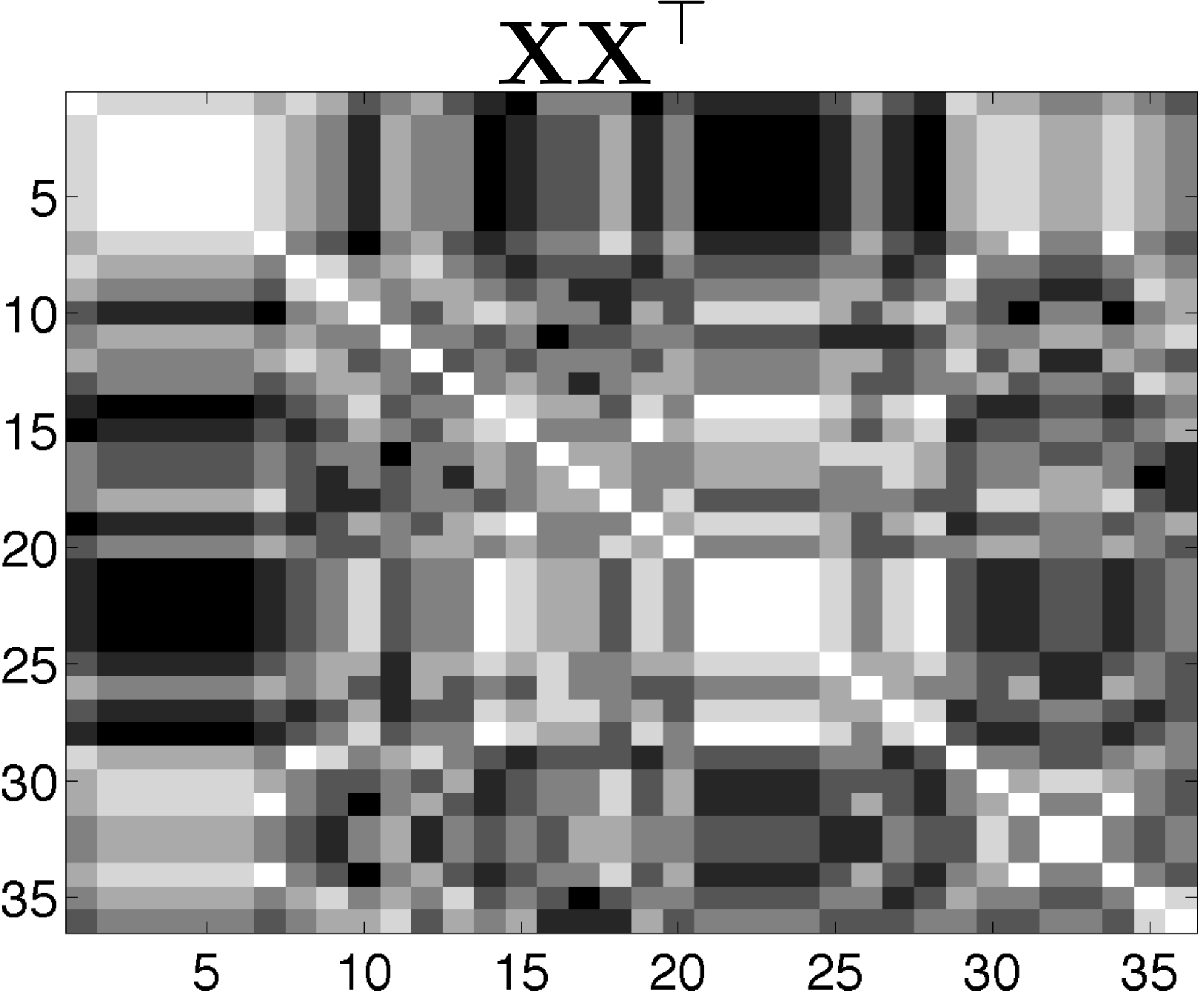}&\includegraphics[width=0.25\textwidth]{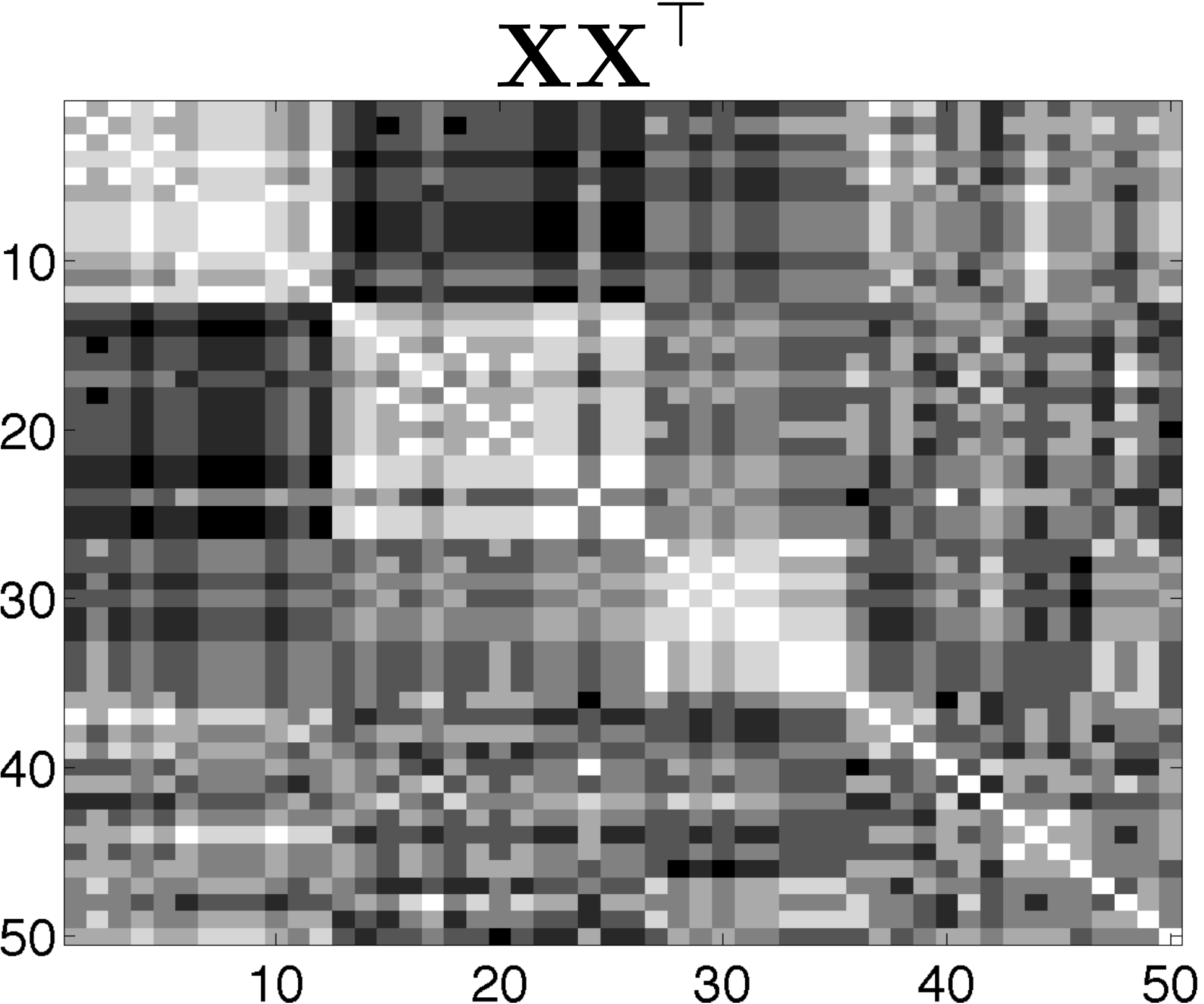} \\
		(c) & (d)\\
		\includegraphics[width=0.25\textwidth,height= 3.5 cm]{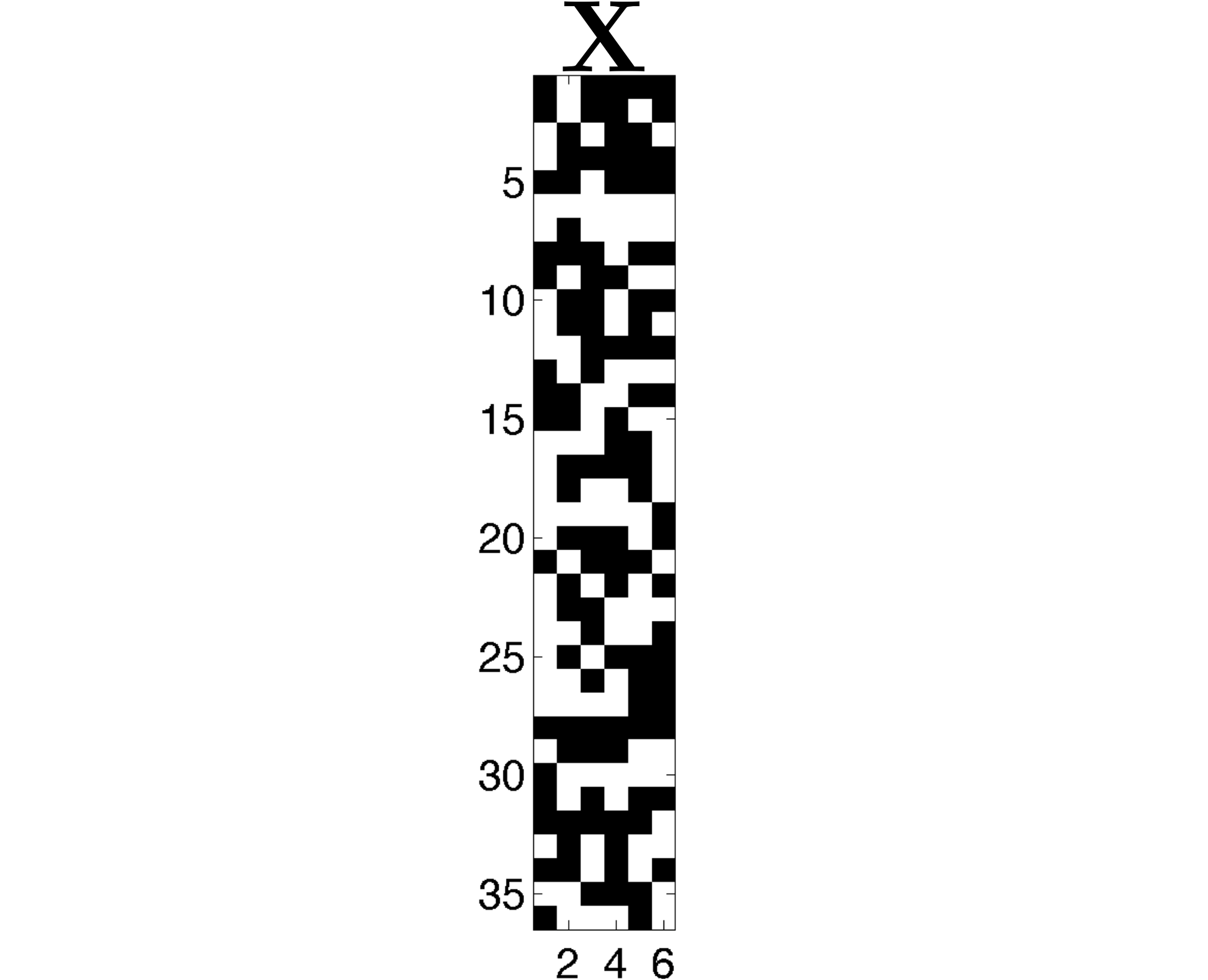} & \includegraphics[height=3.5 cm]{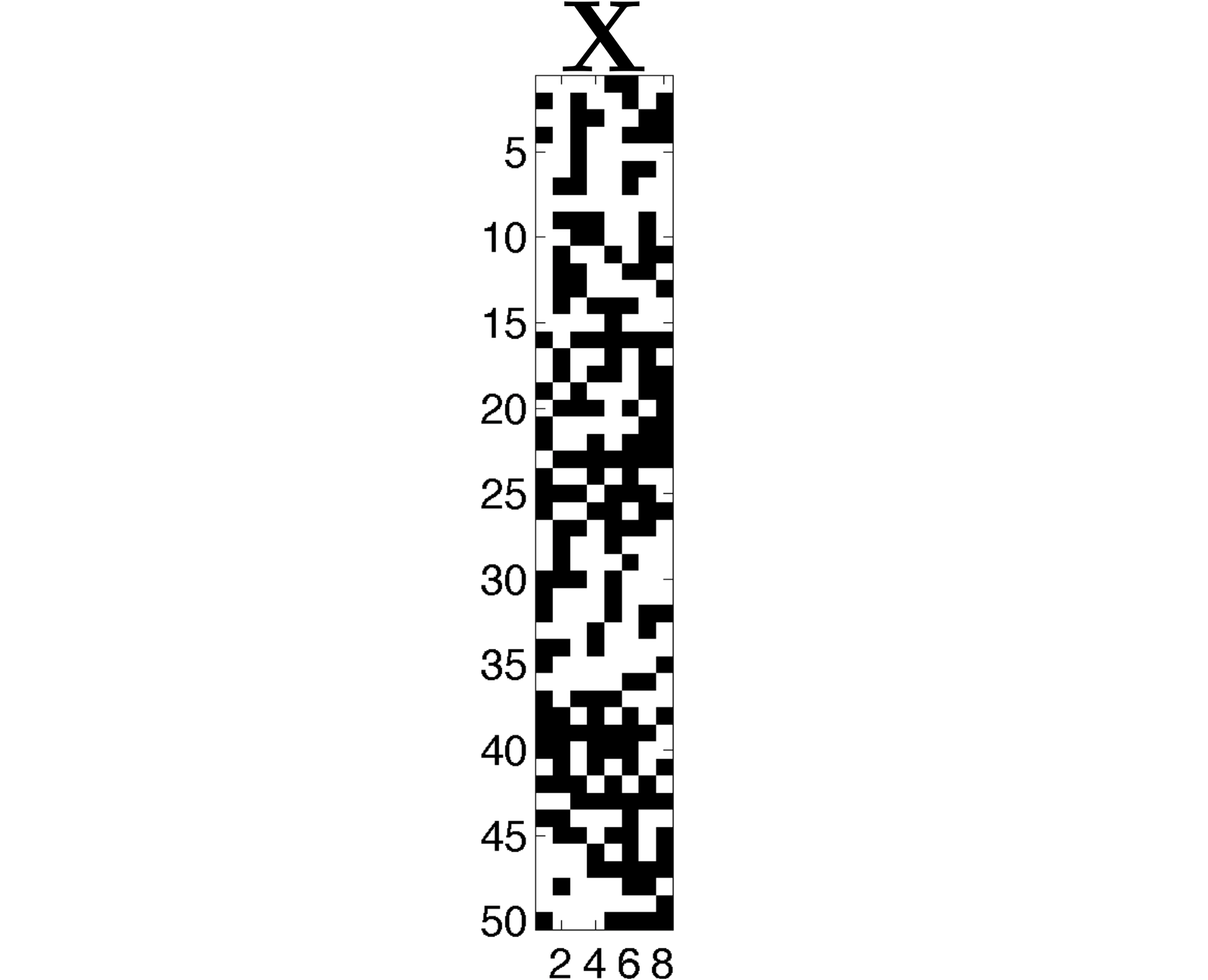}\\
		(e) &(f)\\ 
	\end{tabular}
	\vspace{-0.8 cm}
	\caption{$\D$ matrix for the \textit{Traffic} (a) and \textit{ARFace} (b) datasets. $\X\X^\top$ term obtained via ECF for \textit{Traffic} (c) and \textit{ARFace} (d) datasets. ECOC coding matrix $\X$ obtained with ECF for \textit{Traffic} (e) and \textit{ARFace} (f).}
	\label{fig:opt}
\end{figure}

In general we would like to reconstruct $\D$ with minimal error, and since $\rank(\X) \leq \min(k,l)$ and $k$ (the number of classes) is fixed, we only have to set the number of columns of $\X$ to control the rank. Hence, by setting $\rank(\X)=l=\rank(\D)$, ECF will be able to factorize $\D$ with minimal error. Figure \ref{fig:opt} shows visual results for the ECF method applied on the \textit{Traffic} and \textit{ARFace} datasets. Note how, for the \textit{Traffic} (36 classes) and \textit{ARFaces} (50 classes) datasets the required code length for ECF to full rank factorization is $l=6$ and $l=8$, respectively as shown in Figures \ref{fig:opt}(e)(f).

\subsection{Order of Coordinate Updates}

Coordinate Descent has been applied in a wide span of problems obtaining satisfying results. However, the problem of choosing the coordinate to minimize at each iteration still remains active \cite{richtarik2014iteration,friedman2010regularization,tseng2001convergence,hsieh2011fast}. In particular, \cite{nesterov2012efficiency} derives a convergence rate which is faster when coordinates are chosen uniformly at random rather than on a cyclic fashion. Hence, choosing coordinates at random its a suitable choice when the problem shows some of the following characteristics \cite{richtarik2014iteration}:

\begin{itemize}
	\item Not all data is available at all times.
	
	\item A randomized strategy is able to avoid worst-case order of coordinates, and hence might be preferable.
	
	\item Recent efforts suggest that randomization can improve the convergence rate \cite{nesterov2012efficiency}.
\end{itemize}

\begin{figure}[h]
	\centering
		\setlength{\tabcolsep}{0pt}
	\begin{tabular}{c c}
		\includegraphics[width=0.24\textwidth]{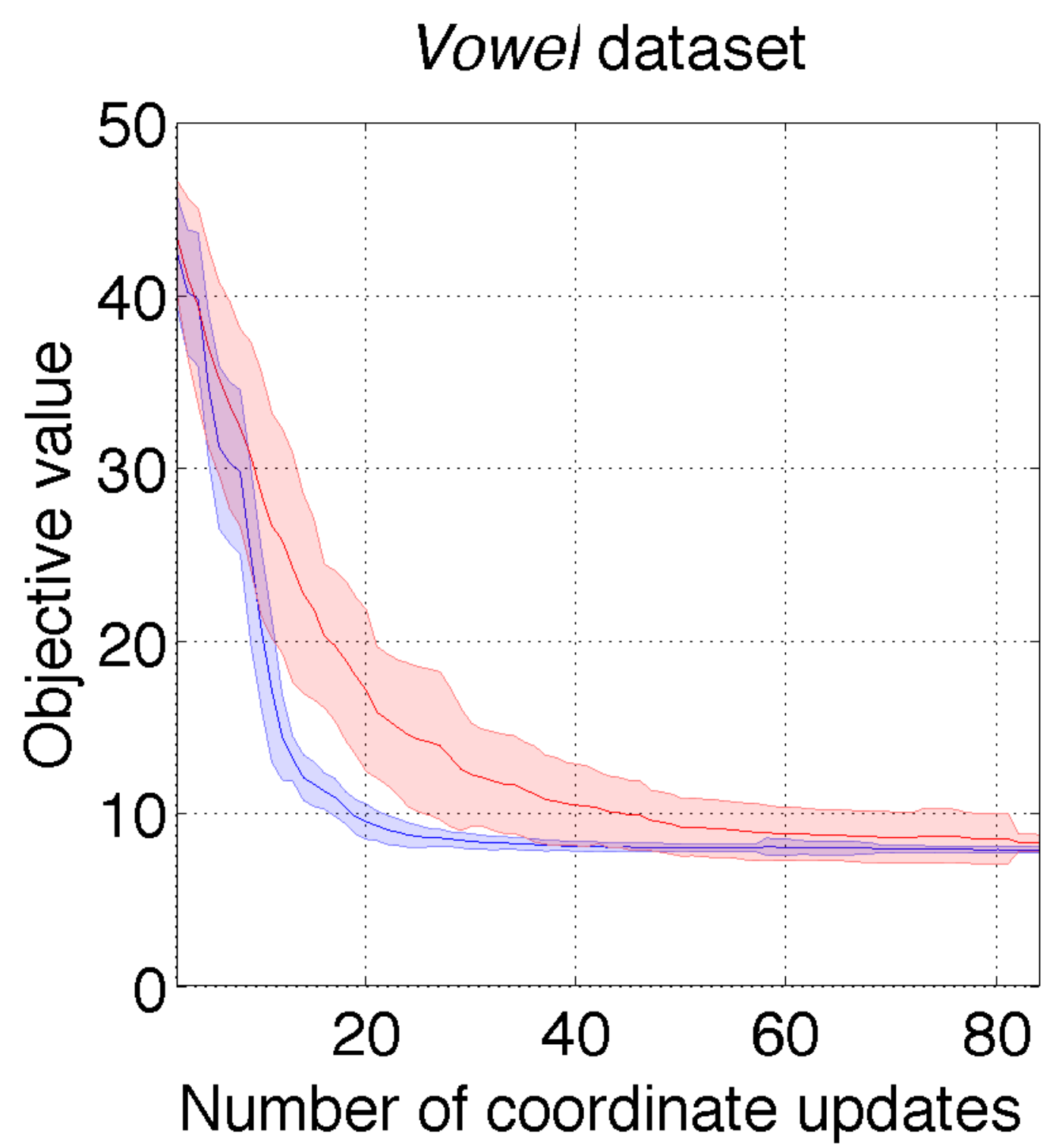} & \includegraphics[width=0.26\textwidth]{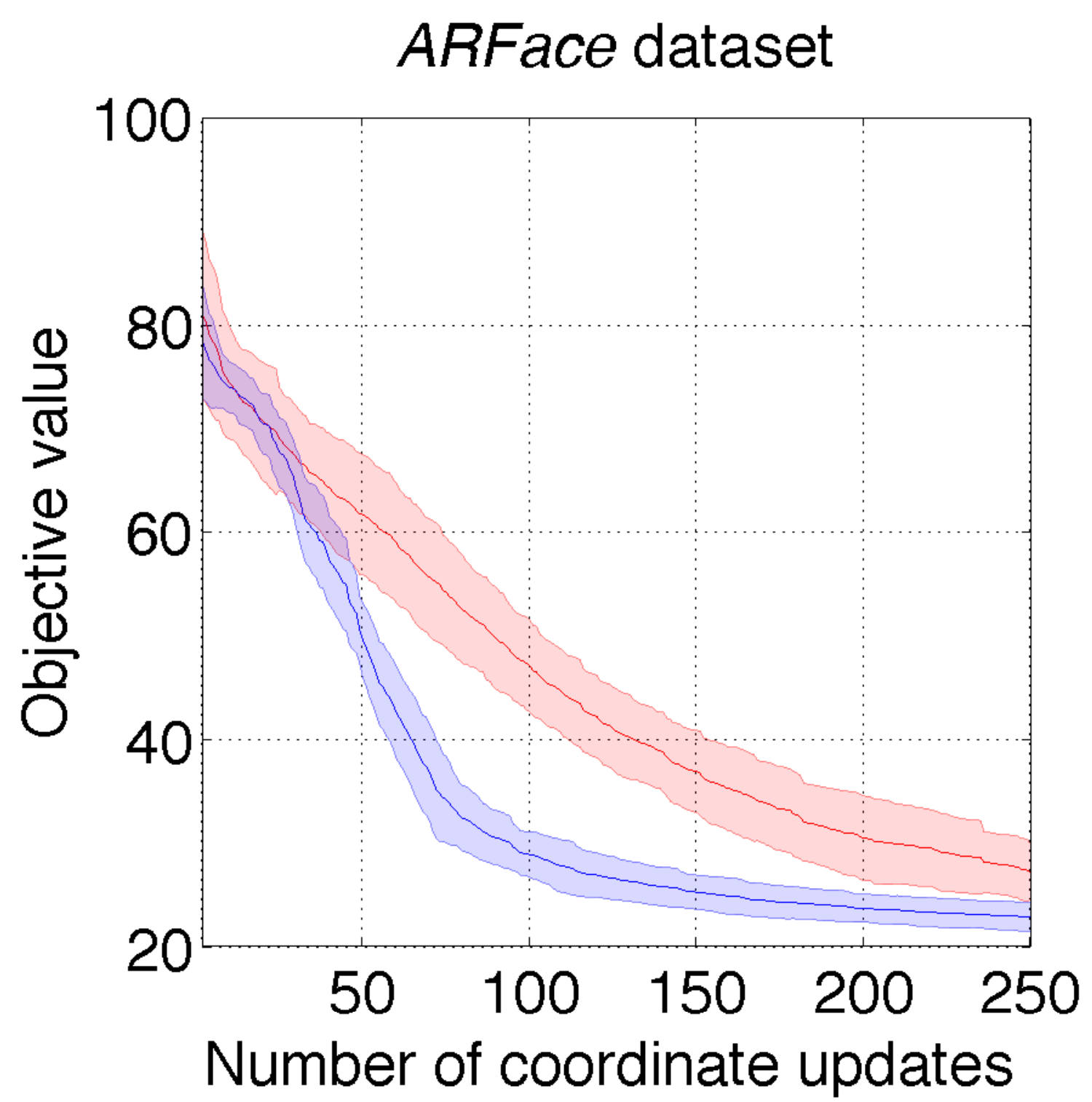}\\ 
		(a) &(b)
	\end{tabular}
	\vspace{-0.4 cm}
	\caption{Mean Frobenius norm value with standard deviation as a function of the number of coordinate updates on $50$ different trials. The blue shaded area corresponds to cyclic update while the red area denotes random coordinate updates for \textit{Vowel} (a) and \textit{ARFAce} (b) datasets.}
	\label{fig:conv}
\end{figure}

However, the structure of ECF is different and calls for a different analysis. In particular, we remark the following points. (i) At each coordinate update of ECF, information about the rest of coordinates is available. (ii) Since our coordinate updates are solved uniquely, a repetition on a coordinate update does not change the objective function.  (iii) The descent on the objective value when updating a coordinate is maximal when all other coordinates have been updated. These reasons leads us to choose a cyclic update scheme for ECF. In addition in Figure \ref{fig:conv} we show a couple of examples in which the cyclic order of coordinates converges faster than the random order for two problems: \textit{Vowel} and \textit{ARFace} (refer to Section \ref{sec:experiments} for further information on the datasets). This behavior is common for all datasets. In particular, note how the cyclic order of coordinates reduces the standard deviation on the objective function, which is denoted by the narrower blue shaded area in Figure \ref{fig:conv}. 

\subsection{Approximation Errors and Convergence results when $\D$ is an inner product of binary data}

The optimization problem posed by ECF in Equation \eqref{eq:minproblem} is non-convex due to the quadratic term $\X\X^\top$, even if the discrete constraint is relaxed. This implies that we cannot guarantee that the algorithm converges to the global optima. Recall that ECF seeks for the term $\X\X^\top$ that is closest to $\D$ under the Frobenius norm. Hence, the error in the approximation can be measured by $\|\X^*\X^{*\top} - \D\|_{F}^{2} \geq 0$, where $\X^*$ is the local optimal point to which ECF converges.  In this sense, we introduce $\D^B$ which is the matrix of inner products of discrete vectors that is closest to $\D$ under the Frobenious norm. Thus, we expand the norm as in the following equation:

\begin{eqnarray}
\|\X^*\X^{*\top} - \D\|_{F}^{2} = \|\X^*\X^{*\top} - \D^B + \D^B - \D  \|_{F}^{2} = \label{eq:initterm} \\
=  \|\X^*\X^{*\top} - \D^B \|_{F}^{2} +  \|\D - \D^B \|_{F}^{2}-  \label{eq:errorterms} \\
- 2 \Tr ((\X^*\X^{*\top} - \D^B)(\D -\D^B)). \label{eq:extraterm}
\label{eq:error}
\end{eqnarray}

\begin{itemize}
\item The optimization error $\varepsilon_o$: measured as the distance between the local optimum where ECF converges and $\D^B$ denoted by $\varepsilon_o = \|\X^*\X^{*\top} - \D^B\|_{F}^{2}$, which is expressed as the first term in Equation \eqref{eq:errorterms}.

\item The discretization error $\varepsilon_d$: computed as, $\varepsilon_d = \|\D-\D^B\|_{F}^{2}$, that is, the distance between $\D$ and the closest inner product of discrete vectors $\D^B$, expressed as the second term in Equation \eqref{eq:errorterms}.

\end{itemize}

\begin{figure}
	\centering
	\setlength{\tabcolsep}{0pt}
	\begin{tabular}{c c}
	\includegraphics[width=0.25\textwidth,height = 3.53 cm]{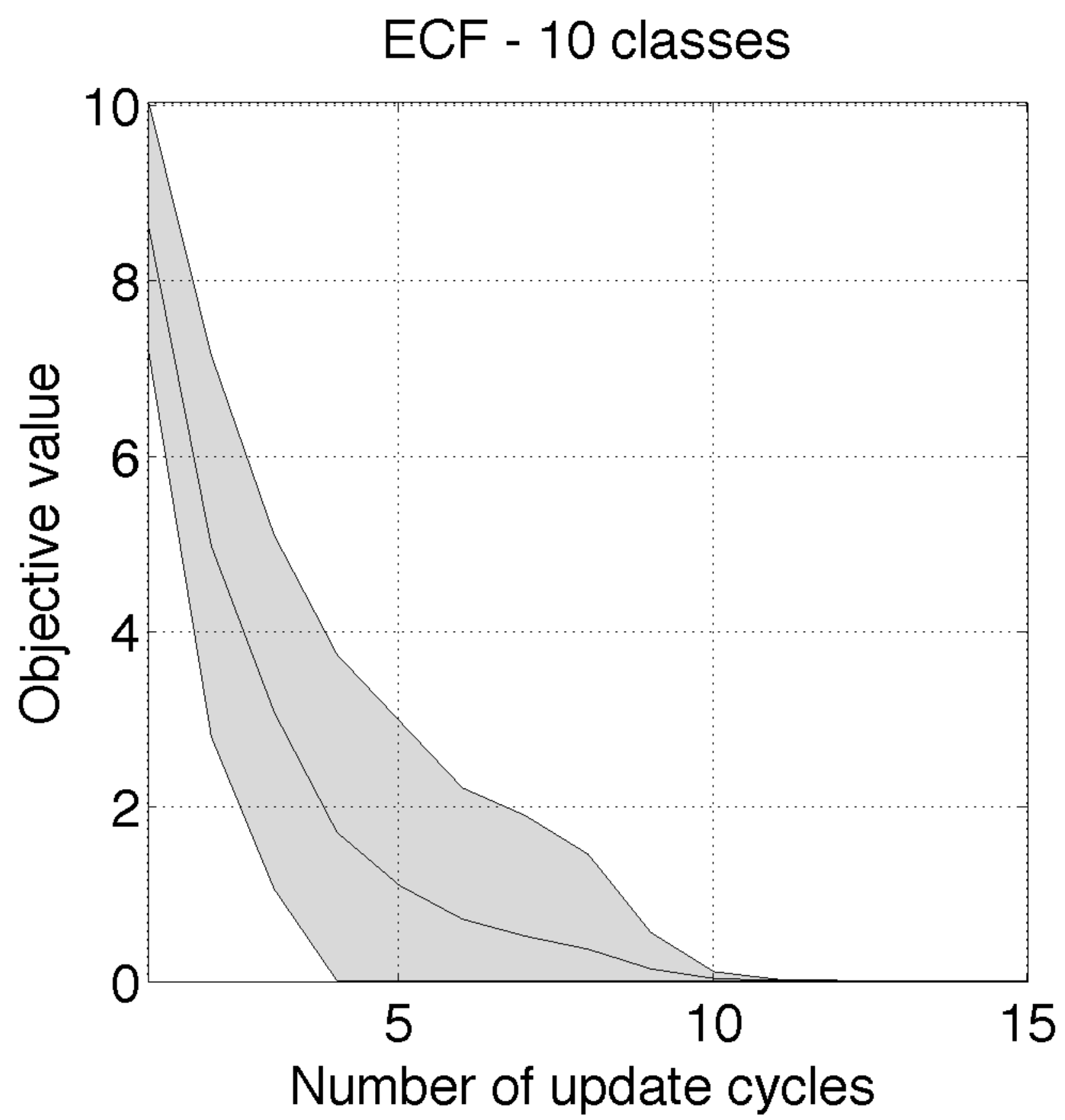}&
	\includegraphics[width=0.25\textwidth,height = 3.5 cm]{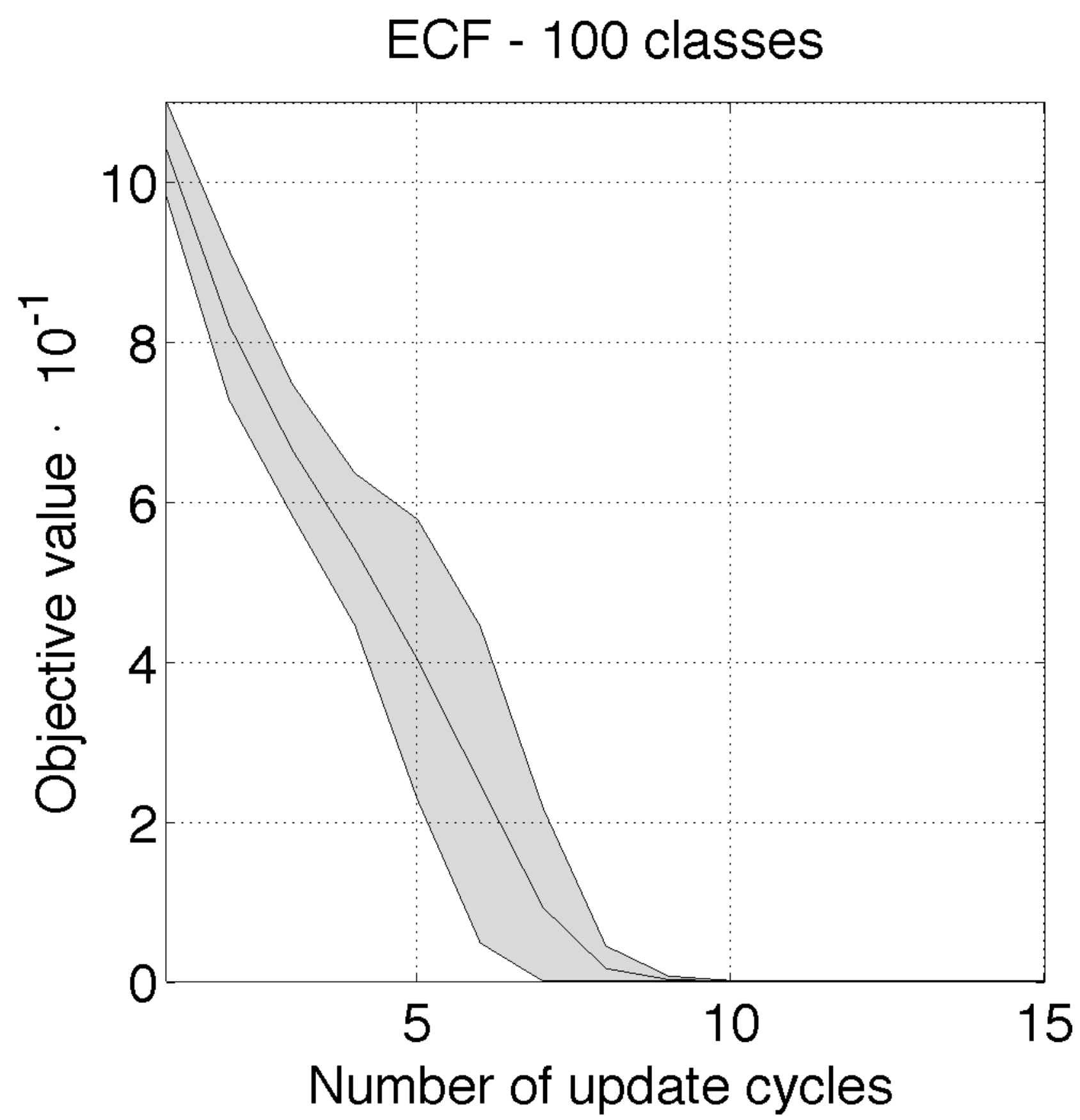}\\
	(a)&(b)\\
	\includegraphics[width=0.25\textwidth,height = 3.53 cm]{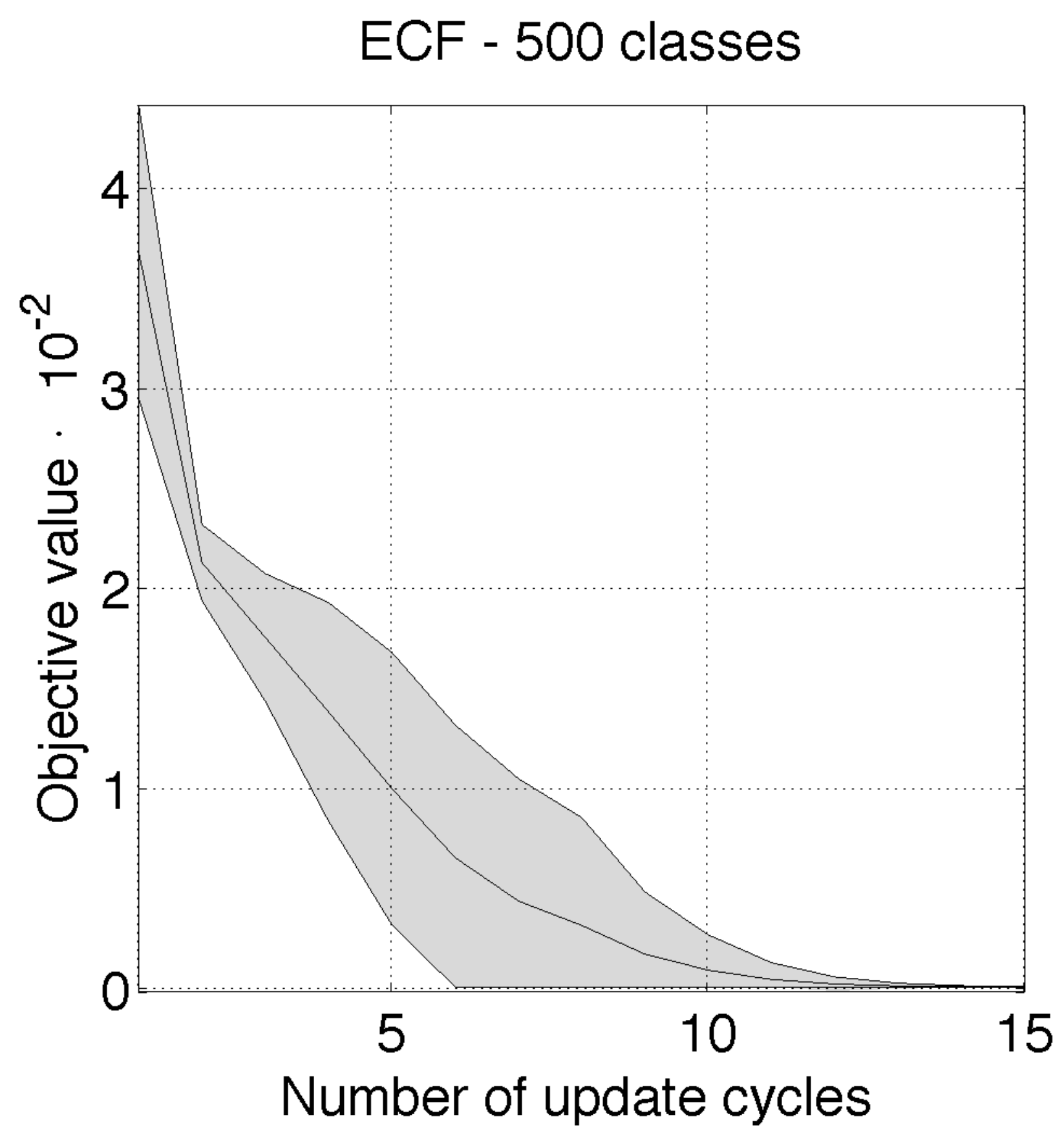}&
	\includegraphics[width=0.25\textwidth]{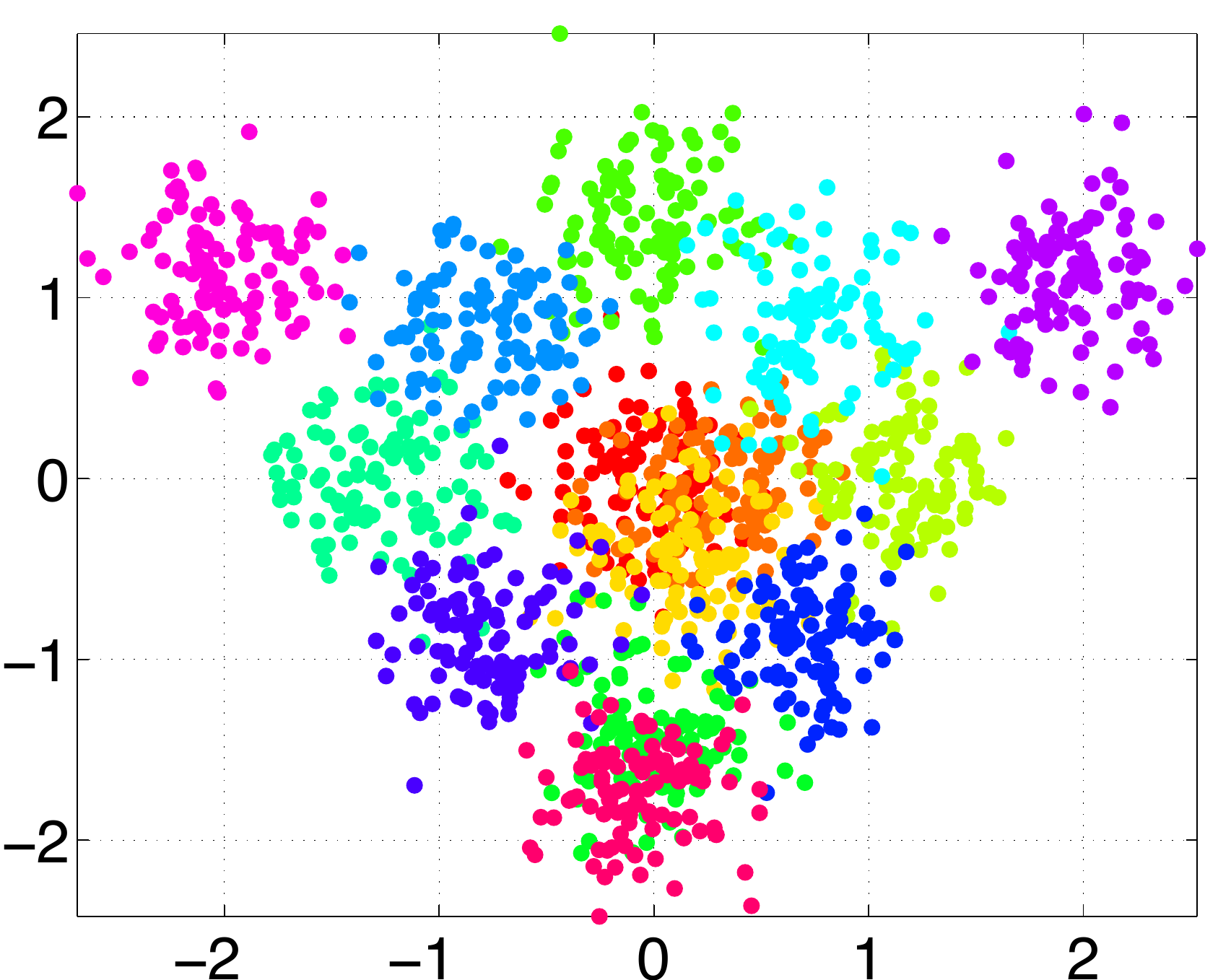}\\
	(c) & (d) \\
	\end{tabular}
	\vspace{-0.4 cm}
	\caption{(Mean objective value and standard deviation for 30 runs of ECF on a random $\D^G$ of $10$ classes (a), $100$ classes (b), and $500$ classes (c). (d) Toy problem synthetic data, where each color corresponds to a different category in the multi-class problem.}
	\label{fig:optimizationDisc}
\end{figure}

In order to better understand how ECF works we analyze both components separately. Then, to analyze if ECF converges to a good solution in terms of Frobenius norm we set $\varepsilon_d=0$ by generating a matrix $\D = \D^B$ which is the inner product matrix of  random discrete vectors, and thus, all the terms except of $\|\X^*\X^{*\top} - \D^B\|_{F}^{2}$ are zero. By doing that, we can empirically observe the magnitude of the optimization error $\varepsilon_o$. In order to do that we run ECF $30$ times on $100$ different $\D^B$ matrices of different sizes and calculate the average $\bar{\varepsilon_o}$. Figure \ref{fig:optimizationDisc} shows examples for different $\D^G$ matrices of size $10\times 10$, $100 \times 100$, and $500 \times 500$. In Figure \ref{fig:optimizationDisc} we can see how ECF converges to a solution with almost negligible optimization error after 15 iterations. In fact, the average objective value for all $3000$ runs of ECF on different $\D^B$'s after 15 update cycles (coordinate updates for all $\x^i$'s) is $\bar{\varepsilon_o} < 10^{-10}$. This implies, that ECF converges in average to a point with almost negligible objective value, and when applied to $\D$'s which are not computed from binary components the main source of the approximation error is the discretization error $\varepsilon_d$. Since ECF seeks to find a discrete decomposition of $\D$ this discretization error is unavoidable, and as we have seen empirically, ECF converges in average to a solution with almost negligible objective value.

\section{Experiments} \label{sec:experiments}
In this section we present the experimental results of the proposed Error-Correcting Factorization method. In order to do so, we first present the data, methods and settings.

\subsection{Data}
The proposed Error-Correcting Factorization method was applied to a total of $8$ datasets. In order to provide a deep analysis and understanding of the method, we synthetically generated a \textit{Toy} problem consisting of $k=14$ classes, where each class contained $100$ two dimensional points sampled from a Gaussian distribution with same standard deviation but different means. Figure \ref{fig:optimizationDisc}(d) shows the synthetic multi-class generated data, where each color corresponds to a different category. We selected $5$ well-known UCI datasets: \textit{Glass}, \textit{Segmentation}, \textit{Ecoli}, \textit{Yeast} and \textit{Vowel} that range in complexity and number of classes. Finally, we apply the classification methodology in two challenging computer vision categorization problems. First, we test the methods in a real traffic sign categorization problem consisting of 36 traffic sign classes. Second, 50 classes from the ARFaces~\cite{ARFaces} dataset are classified using the present methodology. These datasets are public upon request to the authors. Table \ref{tab:data} shows the characteristics of the different datasets.

\begin{figure}
	\centering
	\begin{tabular}{cc}
    \includegraphics[height = 2.3 cm]{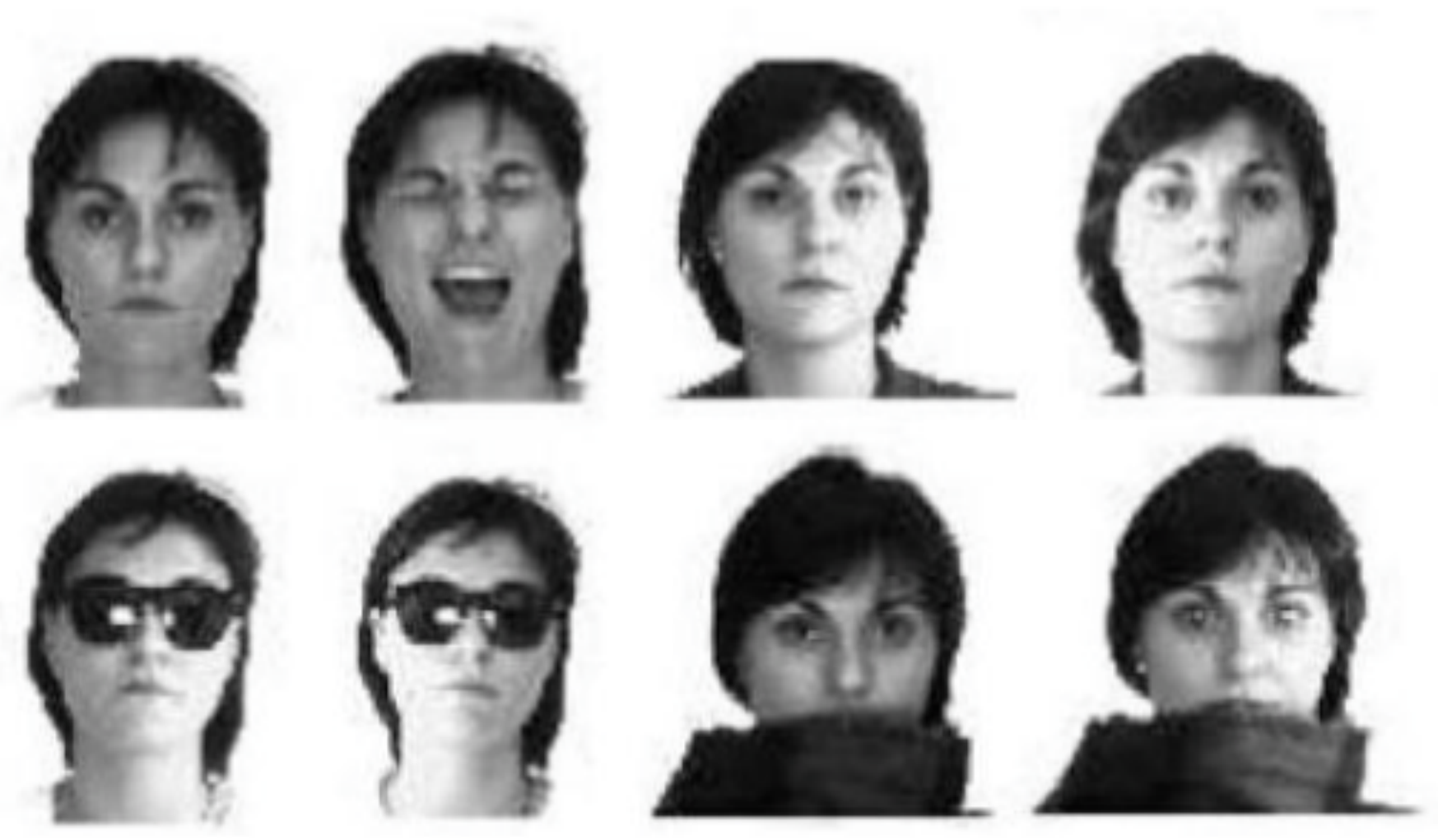}&\includegraphics[height = 2.3 cm]{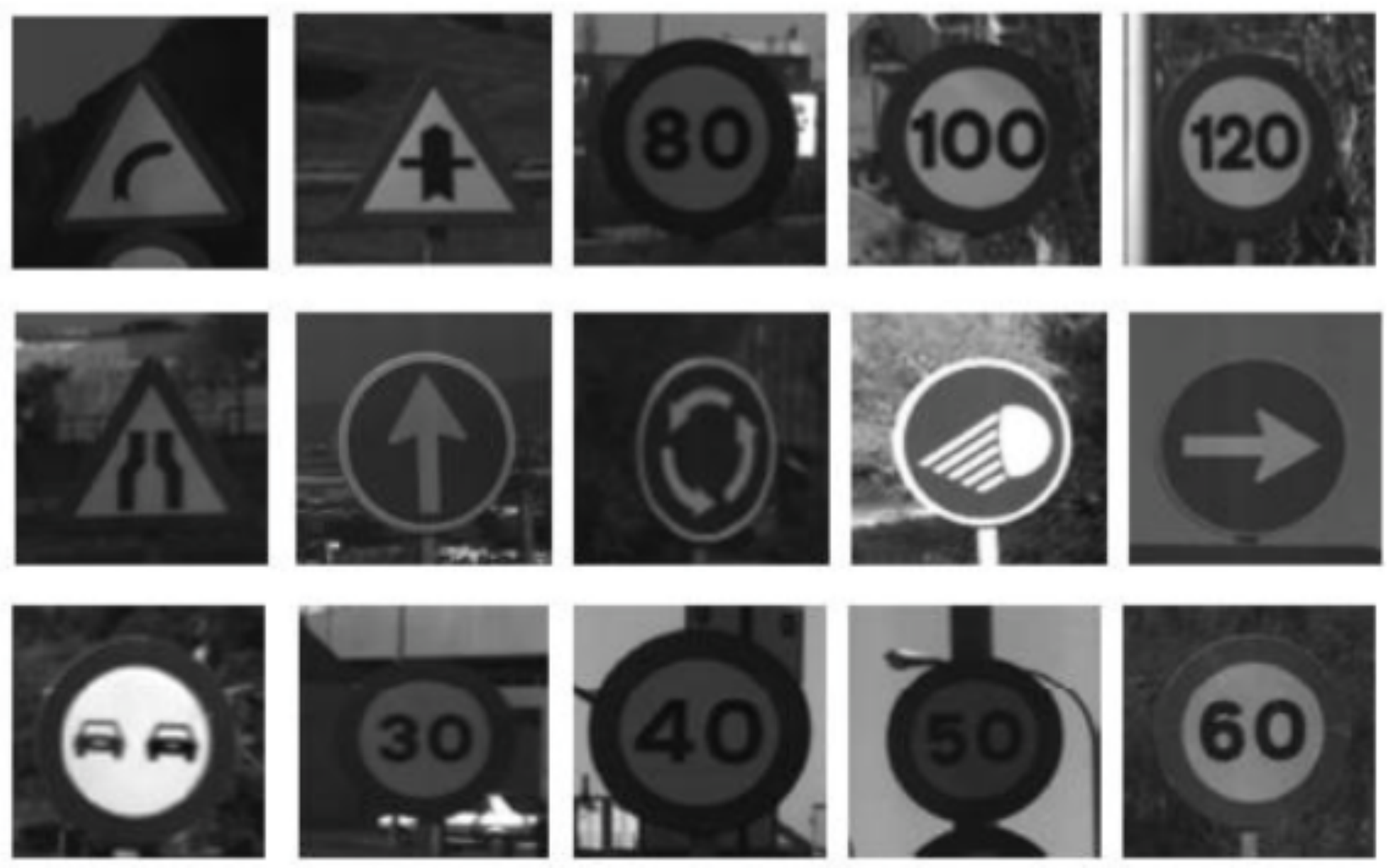} \\
    (a)&(b)\\
    \end{tabular}
    \vspace{-0.2 cm}
\caption{Visual examples for the \textit{ARFace} and \textit{Traffic} datasets.} \label{fig:datasets}
\end{figure}

$\bullet$\textbf{Traffic sign categorization:} We test ECF on a real traffic sign categorization problem, of 36 classes ~\cite{casacuberta}. The dataset contains a total of 3481 samples of size 32$\times$32, filtered using the Weickert anisotropic filter, masked to exclude the background pixels, and equalized to prevent the effects of illumination changes. These feature vectors are then projected into a 100 feature vector by means of PCA. A visual sample is show in Figure \ref{fig:datasets}(a).

$\bullet$\textbf{ARFaces classification:} The ARFace database~\cite{ARFaces} is composed of 26 face images
from 126 different subjects (from which 50 are selected), portraying different expressions and complements. An example is shown in  Figure~\ref{fig:datasets}(b).

 \begin{table}
\centering 
\scriptsize    
\caption{Dataset characteristics.}
   	   	\begin{tabular}{|c|c|c|c|c|c|c|c|c|}
   	   		\hline
   	   		& Glass & Segment. & Ecoli & Yeast & Vowel & Toy & Traffic & ARFace \\ \hline
   	   		\#s & 214 & 2310 & 336 & 1484 & 990 & 400 & 3481  & 1300 \\\hline
   	   		\#f & 9 & 19 & 8 & 8 & 10 & 2 & 100  & 120 \\\hline
   	   		\#c & 7 & 7 & 8 & 10 & 11 & 14 & 36  & 50 \\\hline
   	   	\end{tabular}
   	   	\label{tab:data}
 \end{table}

\subsection{Methods and settings}

We compared the proposed Error-Correcting Factorization method, with the standard predefined One vs. All (\textit{OVA}) and One vs. One (\textit{OVO}) approaches \cite{rifkin,hastie}. In addition, we introduce two random designs for ECOC matrices. In the first one, we generated random ECOC coding matrices fixing the general correction capability to a certain value (\textit{RAND}). In the second, we generate a Dense Random coding matrix \cite{Allwein00reducingmulticlass} (\textit{DENSE}). These comparisons enable us to analyze the effect of reorganizing the inter-class correcting capabilities of an ECOC matrix. Finally, in order to compare our proposal with state-of-the-art methods, we also used the Spectral ECOC (\textit{S-ECOC})  method \cite{spectralECOC} and the Relaxed Hierarchy \cite{iccv2011gao} (\textit{R-H}) . Finally we propose two different flavors of ECF, \textit{ECF-H} and \textit{ECF-E}.  In \textit{ECF-H} we compute the design matrix $\D$ in order to allocate the correction capabilities on those classes that are \textbf{hard} to discriminate. On the other hand, for \textit{ECF-E} we compute  $\D$ allocating correction to those classes that are \textbf{easy} to discriminate.  $\D$ is computed as the Mahalanobis distance between each pair of classes. Although, there exist a number of approaches to define $\D$ from data \cite{iccv2011gao,zhao2013sparse,spectralECOC}, i.e. the margin between each pair of classes (after training a \textit{One vs. One} SVM classifier), we experimentally observed that the Mahalanobis distance provides good generalization and leverages the computational cost of training a \textit{One vs. One} SVM classifier. All the reported classification accuracies are the mean of a stratified $5-$fold cross-validation on the aforementioned datasets. For all methods we used an SVM classifier with RBF kernel. The parameters $C$ and $\gamma$ were tunned by cross-validation on a validation subset of the data using an inner $2-$fold cross-validation. The parameter $C$ was tunned on a grid-search on a log sampling in the range $[0,10^{10}]$, and the $\gamma$ parameter was equivalently tuned on a equidistant linear sampling in the range $[0,1]$, we used the libsvm implementation available at \cite{libsvm}. For both \textit{ECF-H} and \textit{ECF-E} we run the factorization forcing different minimum distance between classes by setting $\P \in \1 \cdot \{1,3,5,7,10\}$ . For the Relaxed Hierarchy method \cite{iccv2011gao} we used values for $\rho \in \{0.1, 0.4, 0.7, 0.9\}$. In all the compared methods that use a decoding function (e.g all tested methods but the one in \cite{iccv2011gao}) we used both the Hamming Decoding (HD) and the Loss-Weighted decoding (LWD) \cite{decoc}.

\subsection{Experimental Results}

\begin{figure*}
	\centering
	\setlength{\tabcolsep}{0pt}
\begin{tabular}{cccc}
	
		\includegraphics[width=0.25\textwidth]{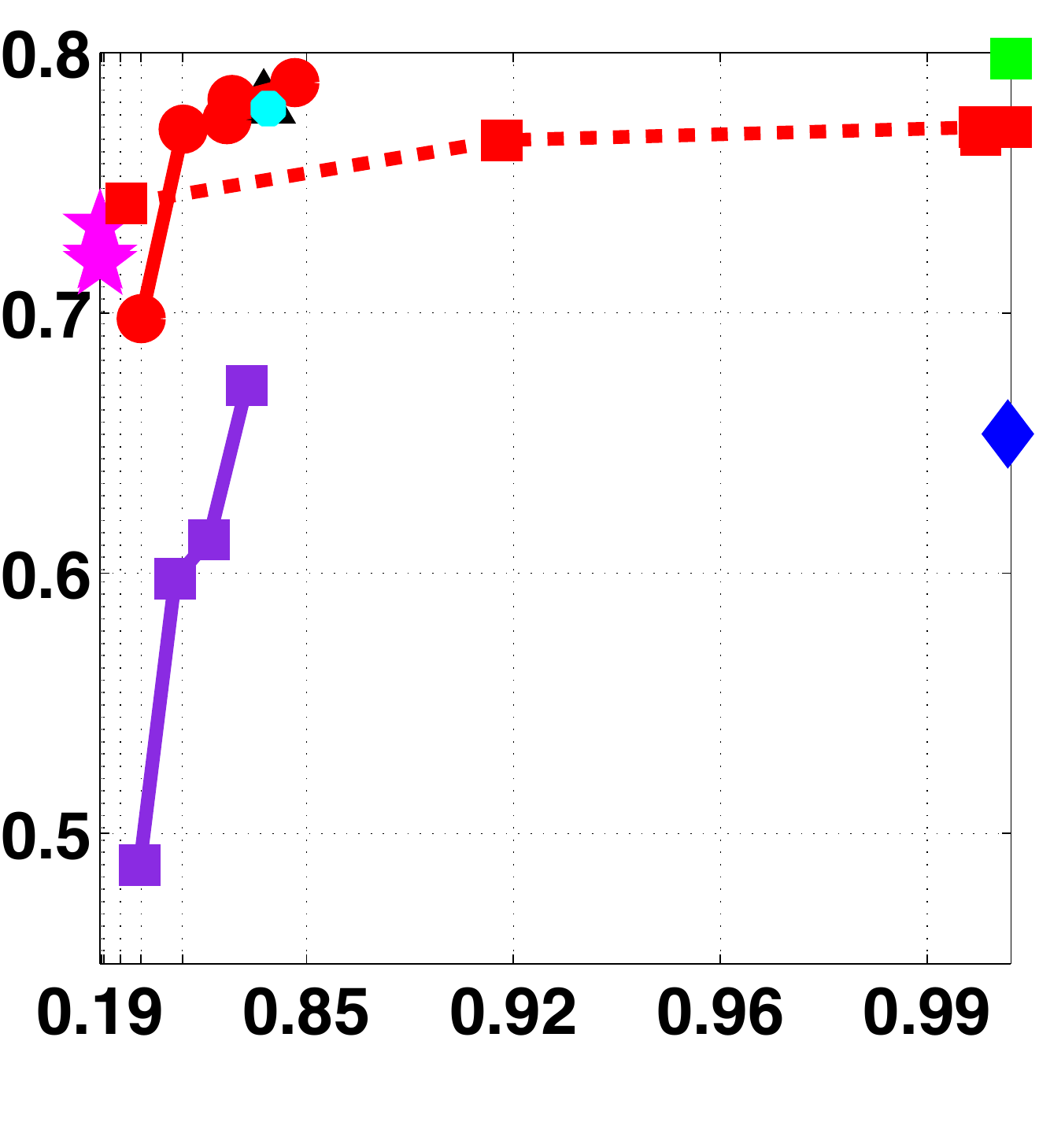} & 
		\includegraphics[width=0.25\textwidth]{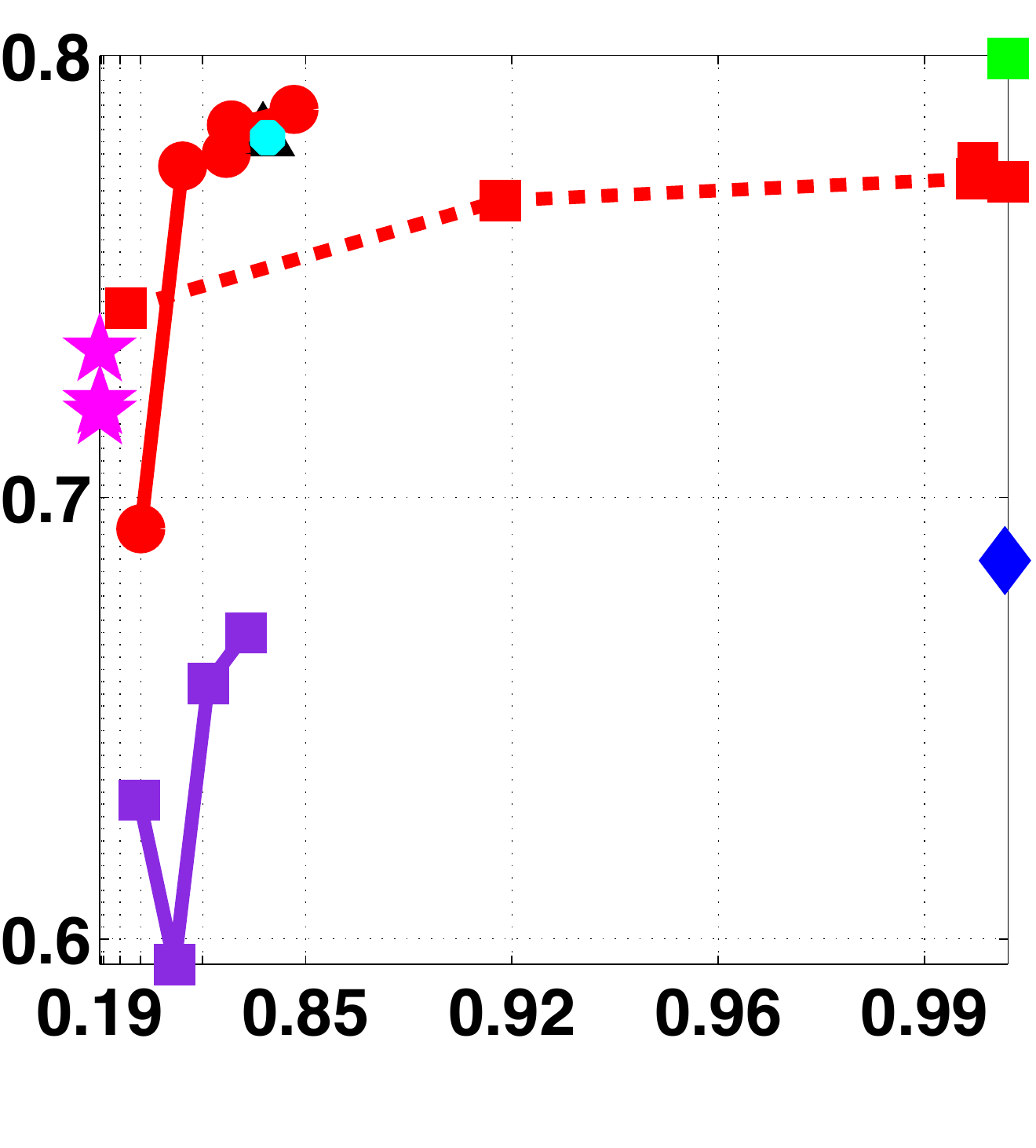} & 
		\includegraphics[width=0.25\textwidth]{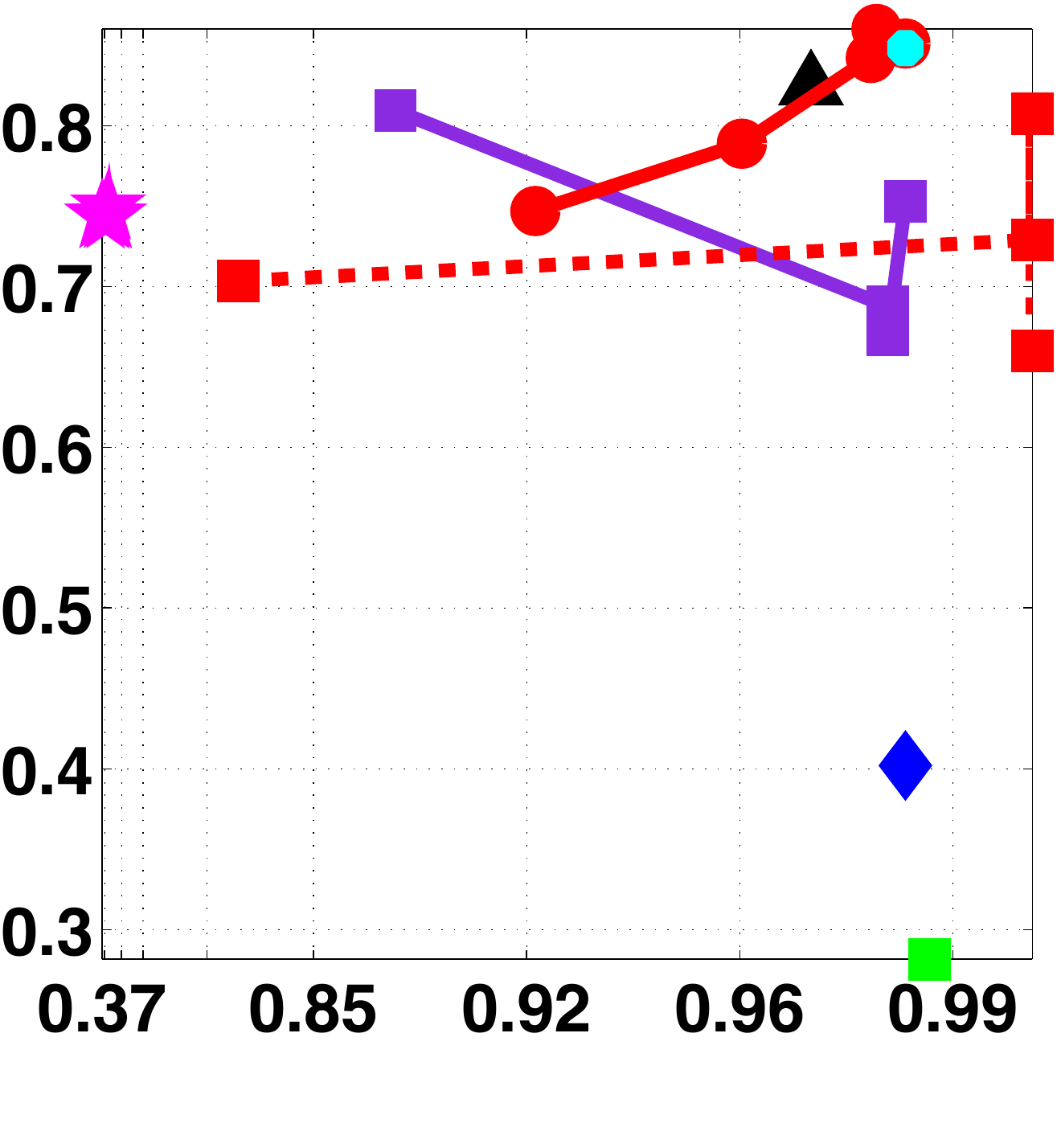} &
		\includegraphics[width=0.25\textwidth]{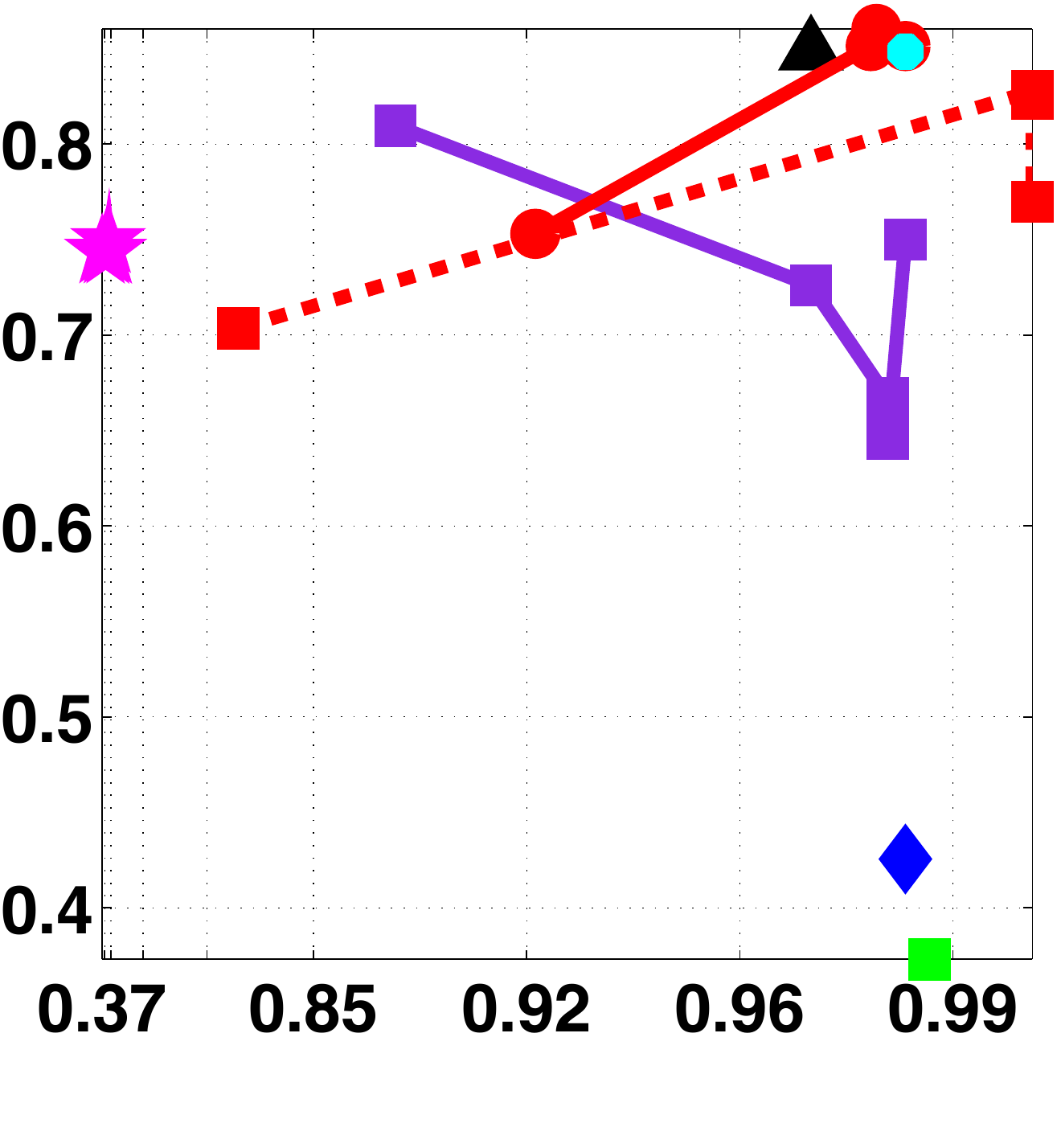} \vspace{-0.4 cm} \\ 
		
		(a) \textit{Toy} dataset HD&(b) \textit{Toy} dataset LWD&(c) \textit{Ecoli} dataset HD&(d) \textit{Ecoli} dataset LWD\\
		
	    \includegraphics[width=0.25\textwidth]{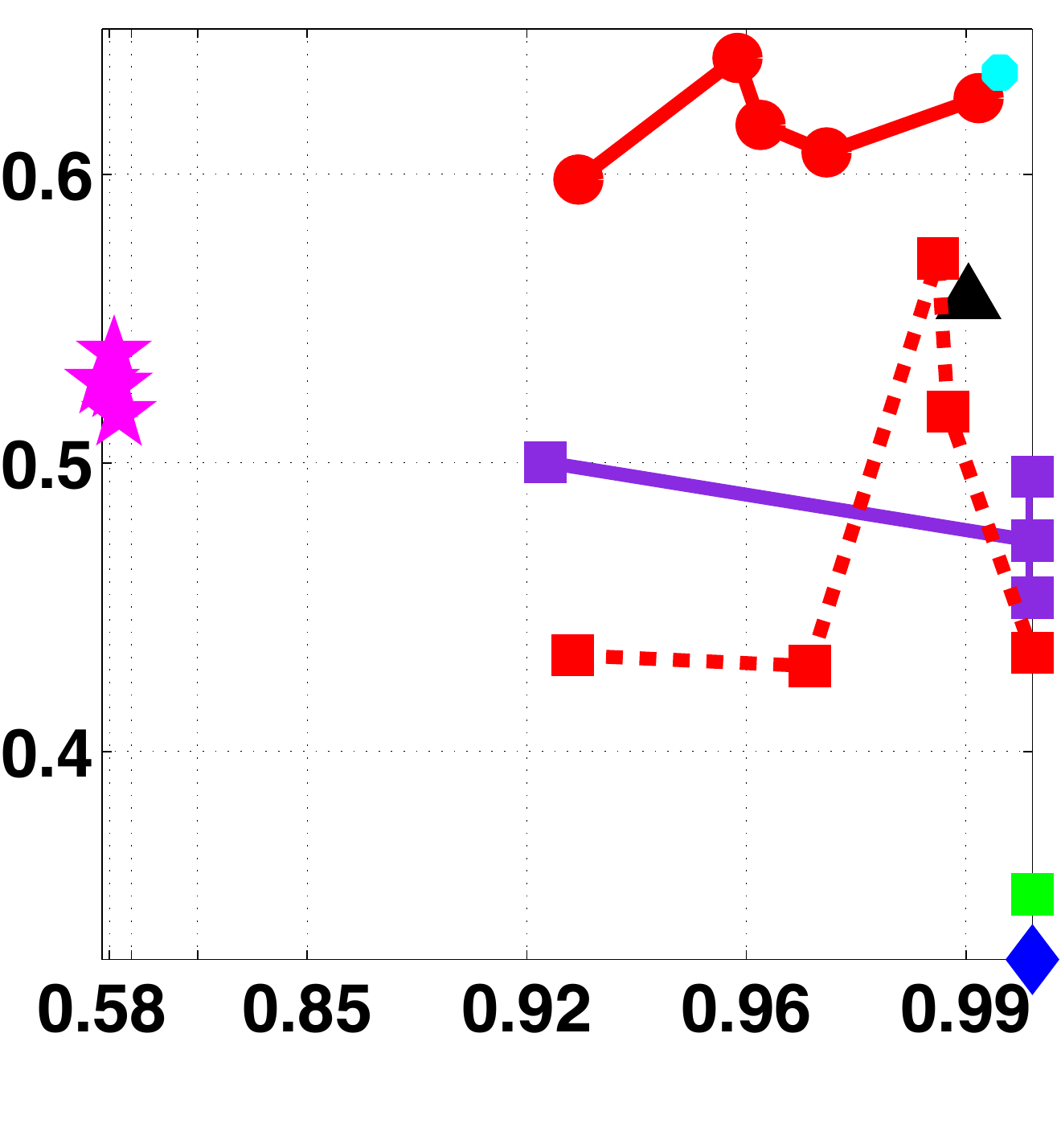} & 
		\includegraphics[width=0.25\textwidth]{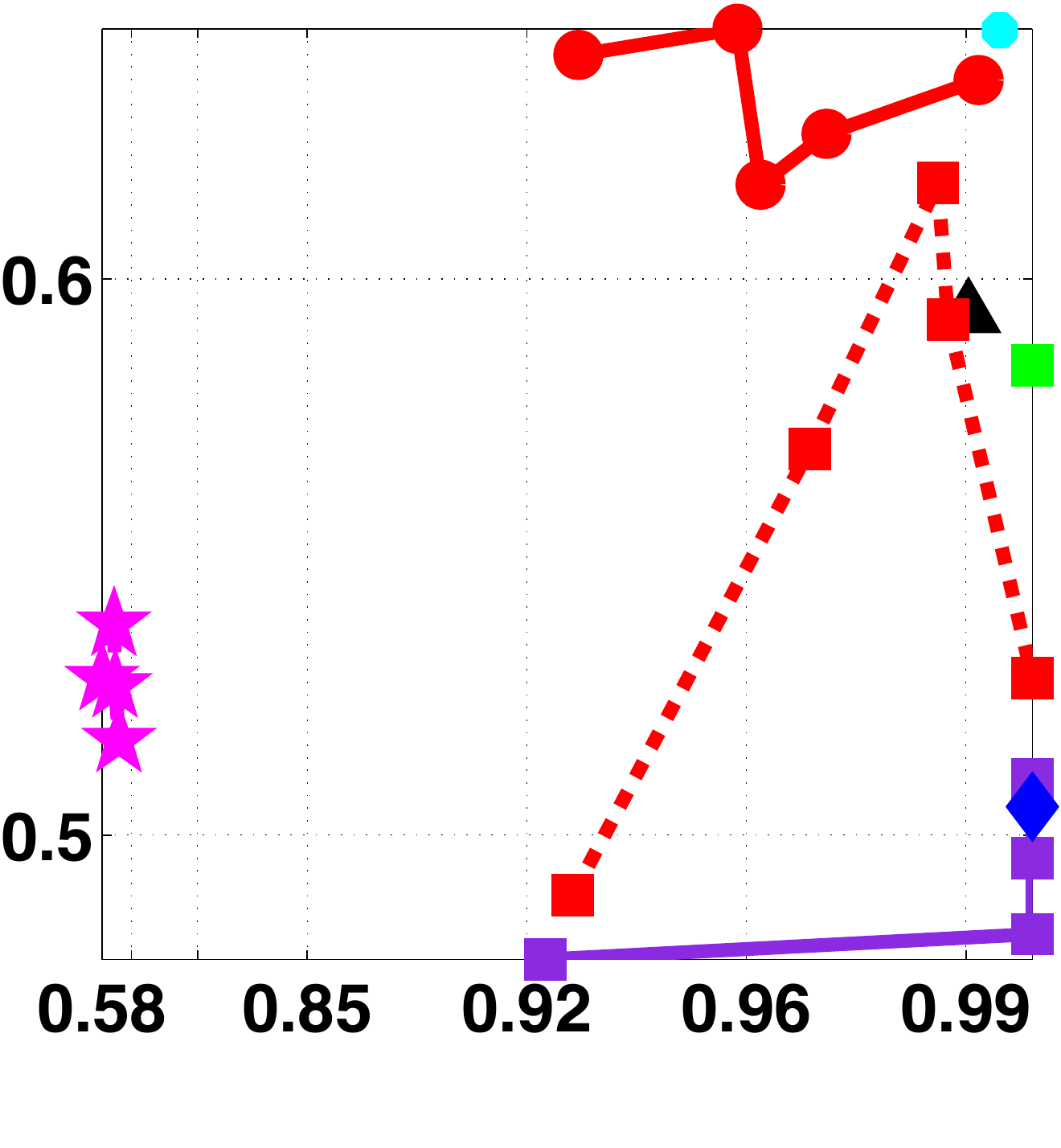} & 
		\includegraphics[width=0.25\textwidth]{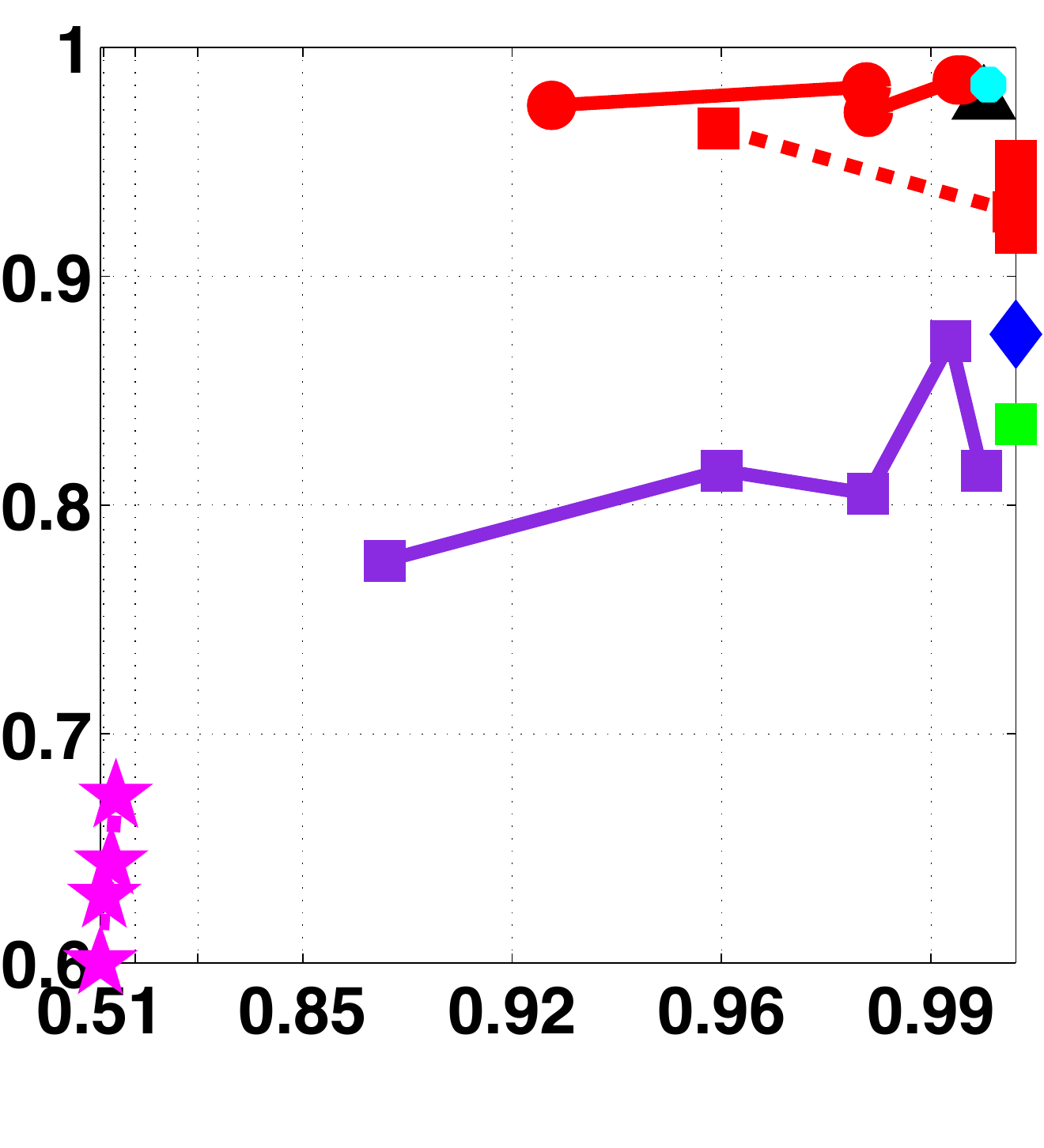} &
		\includegraphics[width=0.25\textwidth]{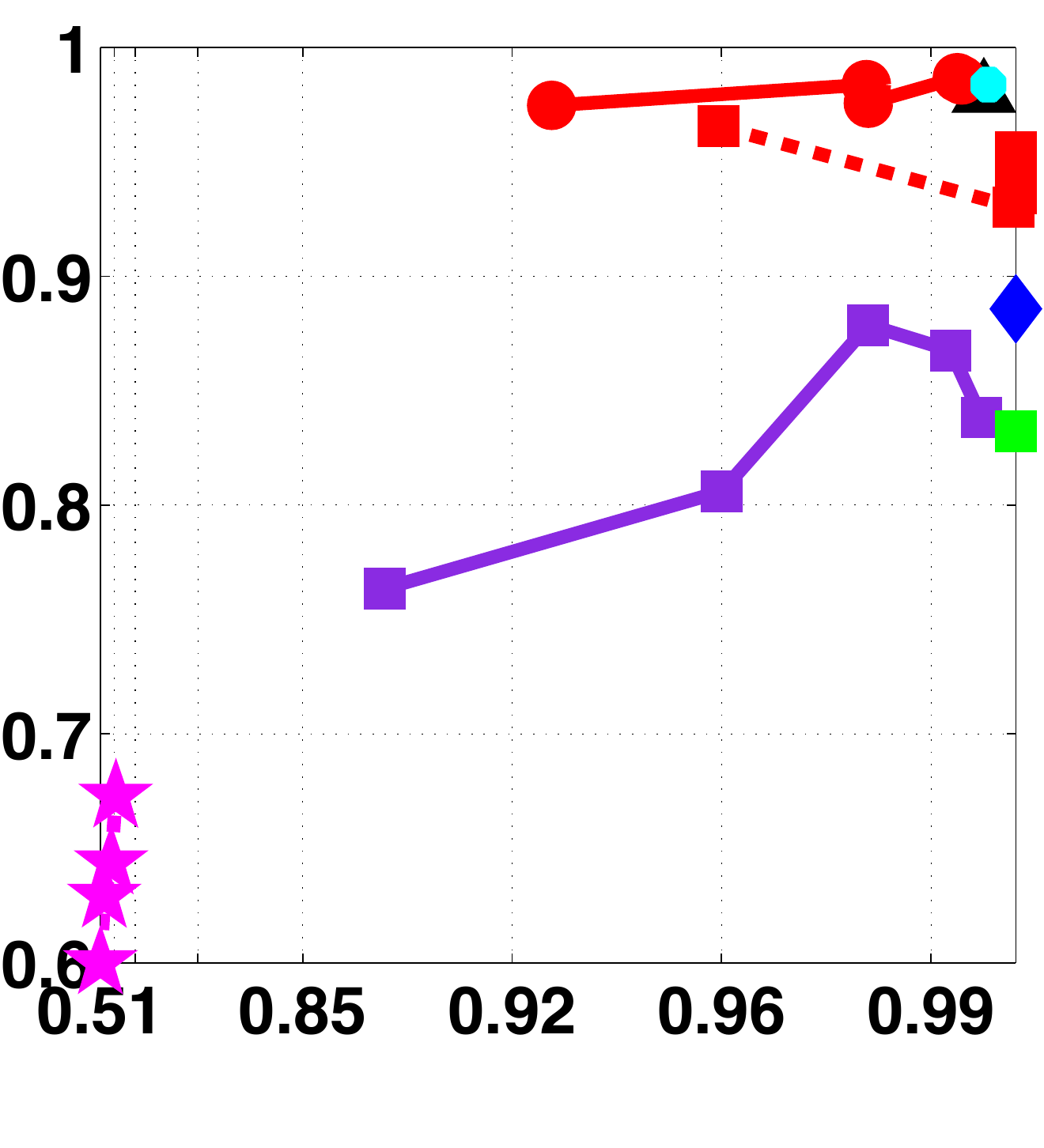}\vspace{-0.4 cm}\\
				
		(e) \textit{Glass} dataset HD&(f) \textit{Glass} dataset LWD&(j) \textit{Vowel} dataset HD&(h) \textit{Vowel} dataset LWD\\

		\includegraphics[width=0.25\textwidth]{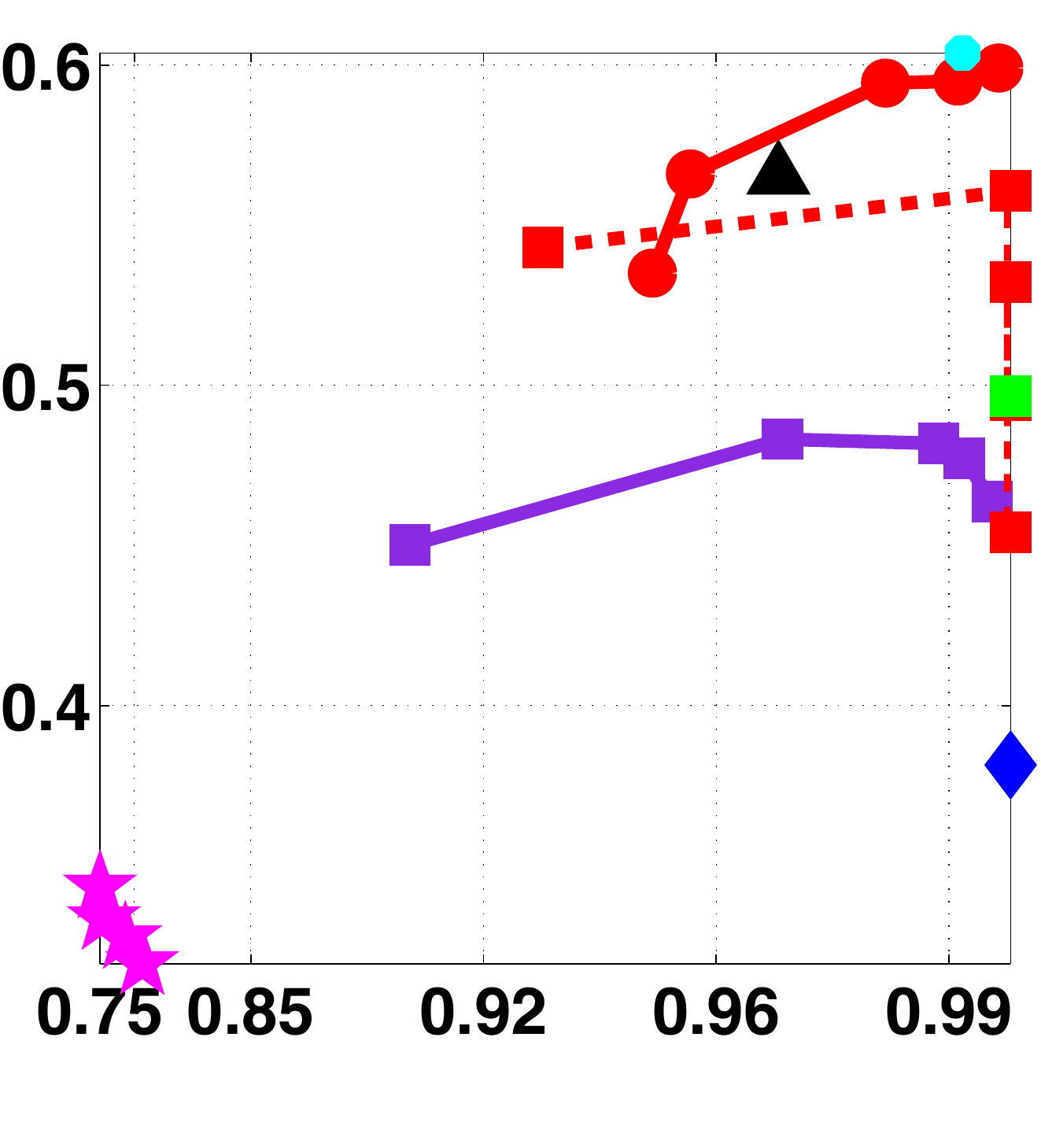} & 
		\includegraphics[width=0.25\textwidth]{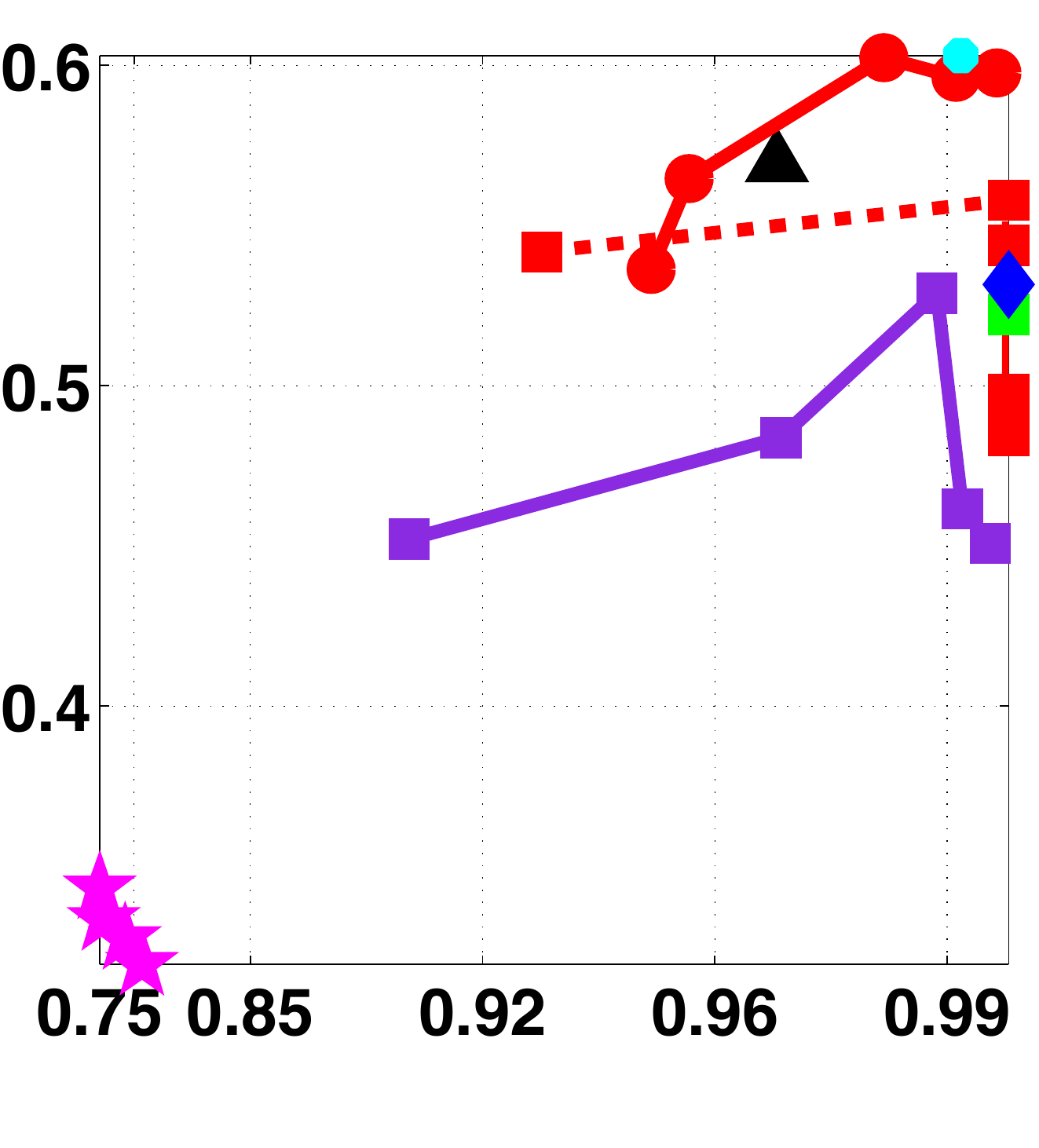} & 
		\includegraphics[width=0.25\textwidth]{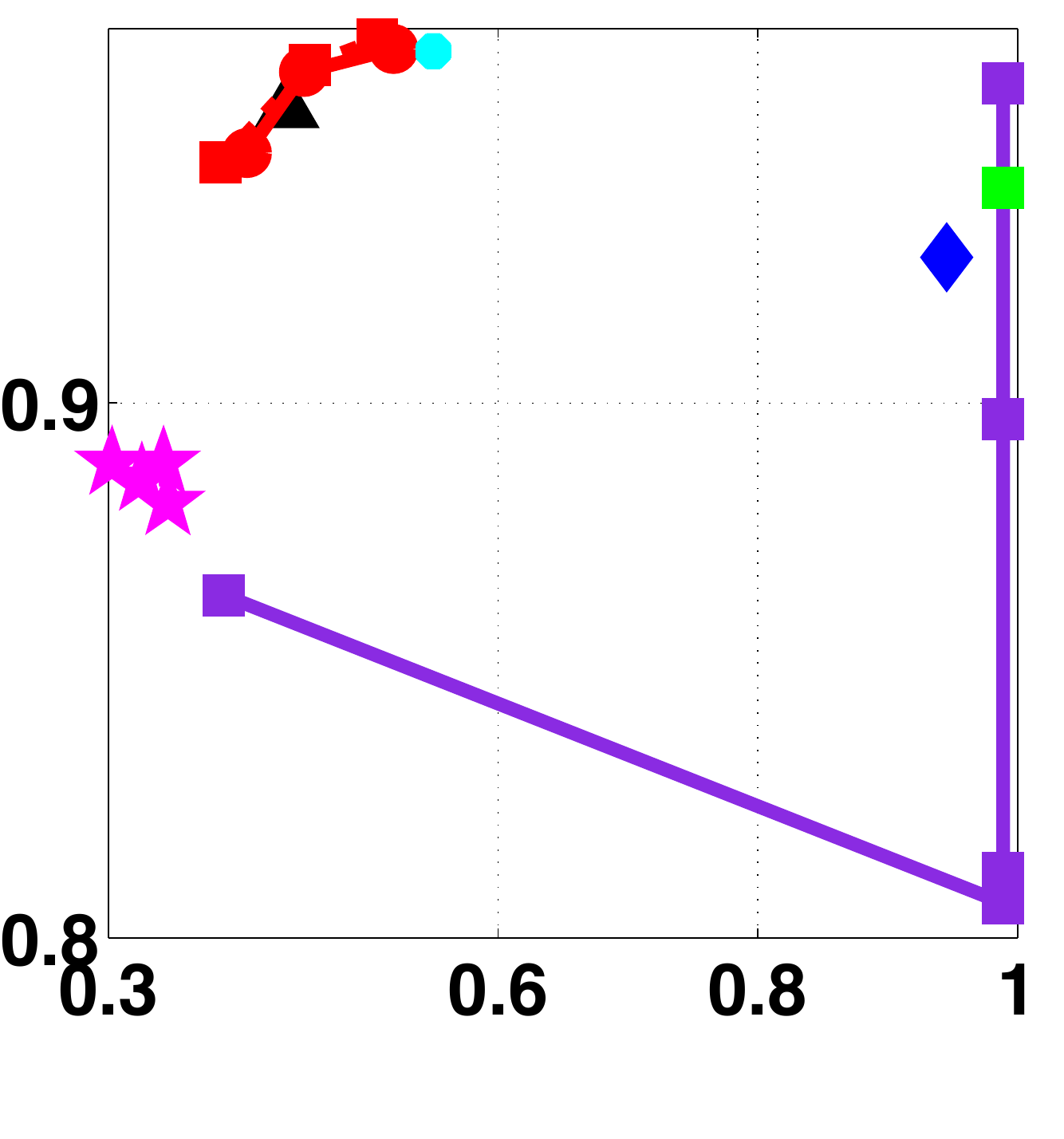} &
		\includegraphics[width=0.25\textwidth]{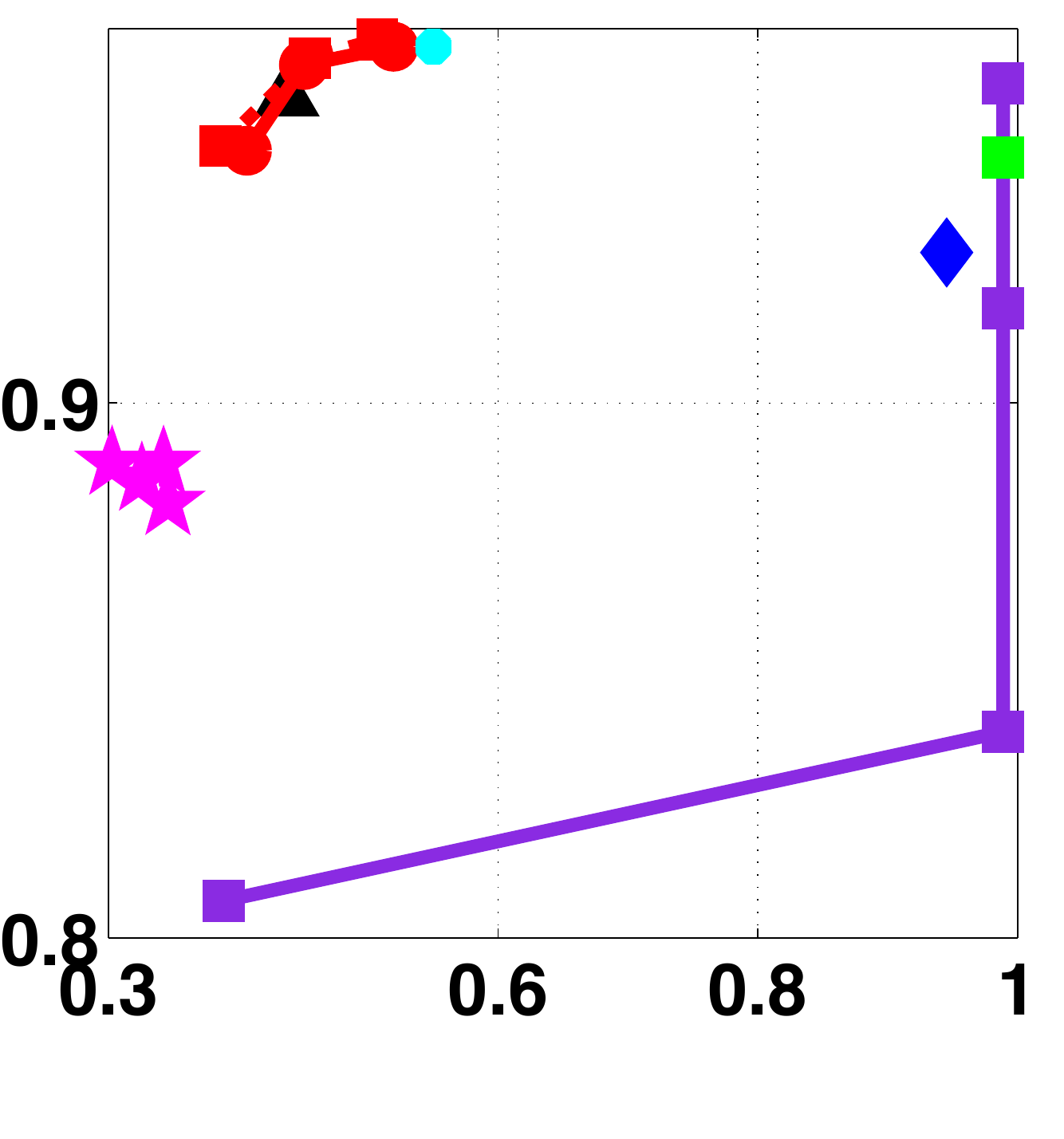} \vspace{-0.4 cm}\\

		(i) \textit{Yeast} dataset HD&(j) \textit{Yeast} dataset LWD&(k) \textit{Segmentation} dataset HD&(l) \textit{Segmentation} dataset LWD\\

		\includegraphics[width=0.25\textwidth, height= 4.5 cm]{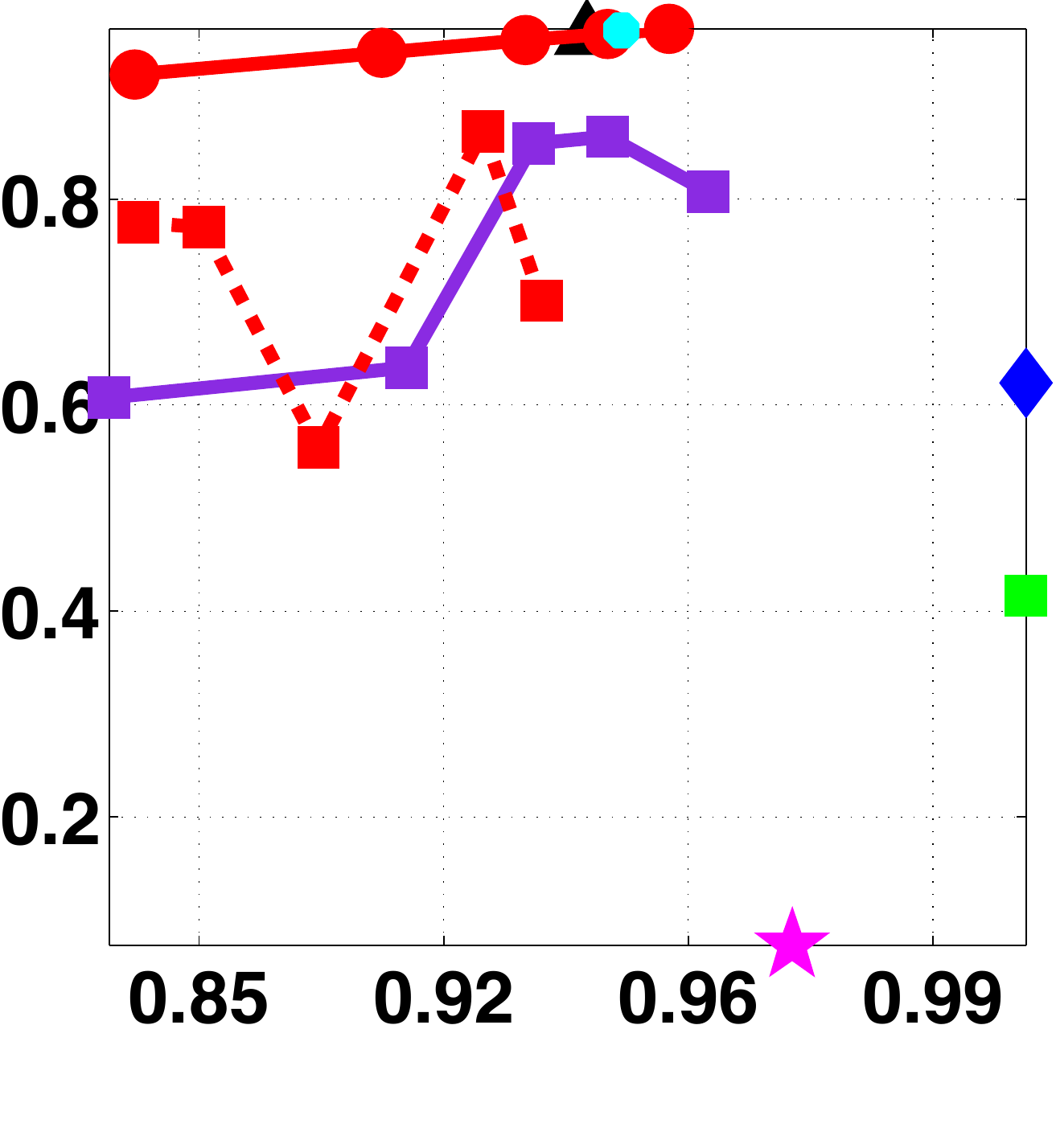} & 
		\includegraphics[width=0.25\textwidth, height= 4.5 cm]{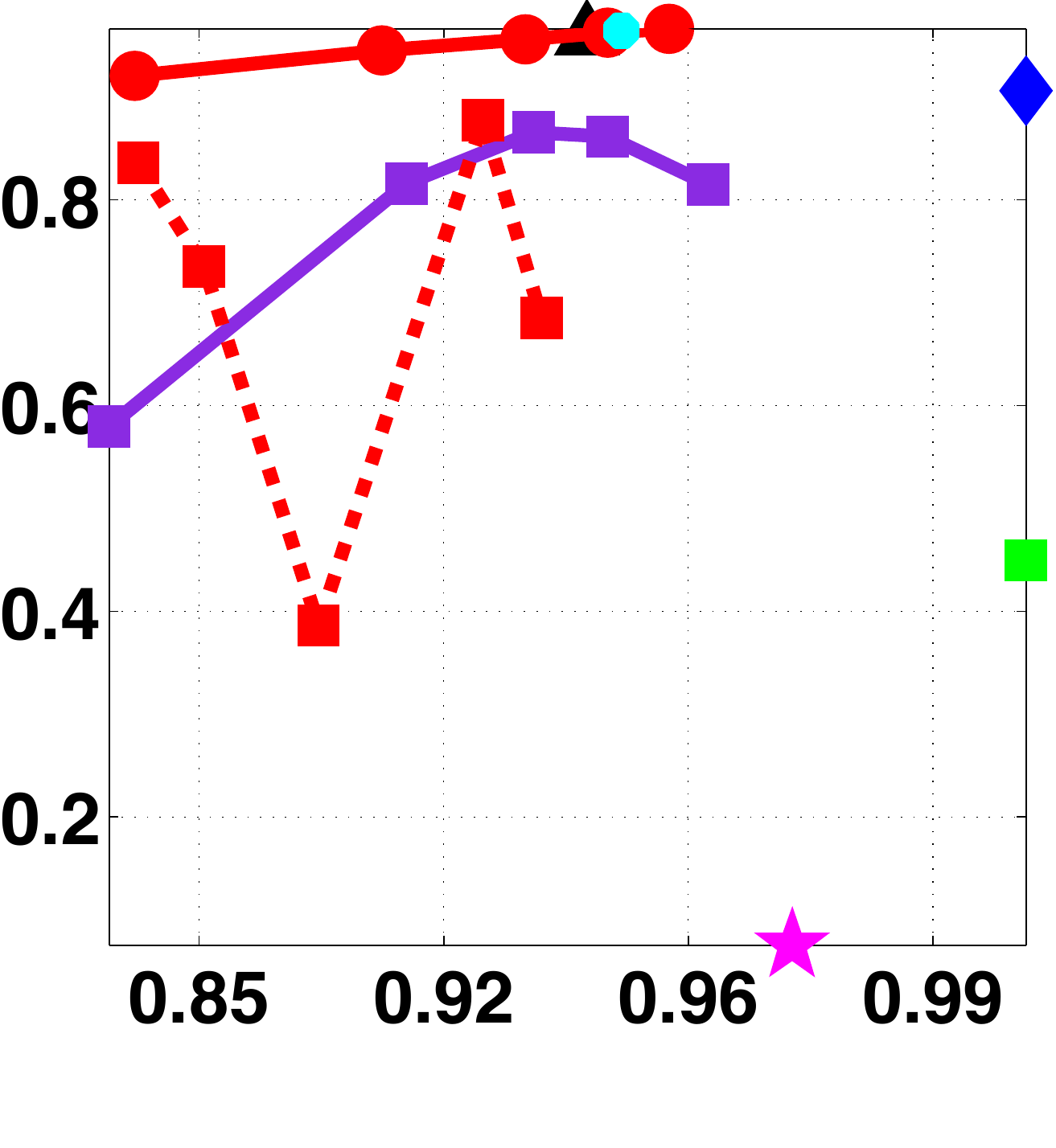} & 
		\includegraphics[width=0.25\textwidth, height= 4.5 cm]{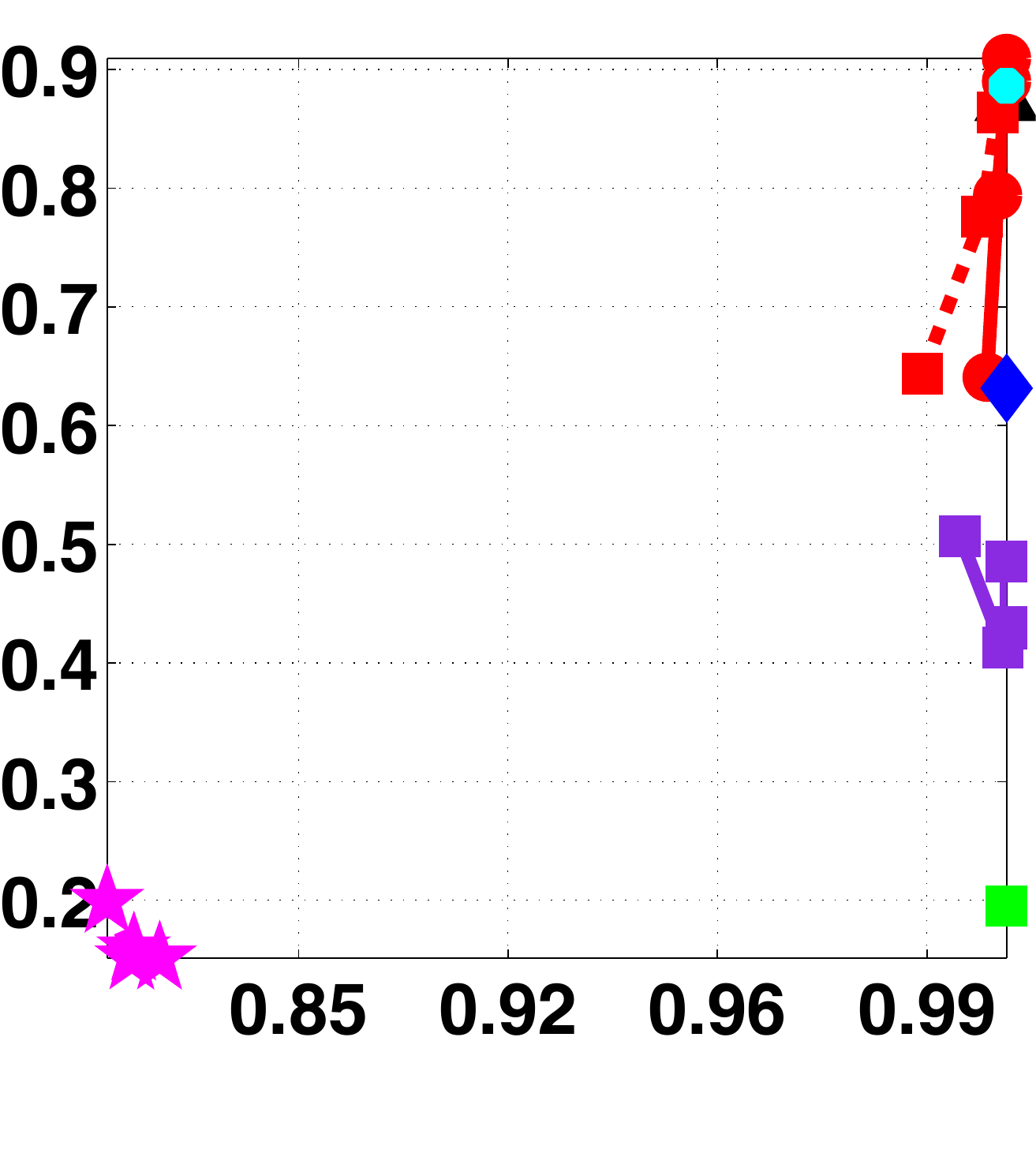} &
		\includegraphics[width=0.25\textwidth, height= 4.5 cm]{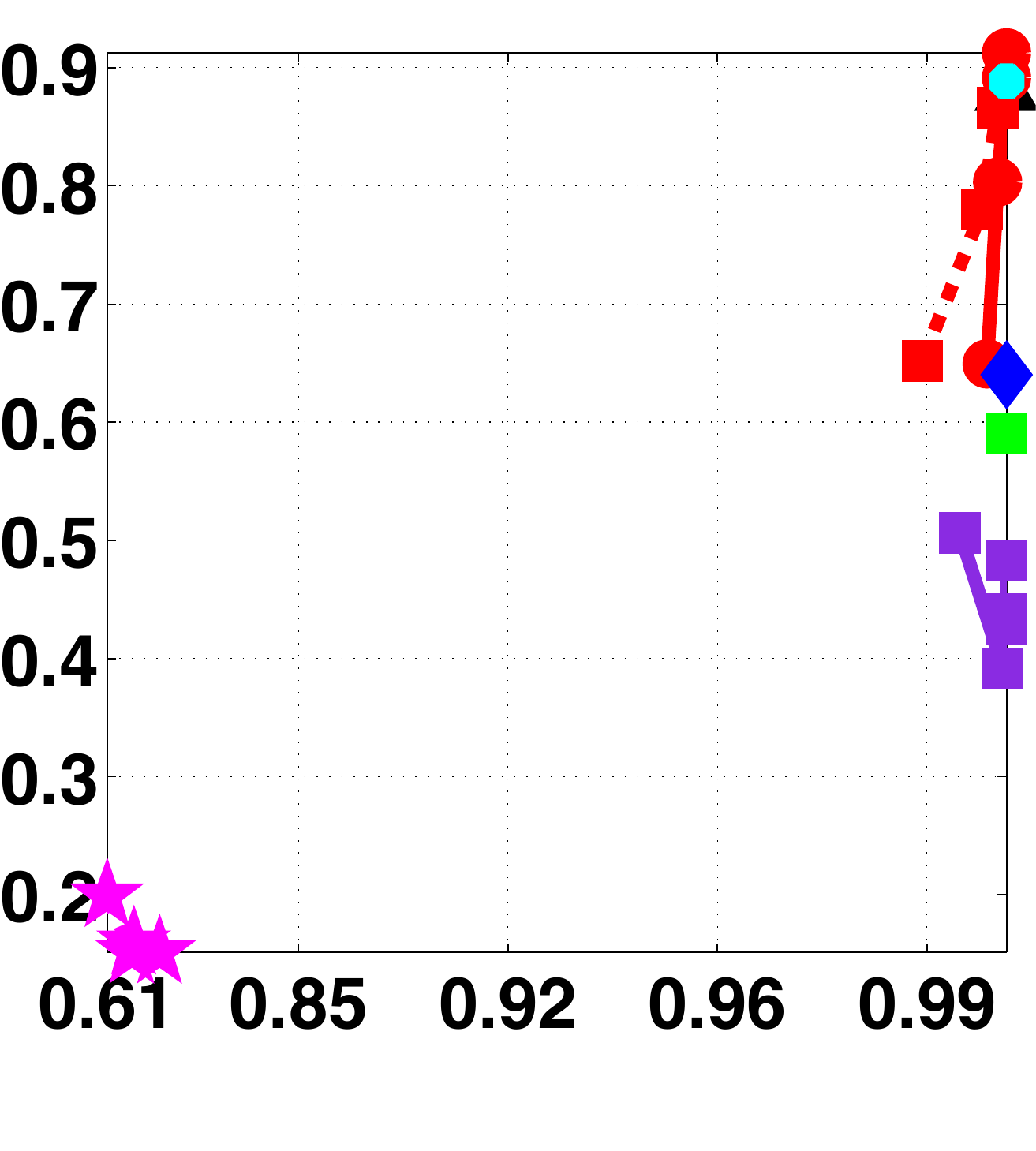} \vspace{-0.4 cm}\\
		(e) \textit{Traffic} dataset HD&(f) \textit{Traffic} dataset LWD&(j) \textit{ARFace} dataset HD&(h) \textit{ARFace} dataset LWD\\

\end{tabular}
\includegraphics[width=12 cm]{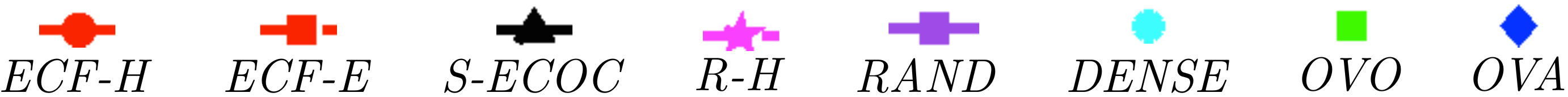} \\
\caption{Multi-class classification accuracy (y axis) as a function of the relative computational complexity (x axis) for all datasets and both decoding measures.}
\label{fig:results_svs}
\end{figure*}

\begin{figure*}
				\centering
				\setlength{\tabcolsep}{0pt}
	\begin{tabular}{c c c c}

		\includegraphics[width=0.25\textwidth]{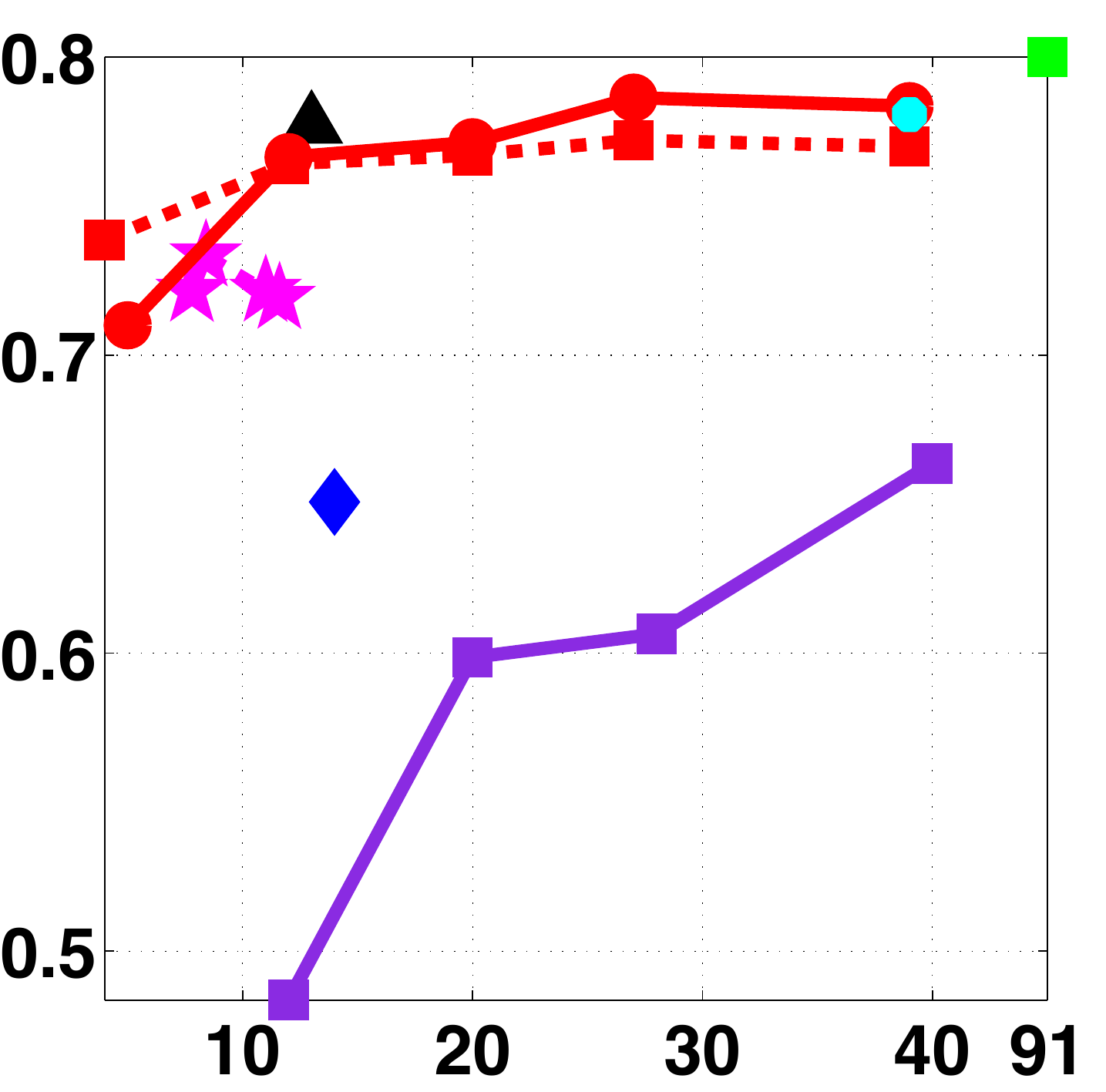} & 
		\includegraphics[width=0.25\textwidth]{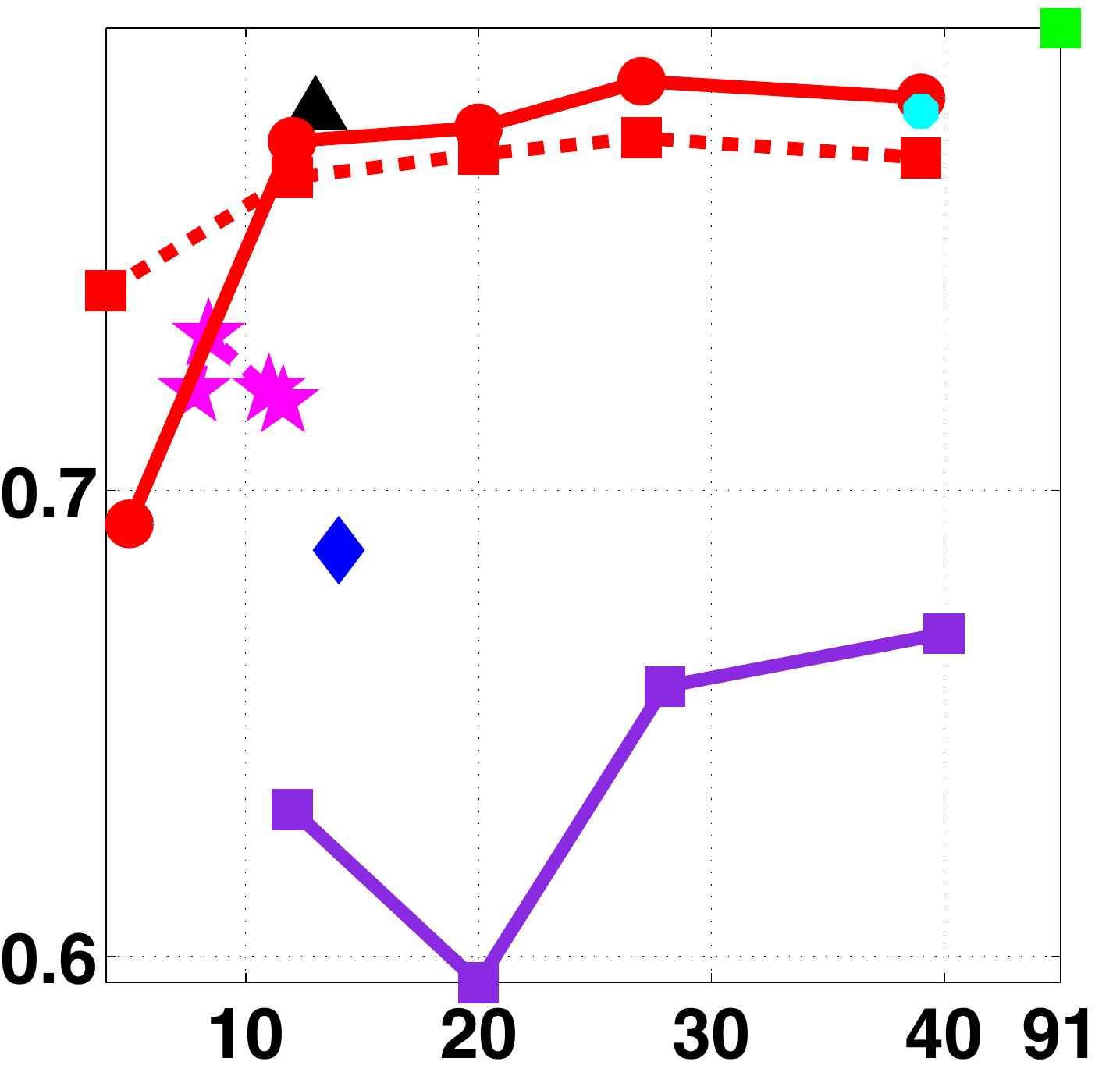} & 
		\includegraphics[width=0.25\textwidth]{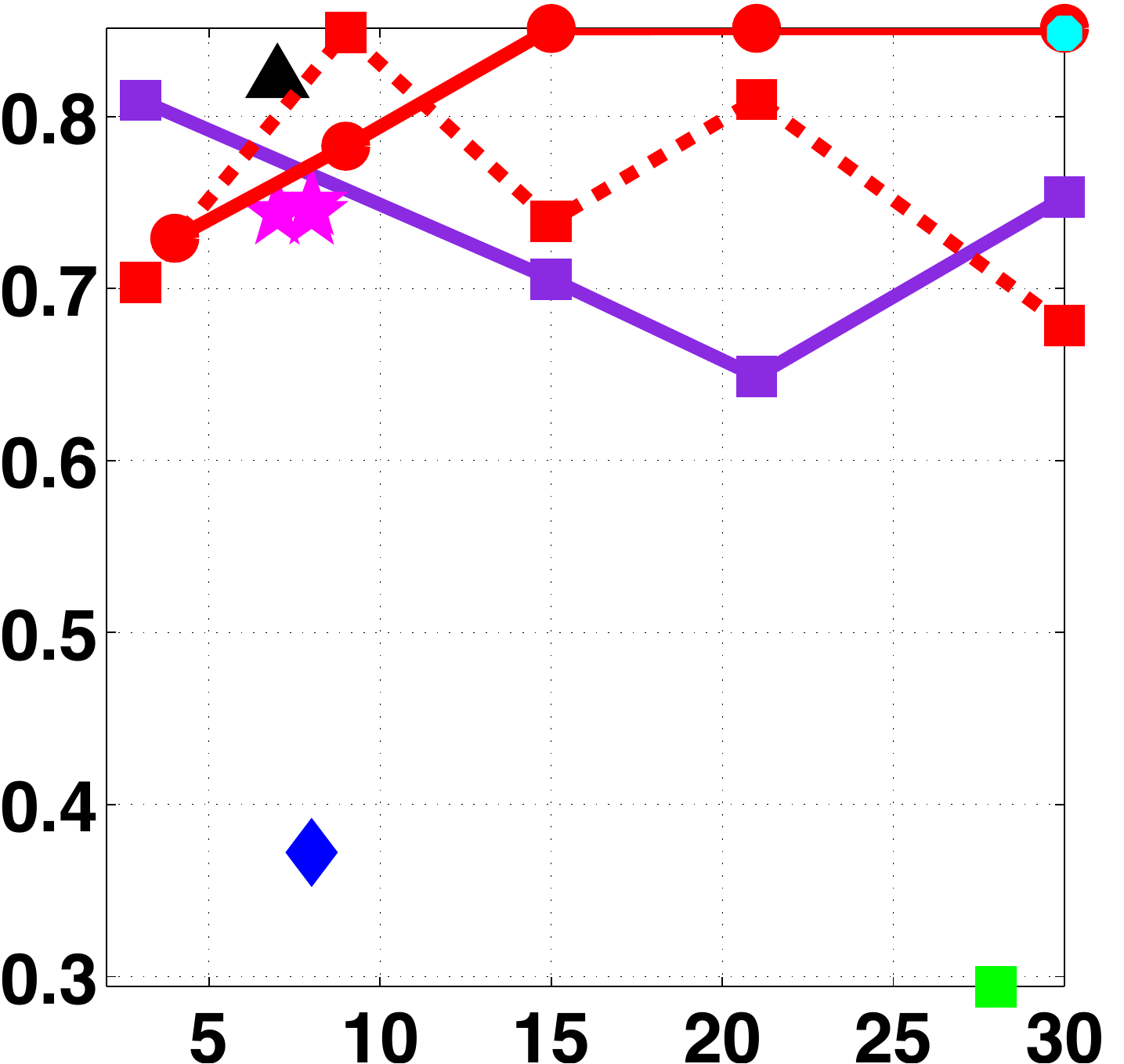} &
		\includegraphics[width=0.25\textwidth]{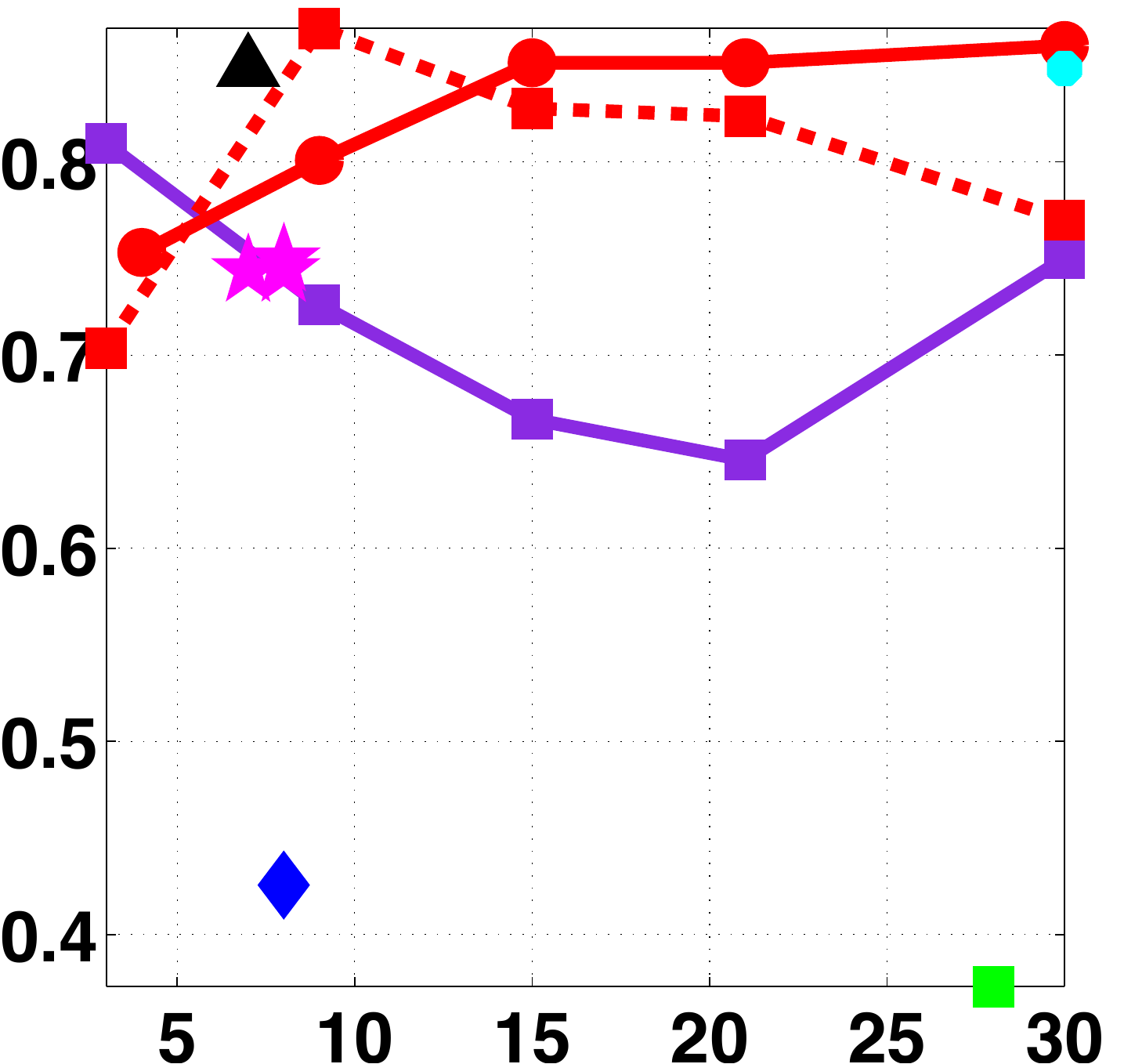} \\

		(a) \textit{Toy} dataset HD&(b) \textit{Toy} dataset LWD&(c) \textit{Ecoli} dataset HD&(d) \textit{Ecoli} dataset LWD\\
		\includegraphics[width=0.25\textwidth]{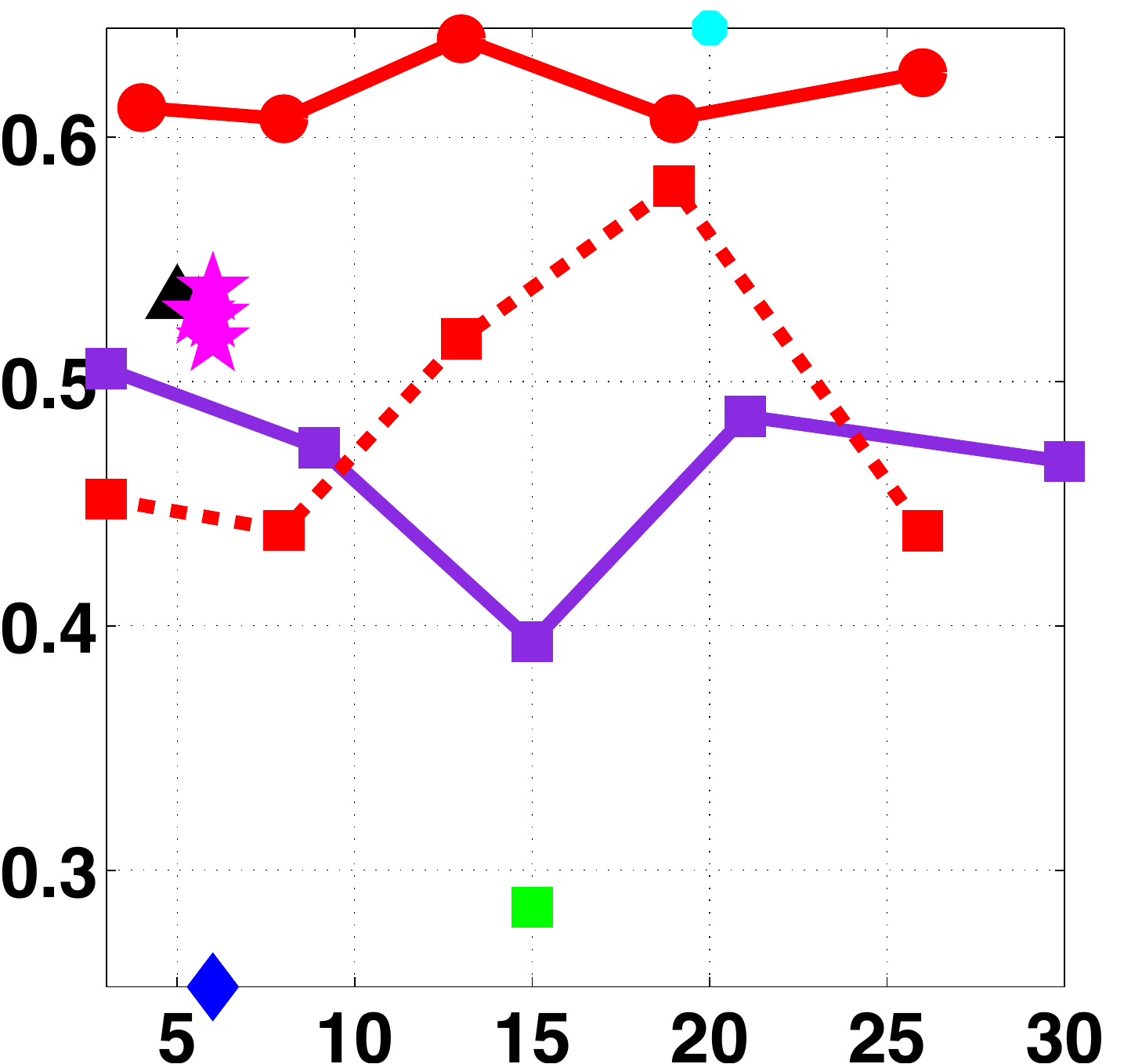} & 
		\includegraphics[width=0.25\textwidth]{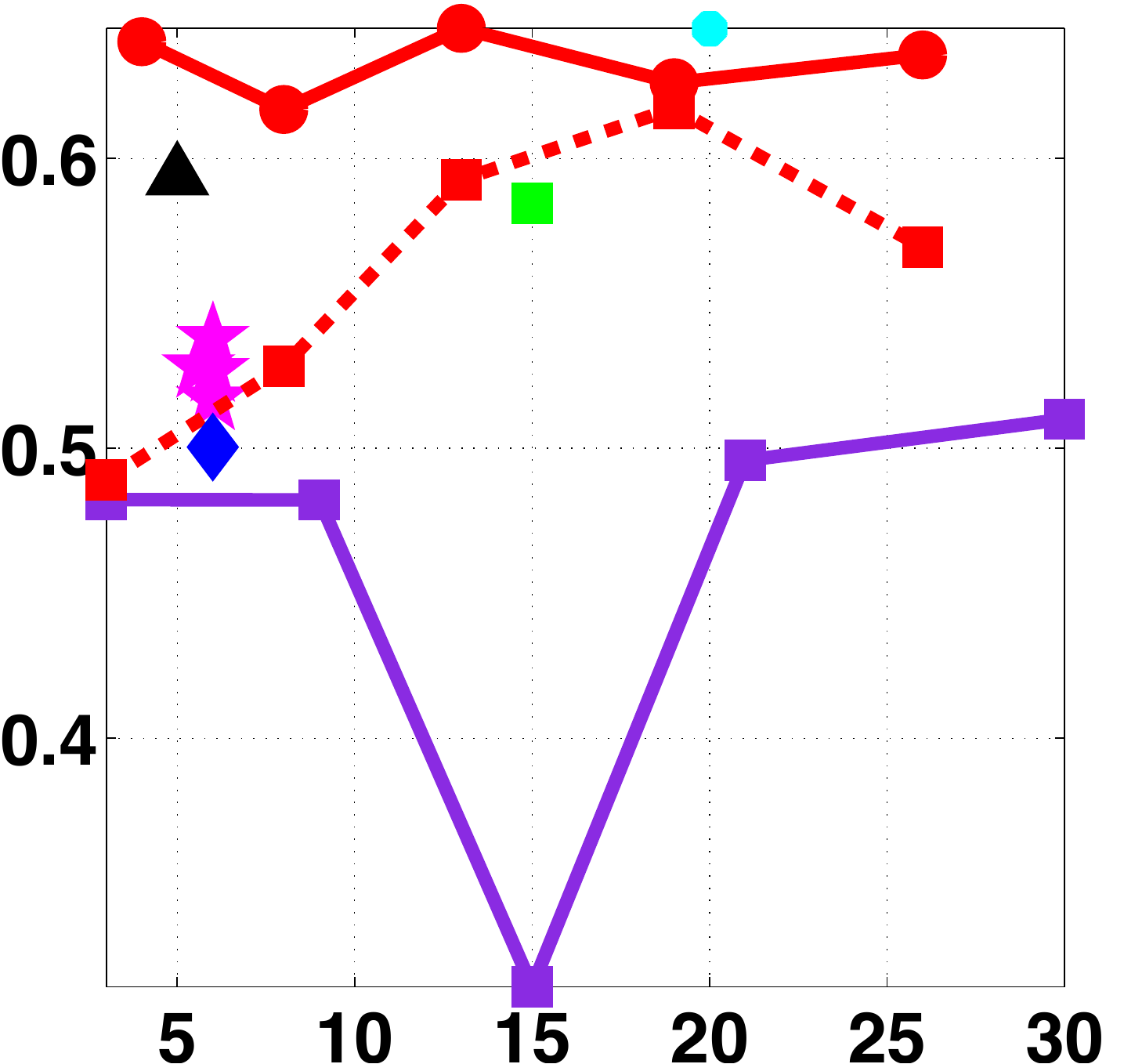} & 
		\includegraphics[width=0.25\textwidth]{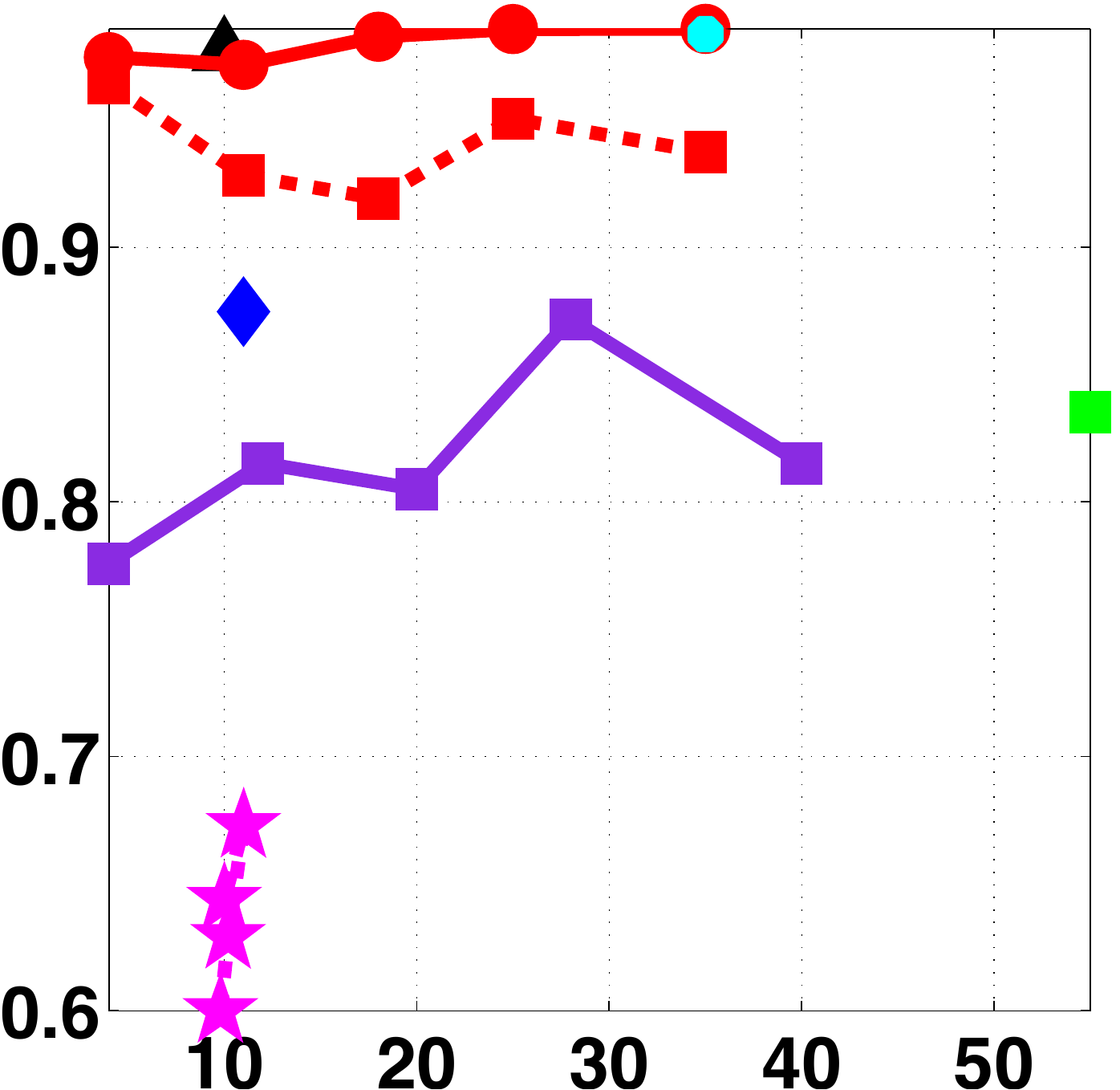} &
		\includegraphics[width=0.25\textwidth]{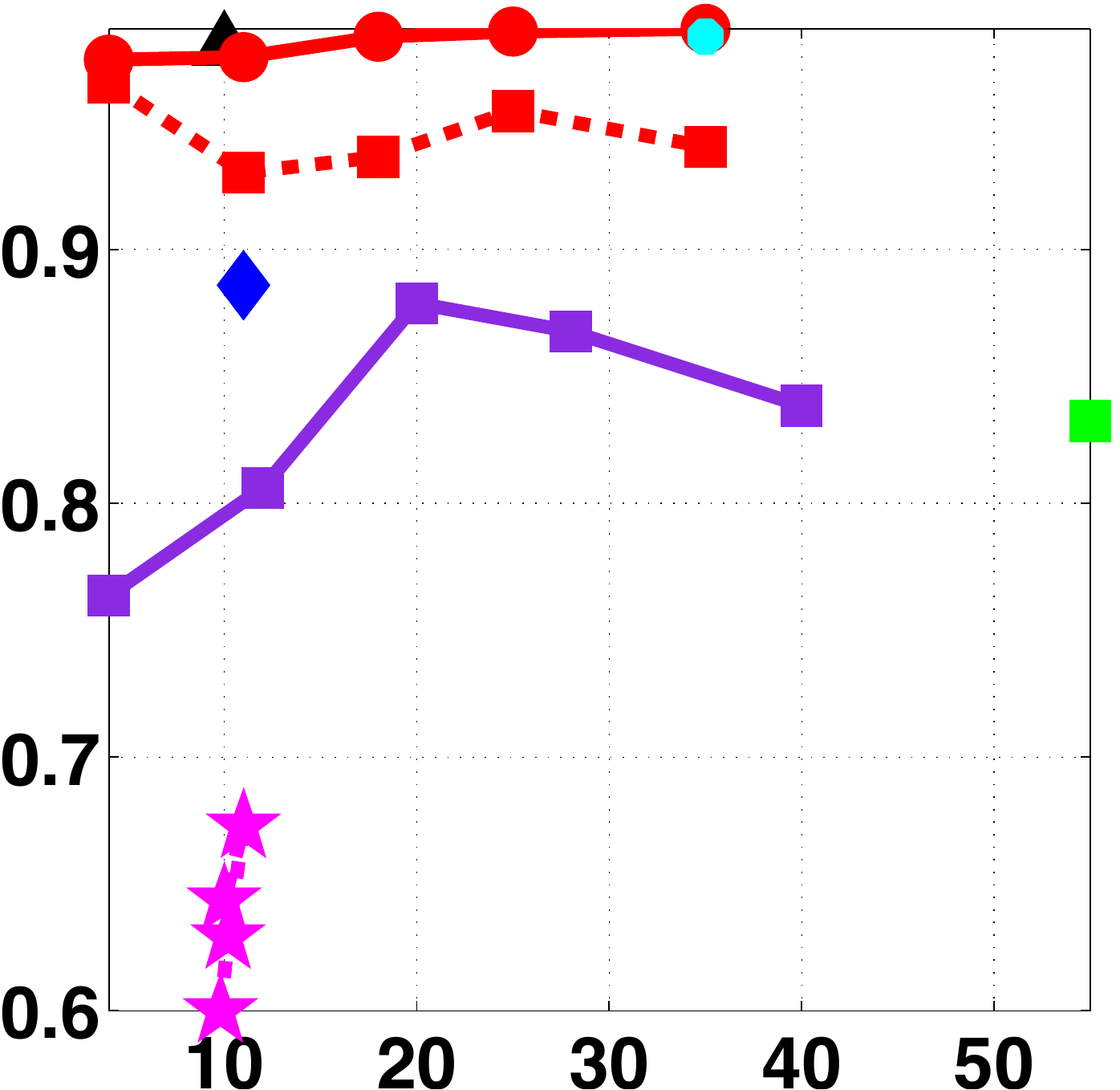} \\
								(e) \textit{Glass} dataset HD&(f) \textit{Glass} dataset LWD&(j) \textit{Vowel} dataset HD&(h) \textit{Vowel} dataset LWD\\
		\includegraphics[width=0.25\textwidth]{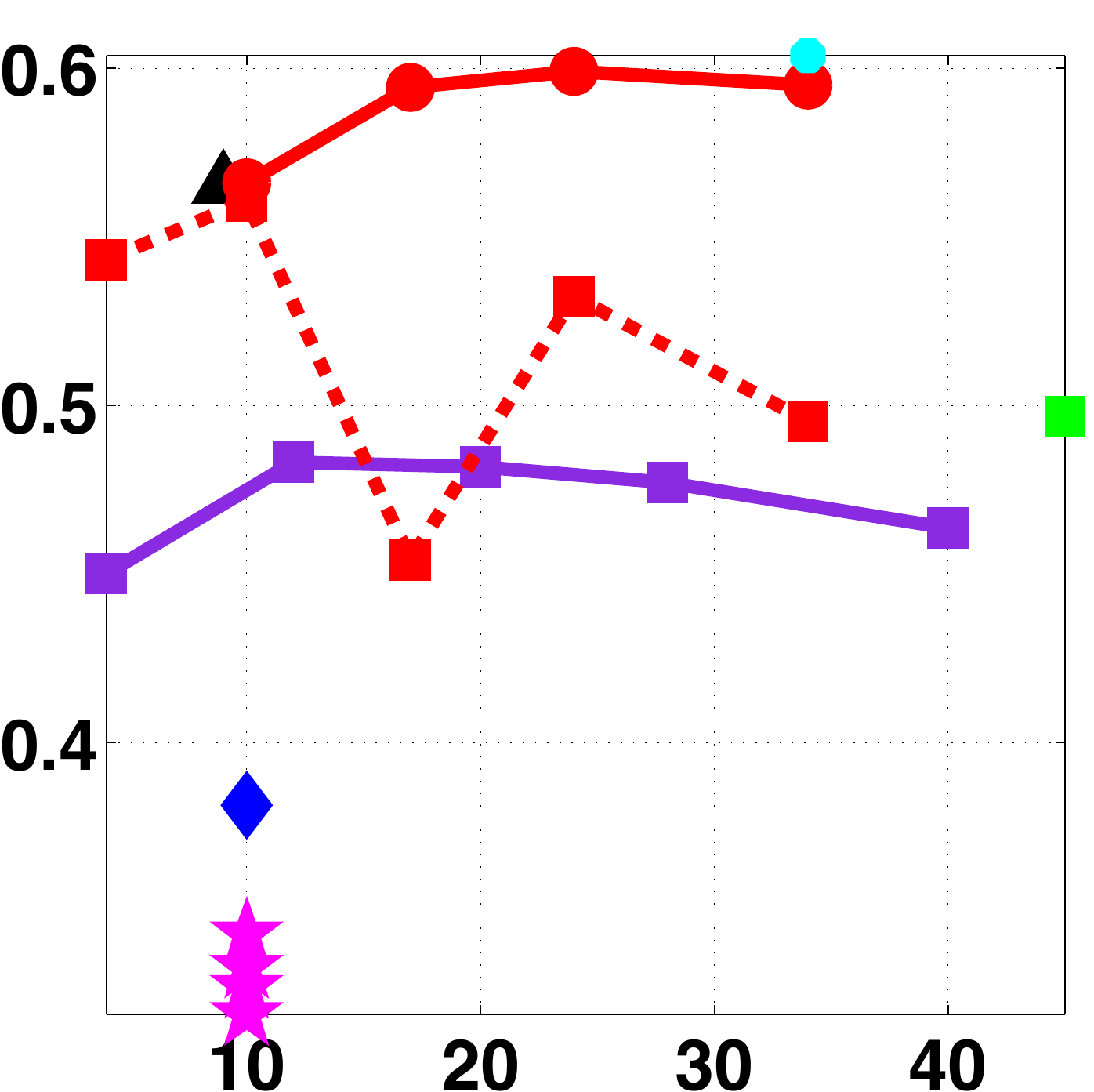} & 
		\includegraphics[width=0.25\textwidth]{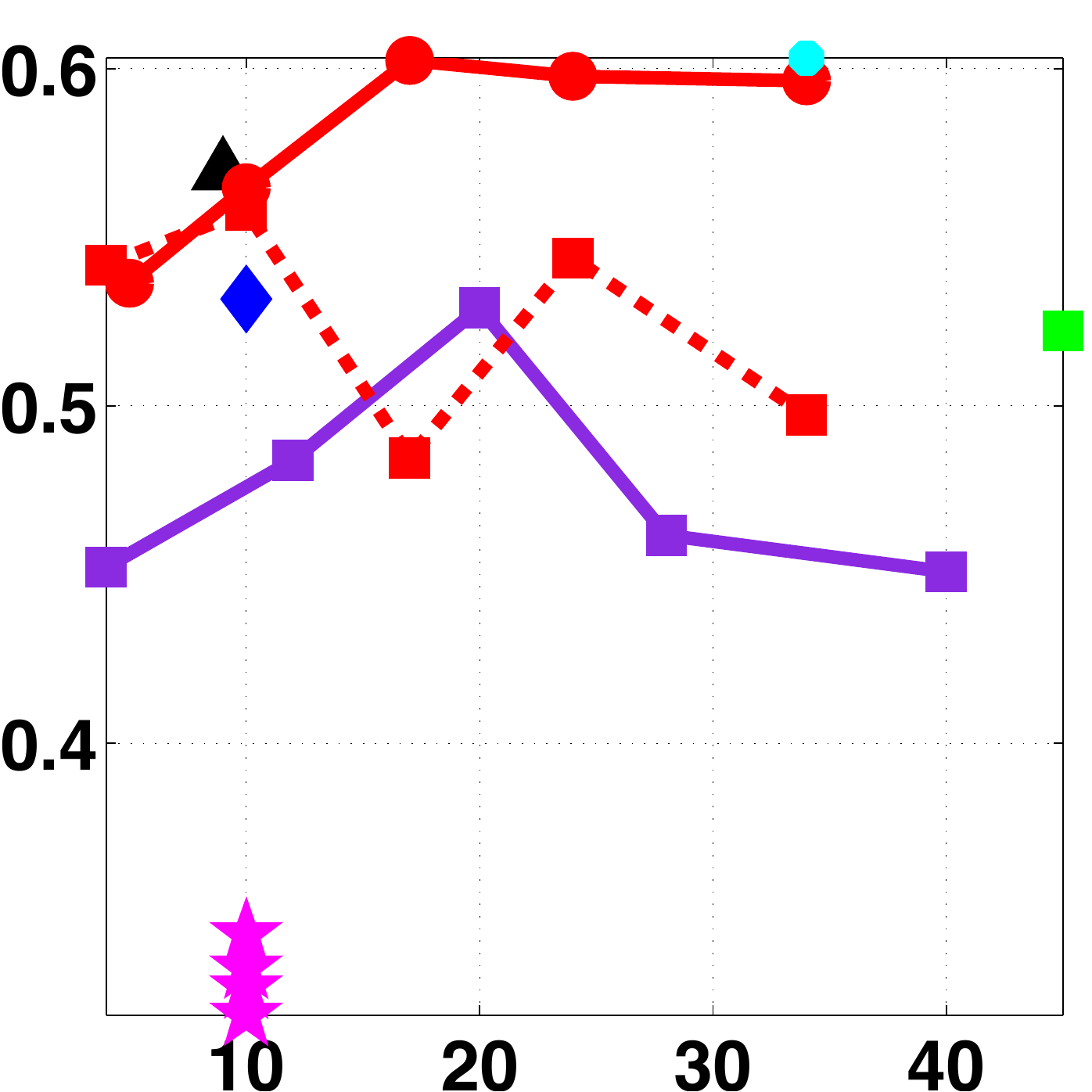} & 
		\includegraphics[width=0.25\textwidth]{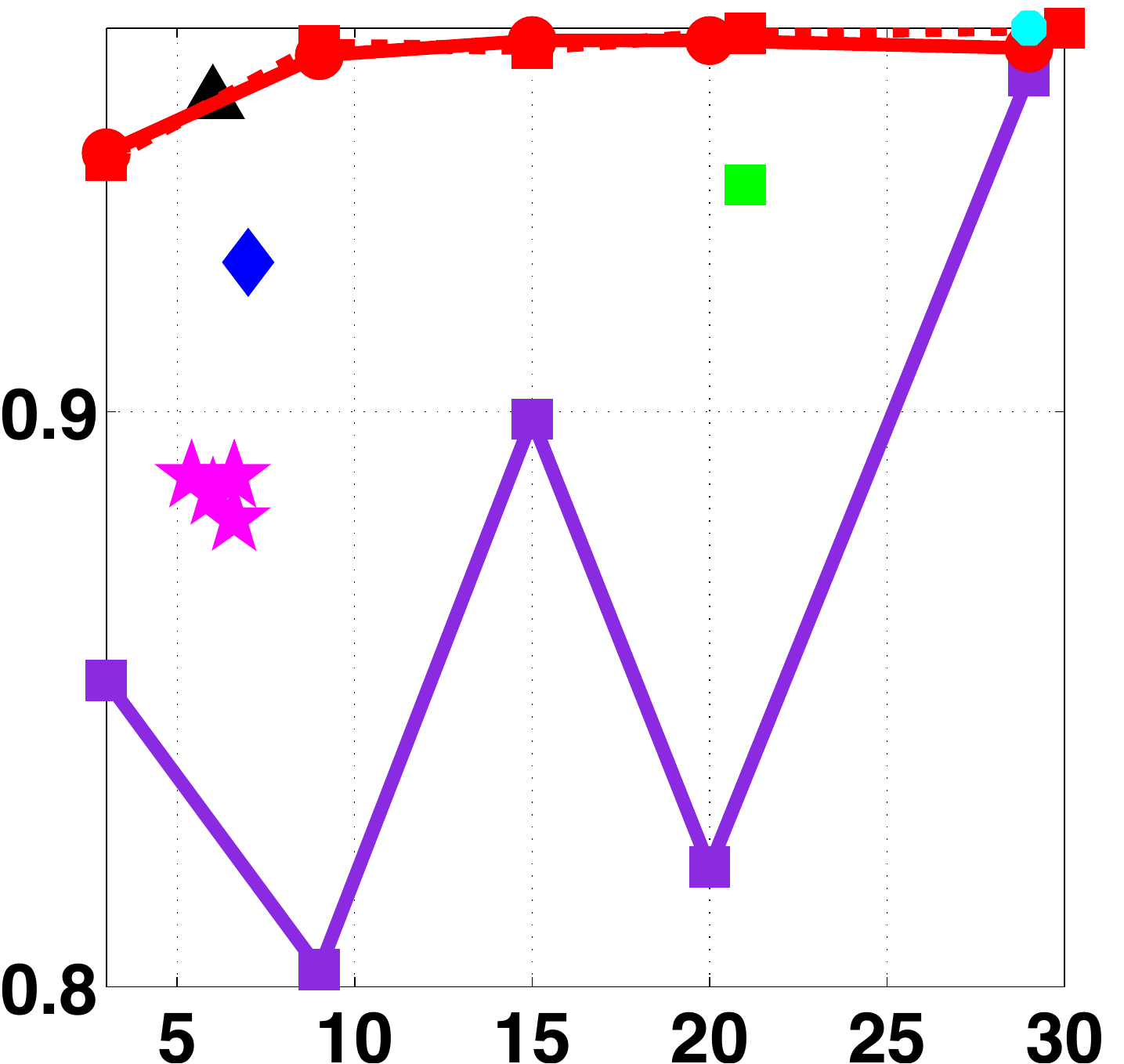} & 
		\includegraphics[width=0.25\textwidth]{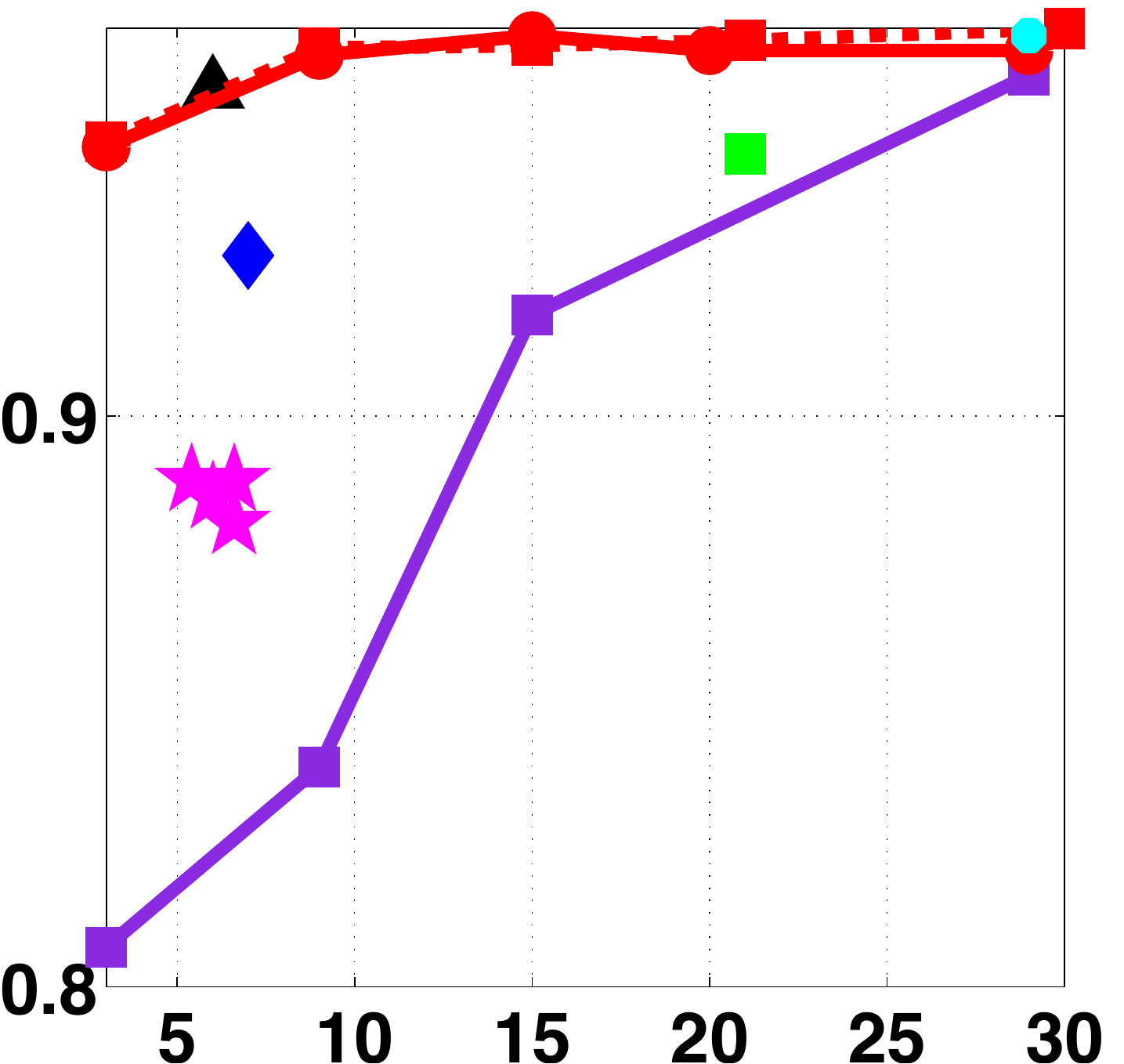} \\
		
		(i) \textit{Yeast} dataset HD&(j) \textit{Yeast} dataset LWD&(k) \textit{Segmentation} dataset HD&(l) \textit{Segmentation} dataset LWD\\
		\includegraphics[width=0.25\textwidth, height= 4.5 cm]{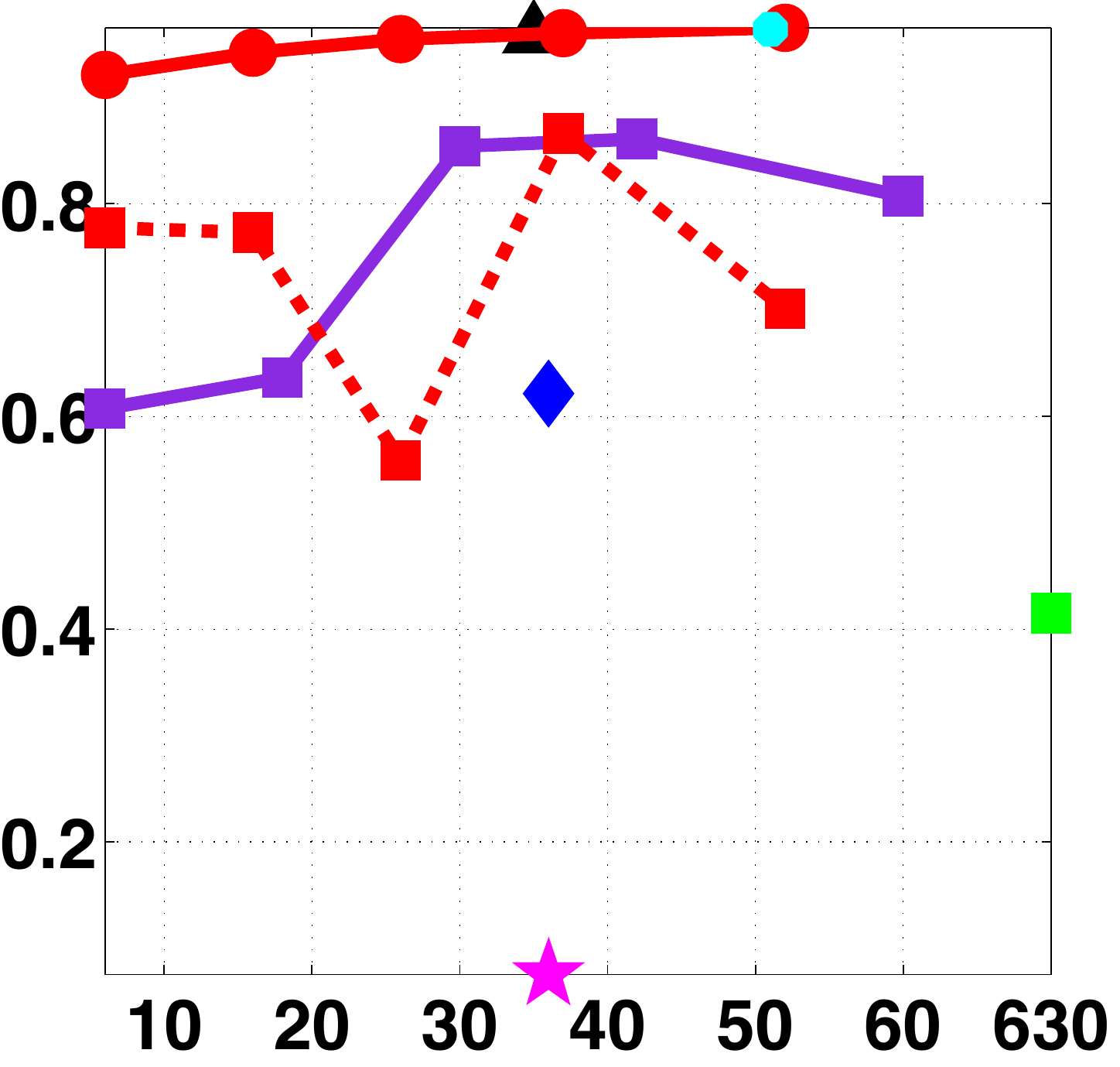} & 
		\includegraphics[width=0.25\textwidth, height= 4.5 cm]{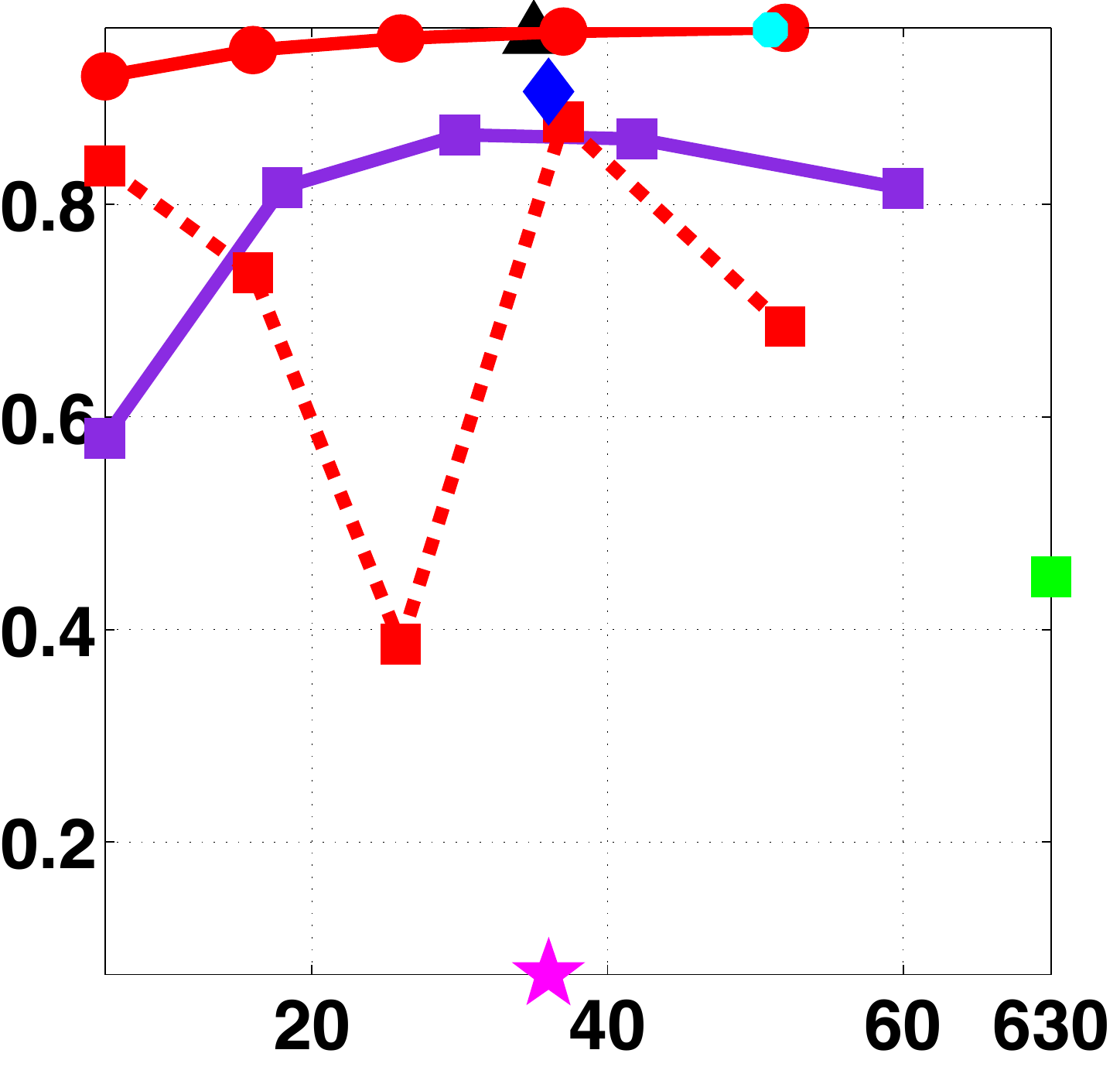} & 
		\includegraphics[width=0.25\textwidth, height= 4.5 cm]{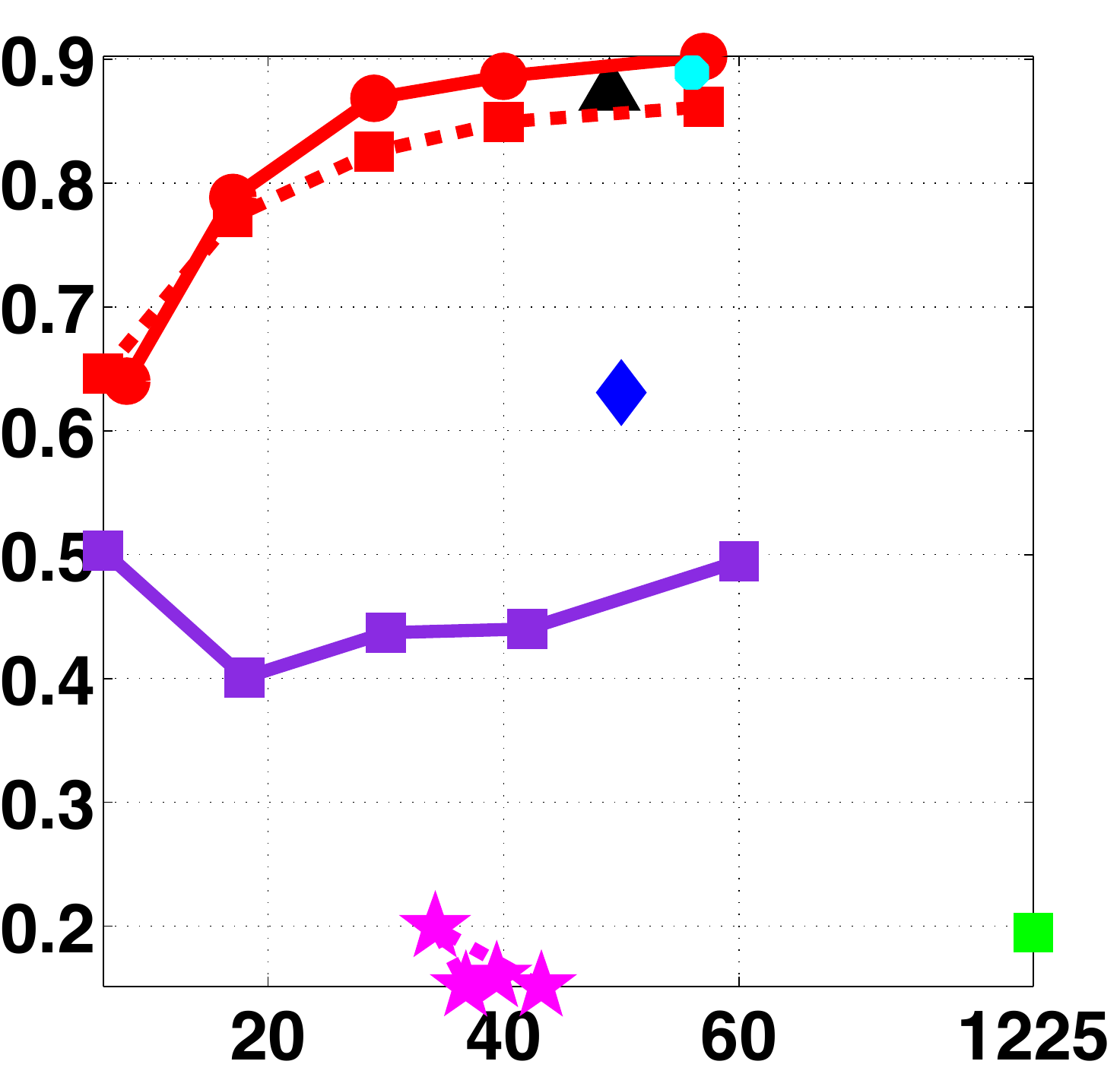} &
		\includegraphics[width=0.25\textwidth, height= 4.5 cm]{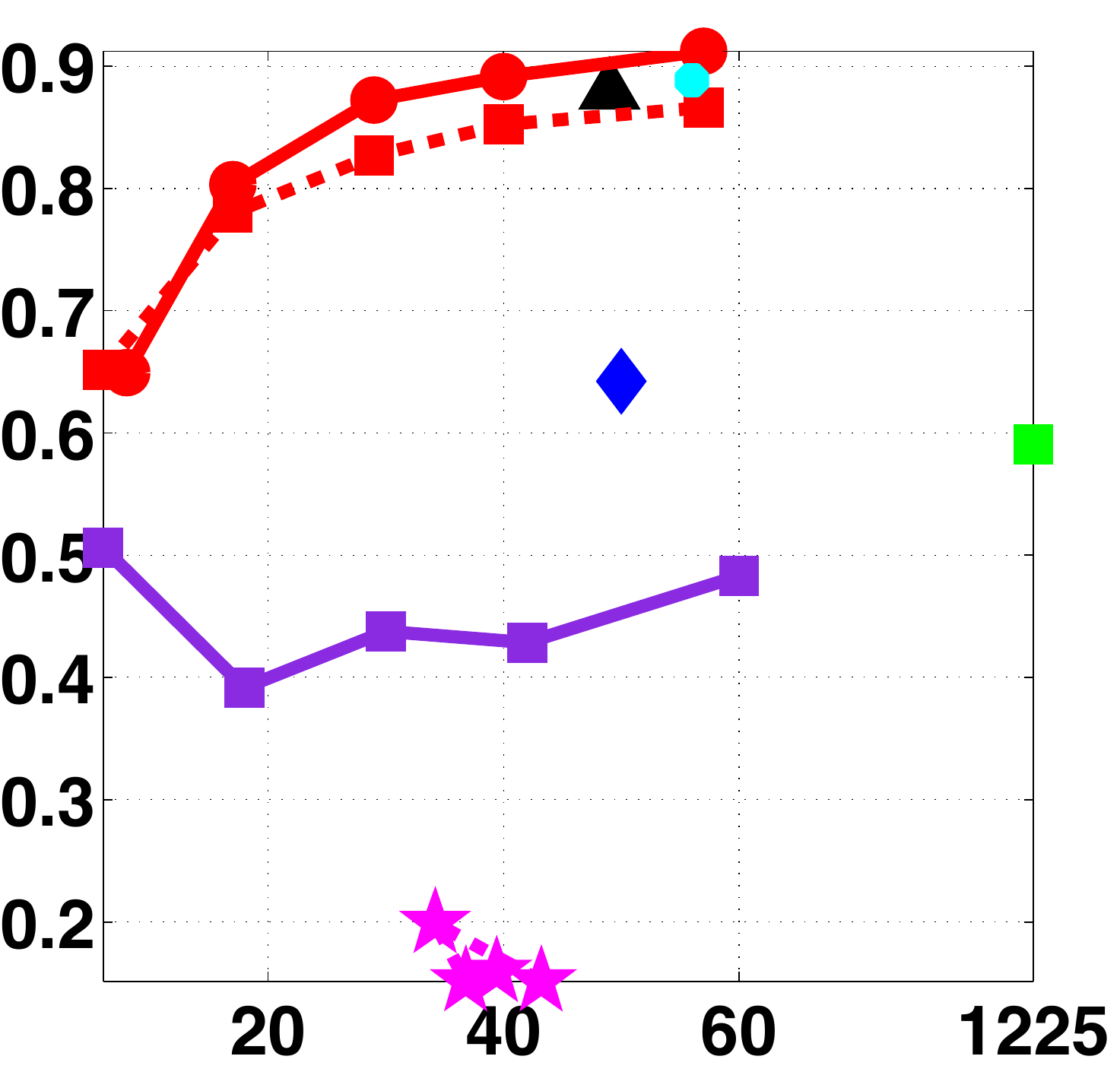} \\
				(m) \textit{Traffic} dataset HD&(n) \textit{Traffic} dataset LWD&(o) \textit{ARFace} dataset HD&(p) \textit{ARFace} dataset LWD\\
		
	\end{tabular}
	\includegraphics[width=12 cm]{legend.pdf} \\
\caption{Multi-class classification accuracy (y axis) as a function of the number of dichotomies for all datasets and both decoding measures (x axis).}
	\label{fig:results_correction}
\end{figure*}

In Figure \ref{fig:results_svs} we show the multi-class classification accuracy as a function of the relative computational complexity for all datasets using both Hamming decoding (HD) and Loss-Weighted Decoding (LWD). We used non-linear SVM classifiers and we define the relative computational complexity as the number of unique Support Vectors (SVs) yielded for each method, as in \cite{iccv2011gao}. For visualization purposes we use an exponential scale and normalize the number of SVs by the maximum number of SVs obtained by a method in that particular dataset. In addition, although the code length cannot be considered as an accurate measure of complexity when using non-linear classifiers in the feature space, it is the only measure of complexity that is available prior to learning the binary problems and designing the coding matrix. In this sense, we show in Figure \ref{fig:results_correction} the classification results for all datasets as a function of the code length $l$, using both Hamming decoding (HD) and Loss-Weighted Decoding (LWD). Figures \ref{fig:results_svs} and \ref{fig:results_correction}  and show how the proposed \textit{ECF-H} obtains in most of the cases better performance than state-of-the-art approaches even with reduced computational complexity. In addition, in most datasets the \textit{ECF-H} is able to boost the boundaries of those classes prone to error, the effect of this is that it attains higher classification accuracies than the rest of methods paying the prize of an small increase on the relative computational complexity. Specifically, we can see how on \textit{Glass} dataset, \textit{Vowel}, \textit{Yeast}, \textit{Segmentation} and \textit{Traffic} datasets (Figs. \ref{fig:results_svs}(e)-(f) and \ref{fig:results_correction}(e)-(f), respectively), the proposed method outperforms the rest of the approaches while yielding a comparable or even lower computational complexity, independently of the decoding function used. We also can see that the \textit{RAND} and \textit{ECF-E} methods present erratic behaviours. This is expected for the random coding design, since incrementing the number of SVs or dichotomies does not imply an increase in performance if the dichotomies are not carefully selected. On the other hand, the reason why \textit{ECF-E} is not stable is not completely straightforward. \textit{ECF-E} focus its design in dichotomies that are very easy to learn, allocating correction to those classes that are separable. 
We hypothesize that when these dichotomies become harder (there exists a limited number of easy separable partitions) to learn the addition of a difficult dichotomy harms the performance by adding confusion to previously learned dichotomies until proper error-correction is allocated. On the other hand, we can see how \textit{ECF-H} usually shows a more stable behaviour since it focuses on categories that are prone to be confused. In this sense, we expect that the addition of dichotomies will increase the correction. Finally, it is worth noting that the Spectral ECOC method yields a code length of  $l=k-1$, corresponding to the full eigendecomposition. Our proposal defines coding matrices which ensure to follow the design denoted by $\D$, fulfilling ECOC properties.

\begin{figure*}[!t]
	\centering
	\setlength{\tabcolsep}{0pt}
	\begin{tabular}{ccc}
		\centering
		\includegraphics[width=0.32\textwidth]{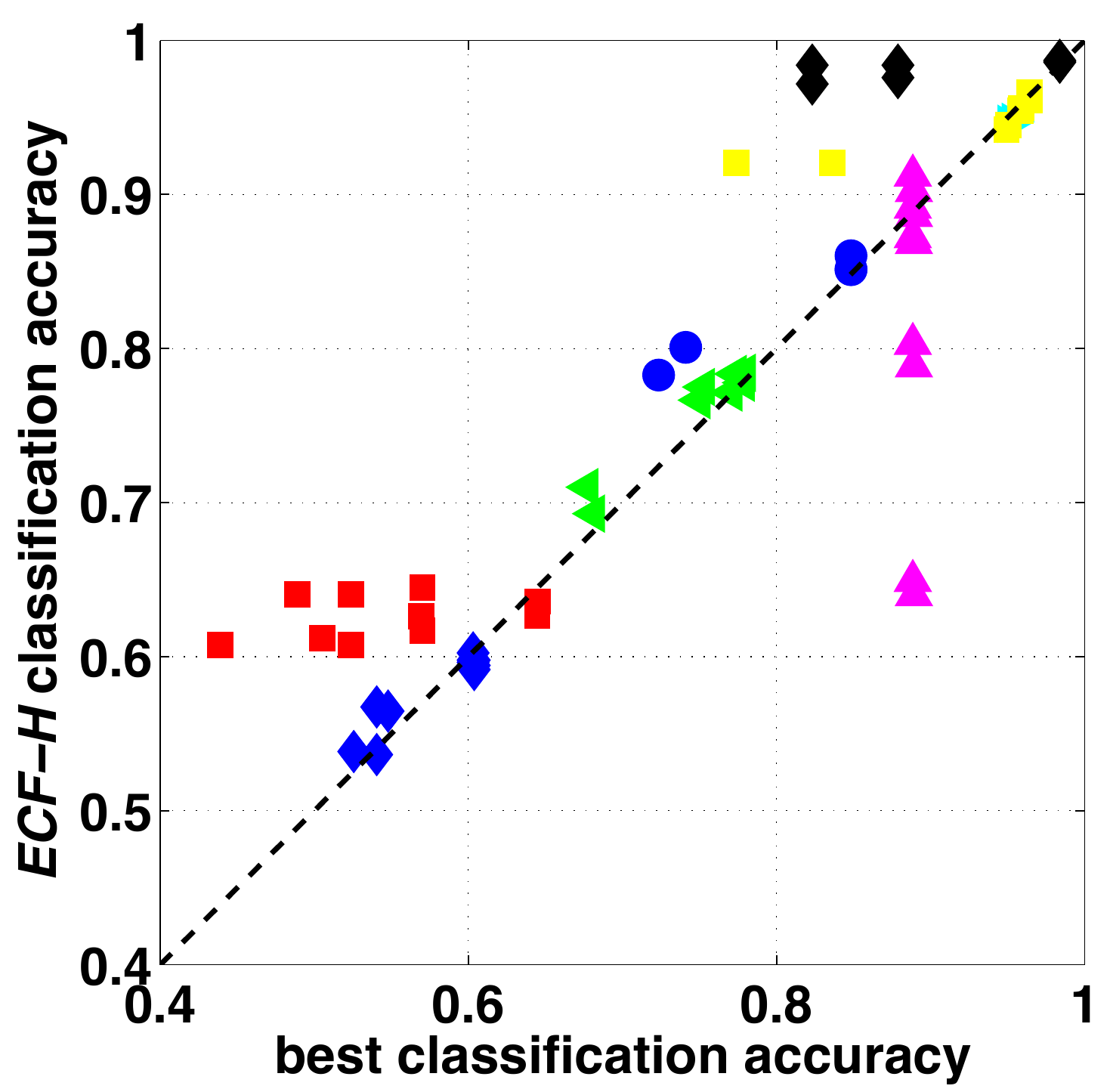}&
		\includegraphics[width=0.32\textwidth]{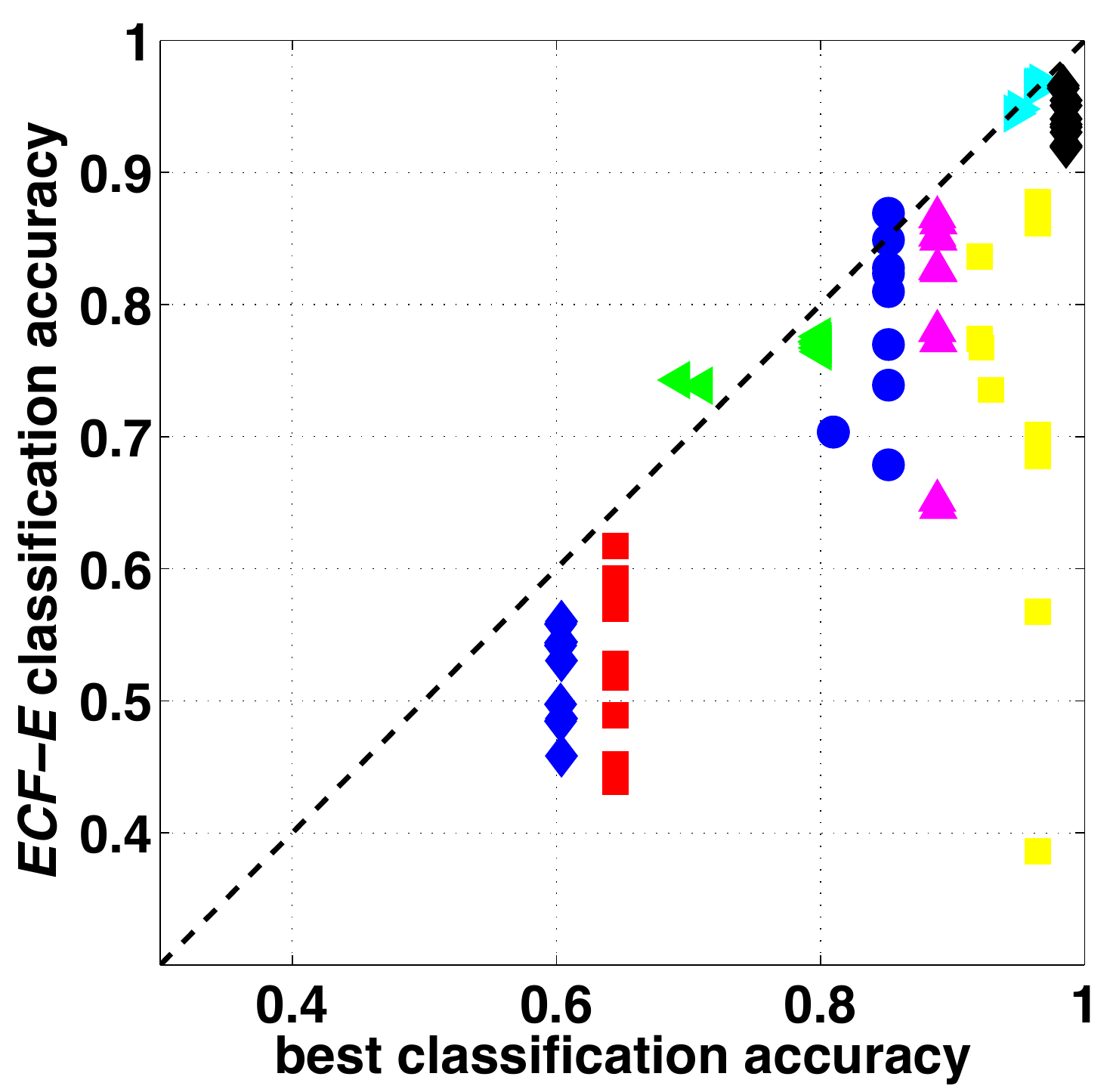}& \includegraphics[width=0.32\textwidth]{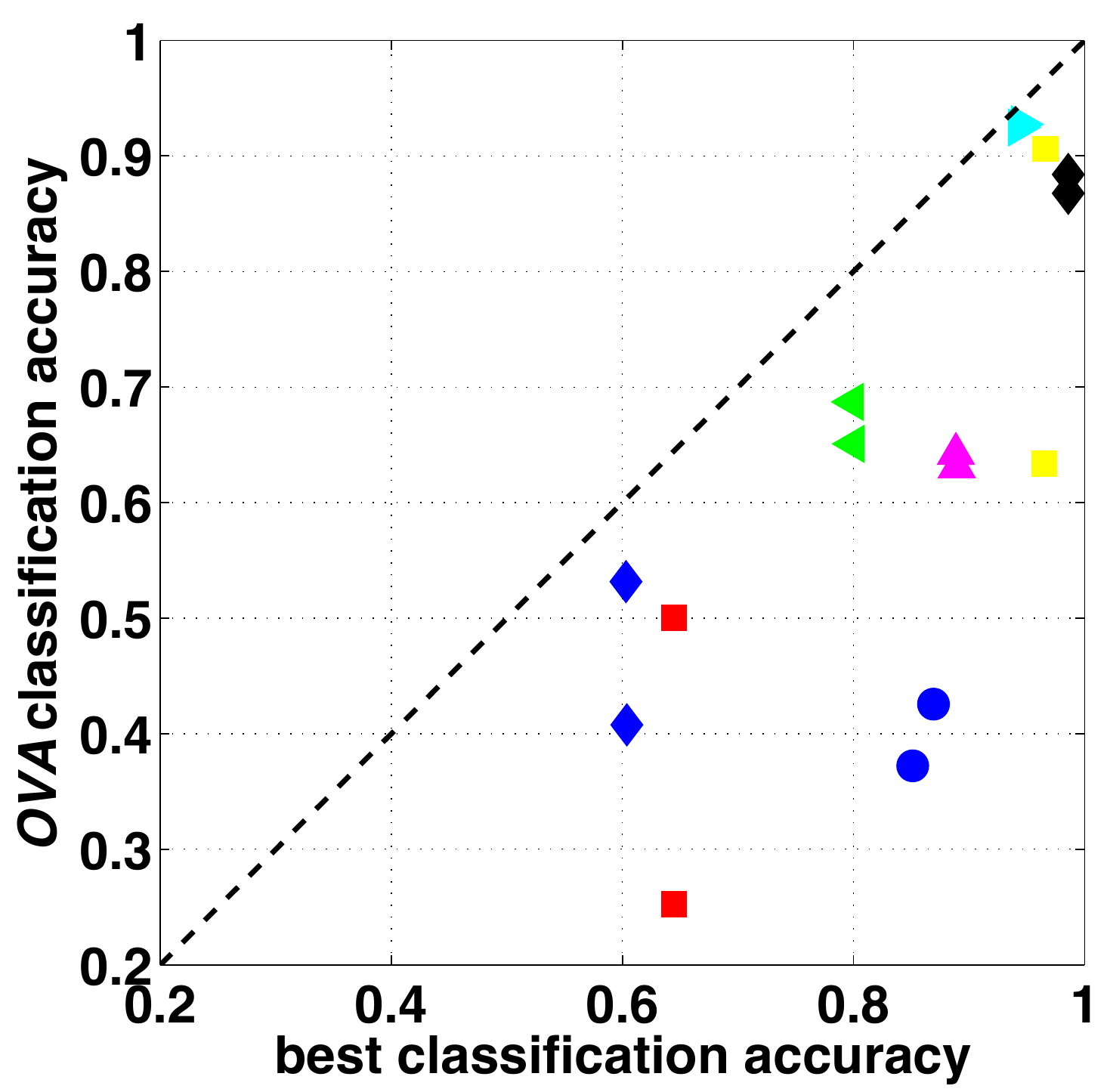}\\
		(a) &(b)  &(c)  \\
		\includegraphics[width=0.32\textwidth]{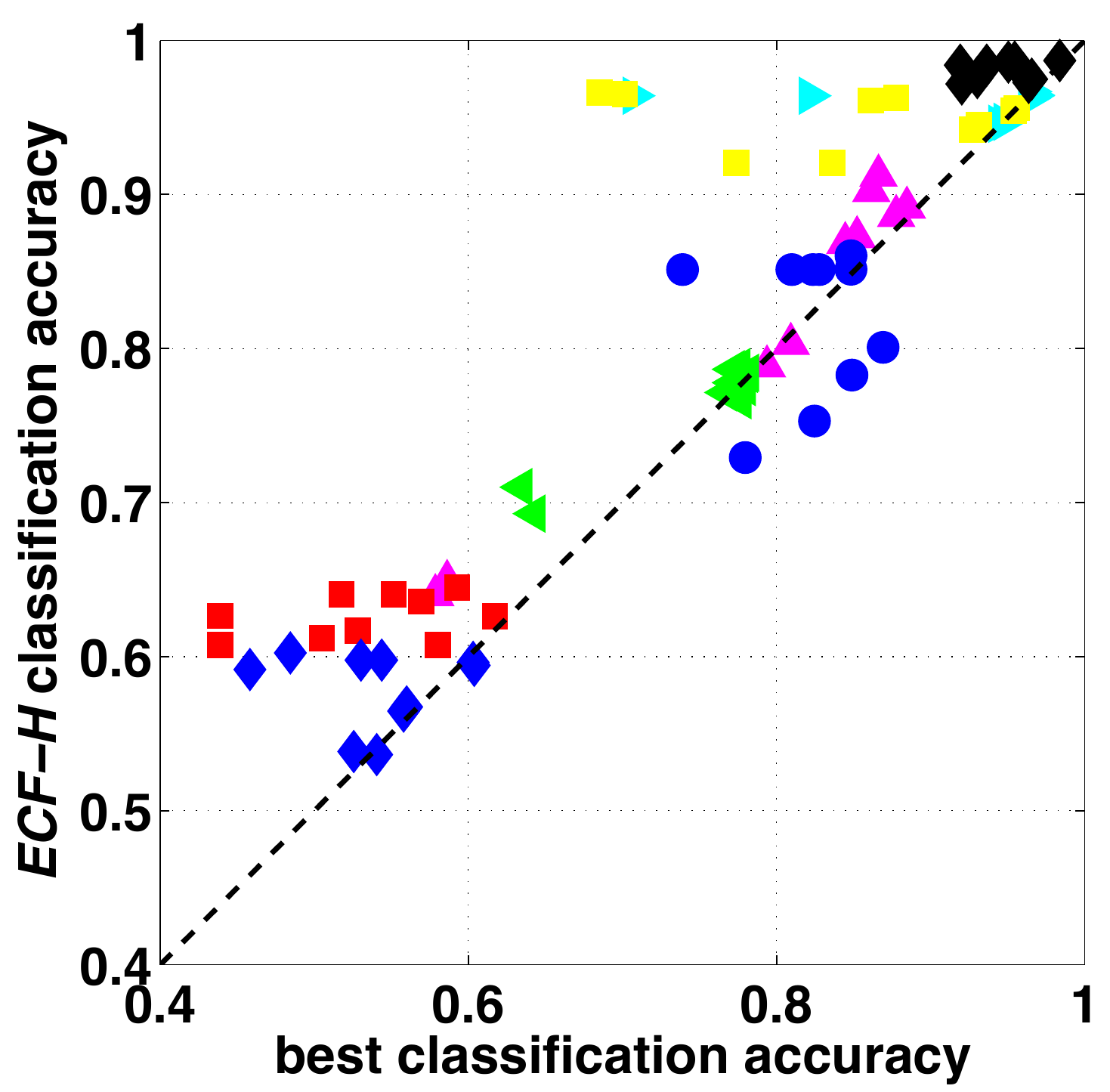}&
		\includegraphics[width=0.32\textwidth]{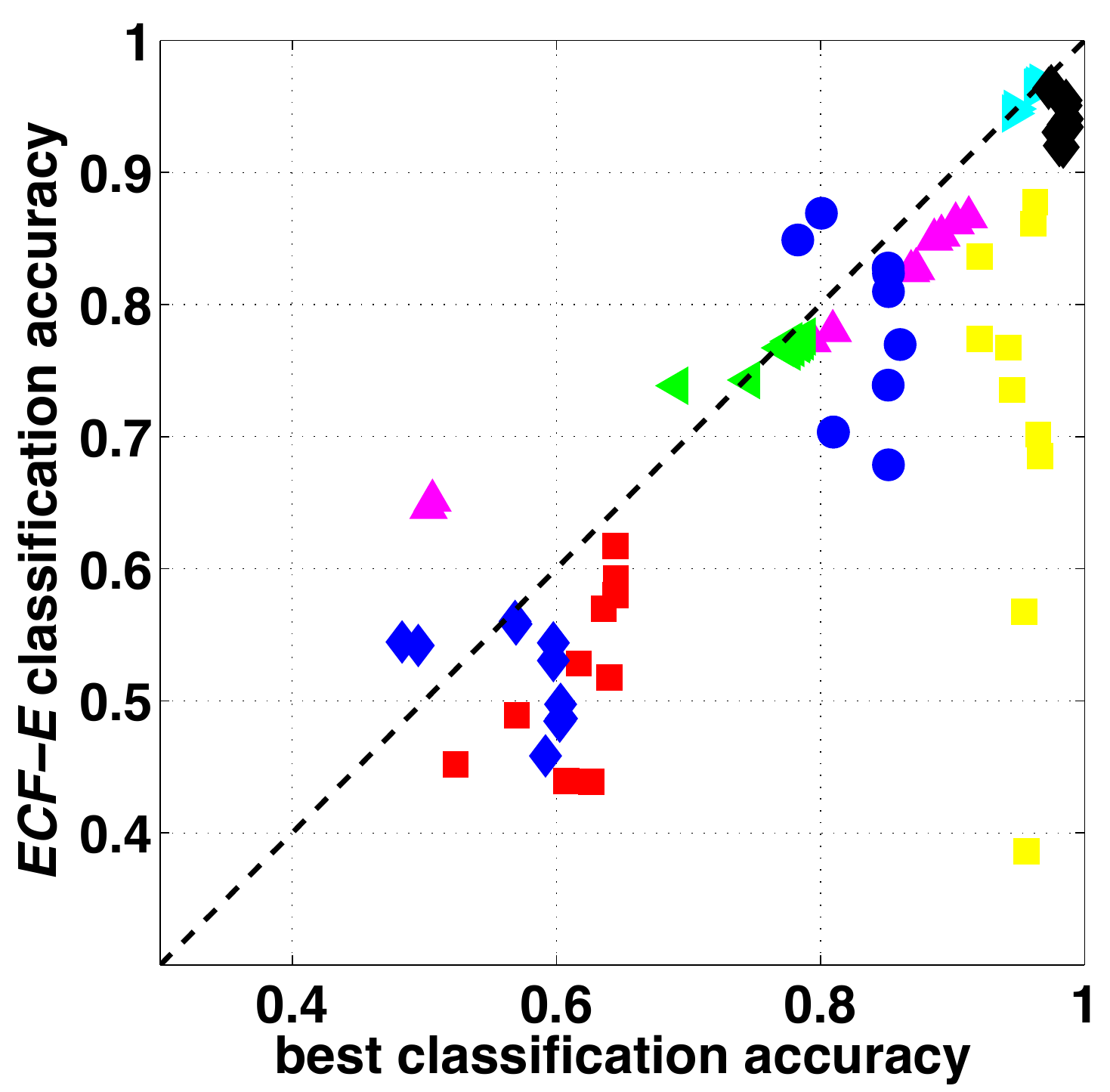}& \includegraphics[width=0.32\textwidth]{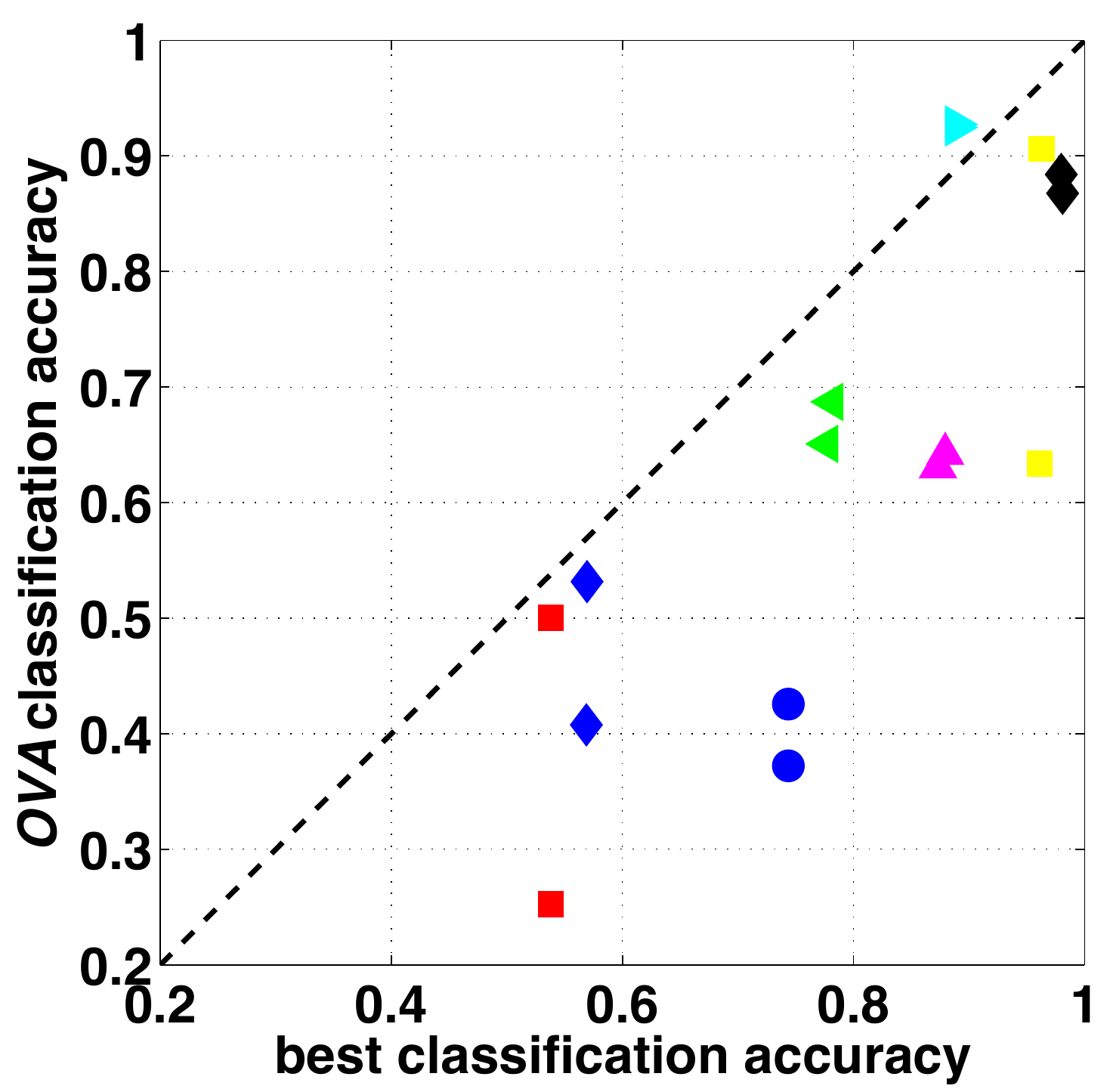}\\
		(d) &(e) &(f)\\
	\end{tabular}
	\vspace{-0.2 cm}
	\caption{(a) Summary of performance of \textit{ECF-H} method over all datasets using the number of SVs and the number of dichotomies as the measure of complexity, respectively for \textit{ECF-H} (a)(d), \textit{ECF-E} (b)(e) and \textit{OVA} (c)(f).} 
	\label{fig:summary}
\end{figure*}

As a summary, we show in Figure \ref{fig:summary} a comparison in terms of classification accuracy for different methods over all datasets. We compare the classification accuracy of a selected method for both decodings (at different operating complexities if available) versus the best performing method in a range of $\pm{5\%}$ of the operative complexity. For consistency we show the comparison using both the number of SVs and the number of dichotomies as the computational complexity. If the compared method dominates in most of the datasets it will be found above the diagonal. In Figures \ref{fig:summary}(a) and \ref{fig:summary}(d) we compare \textit{ECF-H} with the best performant of the rest of the methods and see that \textit{ECF-H} outperforms the rest of the methods $62\% - 70 \%$ of the times depending on the complexity measure. This implies that \textit{ECF-H} dominates most of the methods in terms of performance by focusing on those classes that are more prone to error regardless of the complexity measure used (number of SVs or number of dichotomies). In addition, when repeating the comparison for \textit{ECF-E} in Figures \ref{fig:summary}(b) and \ref{fig:summary}(e) we see that the majority of the datasets are clearly below the diagonal (\textit{ECF-E} is the most suitable choice $10\% -17\%$ of times). Finally, Figures \ref{fig:summary}(c) and \ref{fig:summary}(f) show the comparison for \textit{OVA}, which is a standard method often defended by its simplicity \cite{rifkin}. We clearly see how it never outperforms any method and it is not the recommended choice for almost any dataset. In Table \ref{tab:finalresults} we show  the percentage of wins for all methods\footnote{The R-H method \cite{iccv2011gao} is far less complex than the compared methods, however we compare it to the to the closest operating complexity for each of the rest of the methods.}, in increasing order of complexity averaged over all datasets. Note how, \textit{ECF-H} denoted by \textit{H} in the table although being the third less complex method outperforms by far the rest of the methods with an improvement of at least $12\%-20\%$ in the worst case. In conclusion, the experimental results show that \textit{ECF-H} yields ECOC coding matrices which obtain comparable or even better results than state-of-the-art methods with similar relative complexity. Furthermore, by a allowing a small increase in the computational complexity when compared to state-of-the-art methods, ECF is able to obtain better classification results by boosting the boundaries of classes that are prone to be confused.

\begin{table}
\scriptsize
\caption{Percentage of wins over all datasets for each method using as a complexity measure the number SVs and the number of classifiers. Last row shows the average complexity of each method over all datasets. Abbreviations: ECF-H (H), ECF-E (E), OVA (A), OVO (O), DENSE (D), RAND (R), S-ECOC(S).}	
\vspace{-0.3 cm}
\begin{tabular}{|c|c|c|c|c|c|c|c|c|}
\hline
\textbf{Method}&\textit{R-H$^*$} & \textit{S}&\textit{H}&\textit{E}&\textit{D}& \textit{R} & \textit{A}&\textit{O}\\ \hline

Win $\%$ SVs& 0.0 & 22.5& \textbf{62.1} & 10.3 & 50.0 & 5.7 & 14.2 & 25.0   \\ \hline

Win $\%$ nclass.& 0.0 & 48.5& \textbf{70.0} & 17.5 & 25.0 & 6.9 & 12.5 & 16.6 \\ \hline

Avg. Comp. & \textbf{0.58} & 0.87& 0.88 & 0.89 & 0.91 & 0.92 & 0.99 & 0.99 \\ \hline

\end{tabular}
\label{tab:finalresults}
\end{table}

\section{Conclusions} \label{sec:conclusions}

We presented the Error-Correcting Factorization method for multi-class learning which is based on the Error-Correcting Output Codes framework. The proposed method factorizes a design matrix of desired correction properties into a discrete Error-Correcting component  consistent with the design matrix. ECF is a general method for building an ECOC multi-class classifier with desired properties, which can be either directly set by the user or obtained from data using a priori inter-class distances. We note that the proposed approach is not a replacement for ECOC codings, but a generalized framework to build ECOC matrices that follow a certain error-correcting criterion design. The Error-Correcting Factorization is formulated as a minimization problem which is optimized using a constrained Coordinate Descent, where the minimizer of each coordinate is the solution to a least-squares problem with box and linear constraints that can be efficiently solved. By analyzing the approximation error, we empirically show that although ECF is a non-convex optimization problem, the optimization is very efficient. We performed experiments using ECF to build ECOC matrices following the common trend in state-of-the-art works, in which the design matrix priorized the most separable classes. In addition, we hypothesized and showed that a more beneficial situation is to allocate the correction capability of the ECOC to those categories which are more prone to confusion. Experiments show that when ECF is used to allocate the correction capabilities to those classes which are prone to confusion we obtain higher accuracies than state of the art methods with efficient models in terms of the number of Support Vectors and dichotomies.

Finally, there still exists open questions that require a deeper analysis for future work. The results obtained raise a fair doubt regarding the right allocation of error correcting power in several methods found in literature where ECOC designs are based on the premise of boosting the classes which are easily separable. In the light of these results, we may conjecture that a careful allocation of error correction must be made in such a way that balances two aspects: on one hand, simple to classify boundaries must be handled properly. On the other hand, the error correction must be allocated on difficult classes for the ensemble to correct possible mistakes. In addition, it would be interesting to study which are the parameters that affect the suitability of the \textit{no class is left behind} and the \textit{hard classes are left behind} one. Finally we could consider ternary matrices and further regularizations.


\bibliographystyle{plain}
\bibliography{BibFile}

 \begin{IEEEbiography}[{\includegraphics[width=1in,height=1.25in,clip,keepaspectratio]{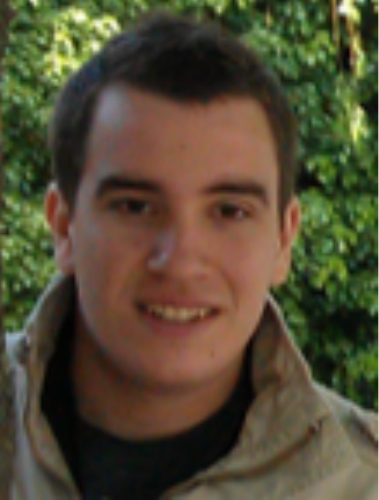}}]{Miguel \'{A}ngel Bautista} received his B. Sc. and M. Sc. degrees in Computer Science and Artificial Intelligence from Universitat de Barcelona and Universitat Politécnica de Catalunya  respectively in 2010. He is a research member at Computer Vision Center at Universitat Autonoma de Barcelona,  Applied Math and Analysis Dept. at Universitat de Barcelona and BCN Perceptual Computing Lab and Human Pose Recovery and Behavior Analysis Group at University of Barcelona . In 2010 Miguel Angel received the first prize from the Catalan Association of Artificial Intelligence Thesis Awards. Currently Miguel Angel is pursuing a Ph. D in Error Correcting Output Codes as a theoretical framework to treat multi-class and multi-label problems. His interests are, between others, Machine Learning, Computer Vision, Convex Optimization and its applications into Human Gesture analysis.
 \end{IEEEbiography}

  \begin{IEEEbiography}[{\includegraphics[width=1in,height=1.25in,clip,keepaspectratio]{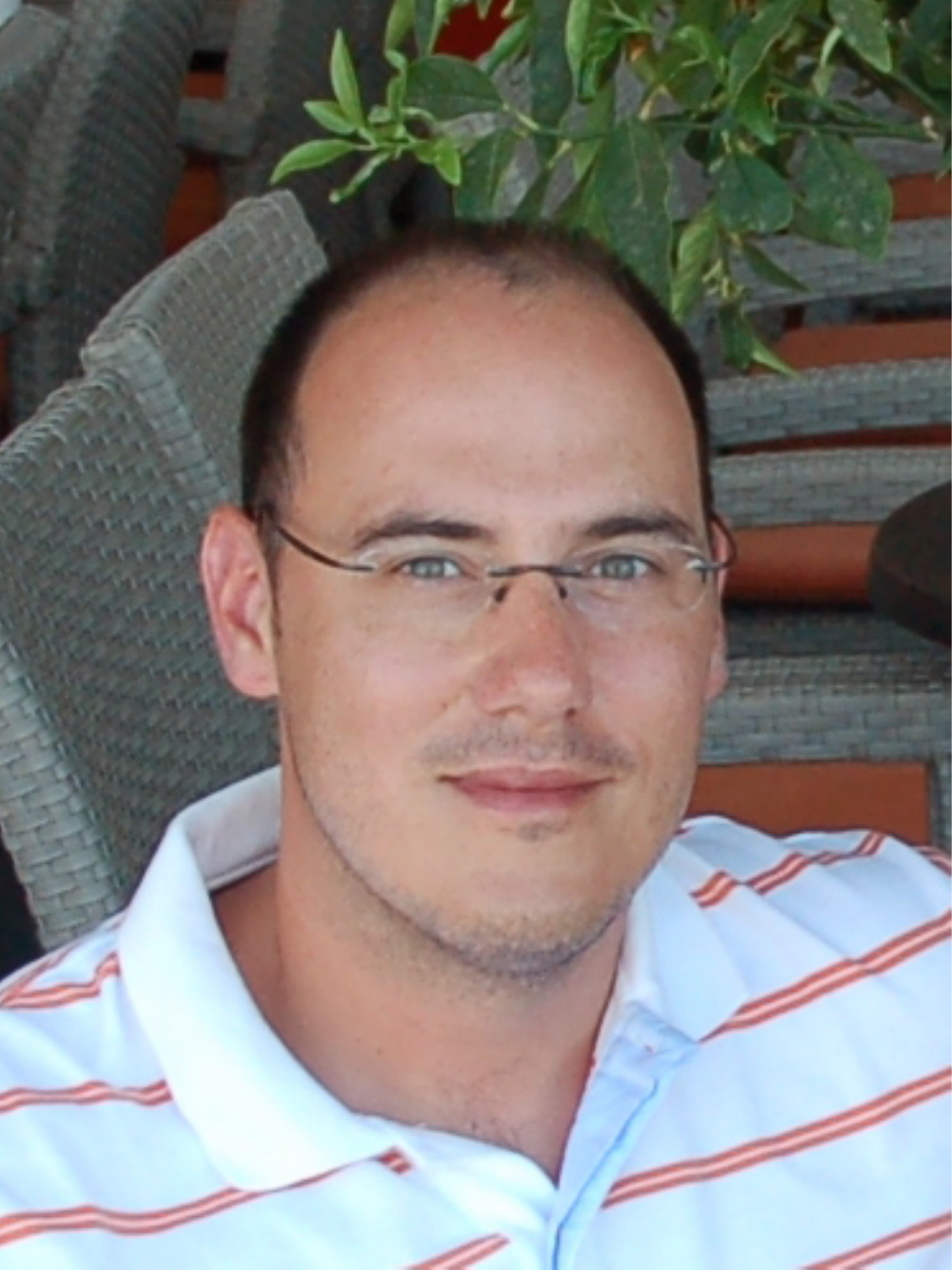}}]{Oriol Pujol}
  	Oriol  Pujol Vila obtained the degree in Telecomunications Engineering in 1998 from the Universitat Politècnica de Catalunya  (UPC). The same year, he joined the Computer Vision Center and the Computer Science Department at Universitat  Autònoma de Barcelona  (UAB). In 2004 he received the Ph.D. in Computer Science at the UAB on work in deformable models, fusion of supervised and unsupervised learning and intravascular ultrasound image analysis. In 2005 he joined the Dept. of Matemàtica Aplicada i Anàlisi at Universitat de Barcelona where he became associate professor. He is member of the BCN Perceptual Computing Lab. He has been since 2004 an active member in the organization of several activities related to image analysis, computer vision, machine learning and artificial intelligence.
  \end{IEEEbiography}
 
 \begin{IEEEbiography}[{\includegraphics[width=1in,height=1.25in,clip,keepaspectratio]{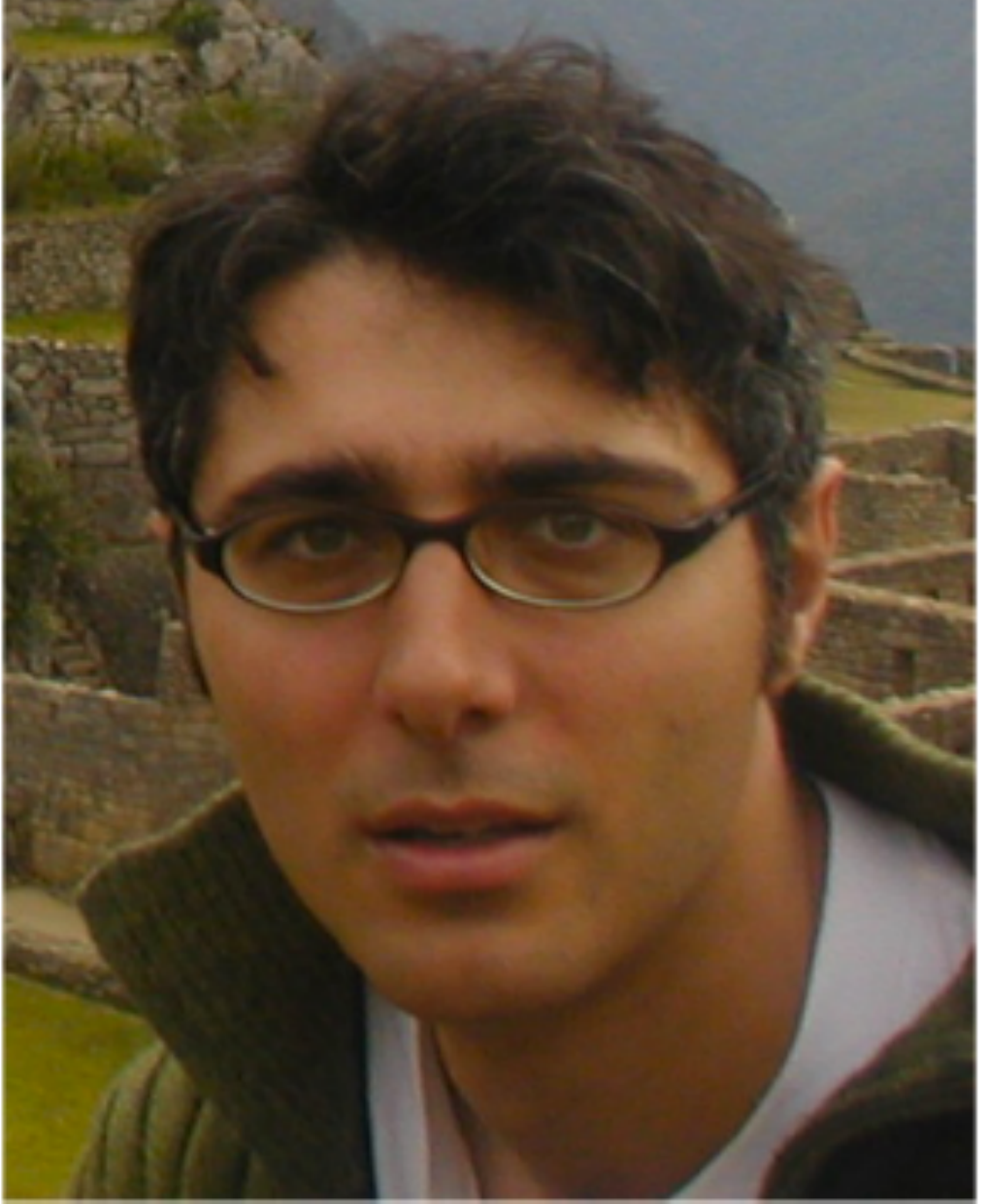}}]{Fernando De la Torre} is an Associate Research
 		Professor in the Robotics Institute at Carnegie Mellon
 		University. He received his B.Sc. degree in
 		Telecommunications, as well as his M.Sc. and Ph.
 		D degrees in Electronic Engineering from La Salle
 		School of Engineering at Ramon Llull University,
 		Barcelona, Spain in 1994, 1996, and 2002, respectively.
 		His research interests are in the fields of
 		computer vision and Machine Learning. Currently,
 		he is directing the Component Analysis Laboratory
 		(http://ca.cs.cmu.edu) and the Human Sensing
 		Laboratory (http://humansensing.cs.cmu.edu) at Carnegie Mellon University.
 		He has over 130 publications in referred journals and conferences. He has
 		organized and co-organized several workshops and has given tutorials at
 		international conferences on the use and extensions of Component Analysis.
 \end{IEEEbiography}

 	 \begin{IEEEbiography}[{\includegraphics[width=1in,height=1.25in,clip,keepaspectratio]{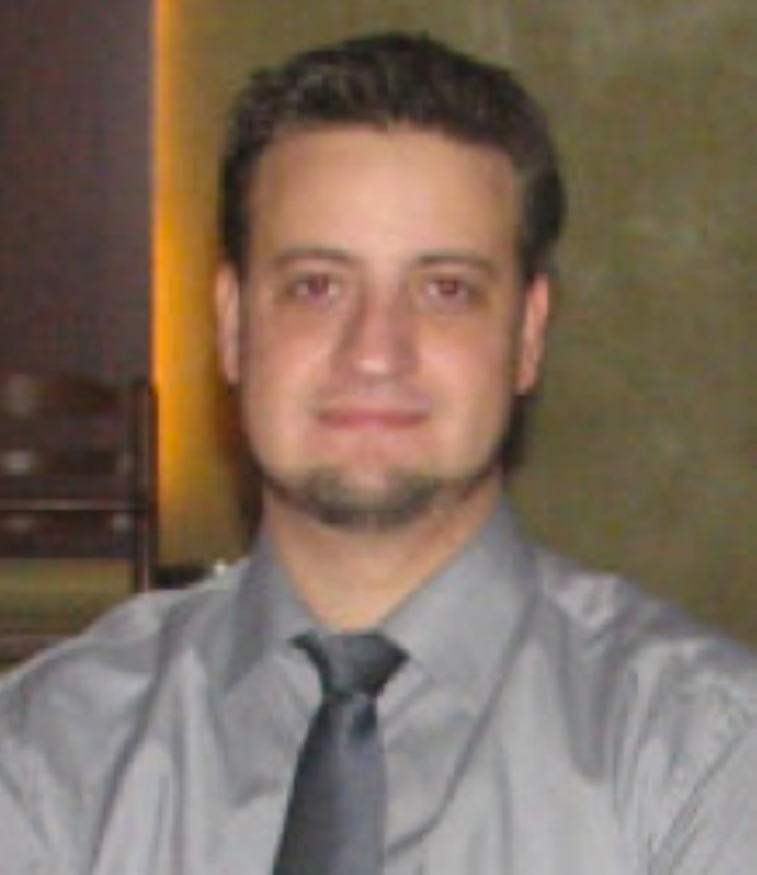}}]{Sergio Escalera}
 	 	Sergio Escalera received the B.S. and M.S. degrees from the Universitat Autònoma de Barcelona (UAB), Barcelona, Spain, in 2003 and 2005, respectively. He obtained the P.h.D. degree on Multi-class visual categorization systems at Computer Vision Center, UAB. He obtained the 2008 best Thesis award on Computer Science at Universitat Autònoma de Barcelona. He lead the Human Pose Recovery and Behavior Analysis Group at University of Barcelona. His research interests include, between others, machine learning, statistical pattern recognition, visual object recognition, and human computer interaction systems, with special interest in human pose recovery and behavior analysis.
 	 \end{IEEEbiography}

\end{document}